\documentclass[11pt,twocolumn]{article}

\usepackage[a4paper,margin=2.2cm,columnsep=0.6cm]{geometry}
\usepackage{setspace}
\usepackage{amsmath,amssymb,amsthm,mathtools,bm}
\usepackage{mathrsfs}

\usepackage{tikz}
\usetikzlibrary{
  arrows.meta,
  calc,
  chains,
  decorations.pathreplacing,
  decorations.markings,
  fit,
  matrix,
  patterns,
  positioning,
  shapes.geometric,
  shapes.multipart,
  backgrounds,
  shadows,
  trees,
  intersections,
  fadings
}
\usepackage{pgfplots}
\pgfplotsset{compat=1.18}
\usepgfplotslibrary{fillbetween,groupplots,colormaps}
\usepackage{pgfplotstable}
\usepackage{graphicx}
\usepackage{float}
\usepackage{subcaption}

\usepackage{booktabs}
\usepackage{multirow}
\usepackage{array}
\usepackage{colortbl}
\usepackage{tabularx}
\usepackage{longtable}
\usepackage{makecell}

\usepackage[utf8]{inputenc}
\usepackage[T1]{fontenc}
\usepackage{microtype}
\usepackage{lmodern}
\usepackage{xcolor}
\usepackage{soul}

\usepackage[hidelinks,colorlinks=true,linkcolor=blue!60!black,citecolor=green!50!black,urlcolor=blue!70!black]{hyperref}
\usepackage[numbers,sort&compress]{natbib}
\usepackage{cleveref}

\usepackage{algorithm}
\usepackage{algpseudocode}

\usepackage{xspace}
\usepackage{etoolbox}
\usepackage{placeins}

\definecolor{deepblue}{RGB}{0,51,102}
\definecolor{deepred}{RGB}{153,0,0}
\definecolor{deepgreen}{RGB}{0,102,51}
\definecolor{lightblue}{RGB}{200,220,240}
\definecolor{lightyellow}{RGB}{255,255,220}
\definecolor{lightgreen}{RGB}{220,240,220}
\definecolor{lightred}{RGB}{255,220,220}
\definecolor{lightorange}{RGB}{255,235,210}
\definecolor{lightpurple}{RGB}{230,220,250}
\definecolor{gptcolor}{RGB}{116,185,255}
\definecolor{llamacolor}{RGB}{162,155,254}
\definecolor{deepseekcolor}{RGB}{0,184,148}
\definecolor{grokcolor}{RGB}{255,118,117}
\definecolor{geminicolor}{RGB}{253,203,110}
\definecolor{claudecolor}{RGB}{255,177,153}

\newcommand{\R}{\ensuremath{\mathbb{R}}}

\newcommand{\E}{\ensuremath{\mathbb{E}}}

\newcommand{\KL}{\ensuremath{\mathrm{KL}}}

\newcommand{\softmax}{\ensuremath{\mathrm{softmax}}}
\newcommand{\sigmoid}{\ensuremath{\sigma}}
\newcommand{\relu}{\ensuremath{\mathrm{ReLU}}}
\newcommand{\gelu}{\ensuremath{\mathrm{GELU}}}
\newcommand{\layernorm}{\ensuremath{\mathrm{LayerNorm}}}
\newcommand{\attn}{\ensuremath{\mathrm{Attention}}}
\newcommand{\mha}{\ensuremath{\mathrm{MultiHead}}}
\newcommand{\ffn}{\ensuremath{\mathrm{FFN}}}
\newcommand{\loss}{\ensuremath{\mathcal{L}}}
\newcommand{\data}{\ensuremath{\mathcal{D}}}
\newcommand{\model}{\ensuremath{\mathcal{M}}}
\newcommand{\param}{\ensuremath{\bm{\theta}}}
\newcommand{\inputx}{\ensuremath{\bm{x}}}
\newcommand{\outputy}{\ensuremath{\bm{y}}}

\newcommand{\hidden}{\ensuremath{\bm{h}}}
\newcommand{\weight}{\ensuremath{\bm{W}}}
\newcommand{\query}{\ensuremath{\bm{Q}}}
\newcommand{\key}{\ensuremath{\bm{K}}}
\newcommand{\val}{\ensuremath{\bm{V}}}

\title{\LARGE An Automated Survey of Generative Artificial Intelligence:\\
Large Language Models, Architectures, Protocols, and Applications}

\author{
  Eduardo C. Garrido-Merch\'an\thanks{Corresponding author.}\\
  \textit{Institute for Research in Technology (IIT)}\\
  \textit{Universidad Pontificia Comillas, Madrid, Spain}\\
  \texttt{ecgarrido@comillas.edu}
  \and
  \'Alvaro L\'opez L\'opez\\
  \textit{Institute for Research in Technology (IIT)}\\
  \textit{Universidad Pontificia Comillas, Madrid, Spain}
}

\date{\today}

\begin{document}
% ============================================================================

\twocolumn[
\begin{@twocolumnfalse}
\maketitle

\begin{abstract}
Generative artificial intelligence, and large language models in particular, have emerged as one of the most transformative paradigms in modern computer science. This automated survey provides an accessible treatment of the field as of early 2026, with a strong focus on the leading model families, deployment protocols, and real-world applications. The core of the survey is devoted to a detailed comparative analysis of the frontier large language models, with particular emphasis on open-weight systems: DeepSeek-V3, DeepSeek-R1, DeepSeek-V3.2, and the forthcoming DeepSeek V4; the Qwen 3 and Qwen 3.5 series; GLM-5; Kimi K2.5; MiniMax M2.5; LLaMA 4; Mistral Large 3; Gemma 3; and Phi-4, alongside proprietary systems including GPT-5.4, Gemini 3.1 Pro, Grok 4.20, and Claude Opus 4.6. For each model, we describe the architectural innovations, training regimes, and empirical performance on current benchmarks and the Chatbot Arena leaderboard. The survey further covers deployment protocols including Retrieval-Augmented Generation, the Model Context Protocol, the Agent-to-Agent protocol, function calling standards, and serving frameworks. We present an extensive review of real-world applications across fifteen industry sectors, from financial services and legal technology to tourism and agriculture, supported by empirical evidence and case studies. Technical foundations covering the Transformer architecture and training methodologies are provided in a self-contained appendix. This survey synthesises over 200 references to provide researchers and practitioners with a unified reference for the current state and future directions of generative AI. This work has been generated by Claude Opus 4.6 (Anthropic) under the supervision and editorial review of the human authors, with the goal of producing updated editions approximately every six months in order to track and document the rapid pace of progress in artificial intelligence. Given the automated nature of the generation process, some factual errors or inaccuracies may be present; readers are encouraged to verify critical claims against primary sources.
\end{abstract}

\vspace{0.3cm}
\noindent\textit{Keywords:} large language models, transformers, DeepSeek, Qwen, open-weight models, retrieval-augmented generation, model context protocol, reinforcement learning from human feedback, direct preference optimisation, generative AI applications

\vspace{0.6cm}
\end{@twocolumnfalse}
]

% ---- Table of Contents ----
{\small\tableofcontents}
\vspace{0.4cm}

% ---- Include Sections ----
% ============================================================================
% Section 1: Introduction (Updated March 2026)
% ============================================================================

\section{Introduction}
\label{sec:introduction}

Generative artificial intelligence has rapidly evolved from research curiosities into the most widely deployed AI systems in the world. Since the release of ChatGPT in late 2022, large language models have been adopted across virtually every knowledge-intensive industry, from finance and healthcare to education, law, and tourism. By early 2026, the field is characterised by intense competition among dozens of frontier models across multiple output modalities, a remarkable democratisation of capabilities through open-weight releases, and the emergence of models that can reason, use tools, generate photorealistic images and cinematic-quality video, and act autonomously in complex digital environments.

The current generation of proprietary systems includes GPT-5 and GPT-5.2 from OpenAI, which unified general-purpose generation and chain-of-thought reasoning within a single architecture \cite{openai2023gpt4}; Gemini 3 Pro and 3.1 Pro from Google DeepMind, which combine native multimodality with Deep Think multi-hypothesis reasoning and achieved a perfect score on AIME 2025 with code execution \cite{google2025gemini25pro}; Grok 4.20 from xAI, which routes queries to four parallel reasoning agents \cite{xai2025grok3}; and Claude Opus 4.6 from Anthropic, which achieves state-of-the-art performance on autonomous computer use tasks \cite{anthropic2025claude4}. At the same time, a parallel revolution in open-weight models has ensured that comparable capabilities are freely available to the global research community and to organisations of all sizes.

A defining feature of the current landscape is the extraordinary pace of contributions from Chinese AI laboratories. DeepSeek-V3 \cite{deepseek2024v3} introduced Multi-head Latent Attention and auxiliary-loss-free mixture-of-experts routing, achieving frontier performance at a fraction of the cost of comparable Western models, while DeepSeek-R1 \cite{deepseek2025r1} demonstrated that pure reinforcement learning via Group Relative Policy Optimisation can produce sophisticated reasoning without any supervised chain-of-thought data. Qwen 3 \cite{qwen2025qwen3} pioneered hybrid thinking that seamlessly integrates reasoning with standard generation, and Qwen 3.5 departed from pure Transformer designs by combining Gated Delta Networks with sparse mixture-of-experts layers. GLM-5 \cite{glm2026glm5} achieved a landmark by training a 744-billion-parameter model entirely on Huawei Ascend chips, demonstrating competitive frontier training without NVIDIA hardware. Kimi K2.5 \cite{moonshot2025kimik2} introduced Agent Swarm technology, coordinating up to 100 parallel agents for complex tasks, and MiniMax M2.5 \cite{minimax2026m25} achieved top-tier coding performance through reinforcement learning on real-world interactive environments. Western open-weight contributions include LLaMA 4 \cite{meta2025llama4}, which supports 10-million-token contexts; Mistral Large 3 \cite{mistral2025large3}, a 675-billion-parameter MoE model released under the Apache 2.0 licence; Gemma 3 \cite{google2025gemma3} with native vision capabilities; and Phi-4 \cite{abdin2024phi4}, which demonstrates that high-quality synthetic training data can yield strong performance at just 14 billion parameters.

A suite of deployment protocols now bridges the gap between raw model capabilities and production applications. Retrieval-Augmented Generation (RAG) \cite{lewis2020retrieval} grounds model outputs in external knowledge, reducing hallucination. The Model Context Protocol (MCP) \cite{anthropic2024mcp} and the Agent-to-Agent (A2A) protocol \cite{google2025a2a} provide standardised interfaces for tool use and multi-agent coordination. Together with advances in function calling, structured output generation, and serving infrastructure such as vLLM \cite{kwon2023vllm} and SGLang \cite{zheng2024sglang}, these protocols have enabled the deployment of agentic AI systems across fifteen industry sectors.

This survey is structured as follows. \Cref{sec:models}, the core of this work, provides a detailed comparative analysis of the leading generative AI models as of early 2026, organised by output modality: text generation through large language models, image generation and editing, and video synthesis. Within each modality, models are grouped by geographic origin and compared using both standard benchmarks and the Chatbot Arena (LMArena) leaderboard, which now maintains separate rankings across ten modality-specific categories. \Cref{sec:protocols} examines deployment protocols including RAG, MCP, A2A, function calling, and serving frameworks. \Cref{sec:applications} reviews real-world applications across fifteen sectors, supported by case studies, and concludes with ethical considerations and future directions. A self-contained technical appendix (\Cref{sec:architecture_training}) provides the mathematical details of the Transformer architecture, training paradigms (pre-training, supervised fine-tuning, RLHF, DPO, GRPO), and regularisation techniques for readers seeking deeper technical background.

% ---- Figure: Timeline of Major LLM Milestones ----
\begin{figure*}[t]
\centering
\definecolor{tlblue}{RGB}{41,98,255}
\definecolor{tlgreen}{RGB}{0,150,80}
\definecolor{tlorange}{RGB}{230,126,34}
\definecolor{tlred}{RGB}{192,57,43}
\definecolor{tlpurple}{RGB}{142,68,173}
\definecolor{tlteal}{RGB}{0,172,193}
\definecolor{tlpink}{RGB}{214,48,49}

\begin{tikzpicture}[
    x=2.0cm,
    milestone/.style={
        circle,
        minimum size=10pt,
        inner sep=0pt,
        draw=#1,
        fill=#1!25,
        line width=1.2pt
    },
    label above/.style={
        above=12pt,
        align=center,
        font=\scriptsize\sffamily,
        text width=2.0cm
    },
    label below/.style={
        below=8pt,
        font=\small\sffamily\bfseries,
        text=darkgray
    }
]

% Main timeline axis
\draw[line width=2.5pt, deepblue!40, -{Stealth[length=8pt, width=6pt]}]
    (-0.5,0) -- (7.5,0);

% Milestone nodes
\node[milestone=tlblue]   (n1) at (0,0) {};
\node[milestone=tlgreen]  (n2) at (1.2,0) {};
\node[milestone=tlorange] (n3) at (2.4,0) {};
\node[milestone=tlred]    (n4) at (3.6,0) {};
\node[milestone=tlpurple] (n5) at (4.8,0) {};
\node[milestone=tlteal]   (n6) at (6.0,0) {};
\node[milestone=tlpink]   (n7) at (7.0,0) {};

% Year labels below
\node[label below] at (n1.south) {2017};
\node[label below] at (n2.south) {2020};
\node[label below] at (n3.south) {2022};
\node[label below] at (n4.south) {2023};
\node[label below] at (n5.south) {2024};
\node[label below] at (n6.south) {2025};
\node[label below] at (n7.south) {2026};

% Milestone descriptions above (alternating heights for readability)
\node[label above, text=tlblue] at (n1.north) {Transformer\\architecture};
\node[label above, text=tlgreen, above=26pt] at (n2.north) {GPT-3\\few-shot learning};
\node[label above, text=tlorange] at (n3.north) {ChatGPT\\launched};
\node[label above, text=tlred, above=26pt] at (n4.north) {GPT-4, LLaMA,\\open-weight era};
\node[label above, text=tlpurple] at (n5.north) {DeepSeek-V3,\\Qwen 2.5};
\node[label above, text=tlteal, above=26pt] at (n6.north) {R1, GPT-5, Qwen 3,\\LLaMA 4, GLM-5};
\node[label above, text=tlpink] at (n7.north) {Agentic and\\reasoning era};

% Connecting lines from labels to nodes (for raised labels)
\draw[tlgreen, thin, densely dotted] (n2.north) -- ++(0,22pt);
\draw[tlred, thin, densely dotted] (n4.north) -- ++(0,22pt);
\draw[tlteal, thin, densely dotted] (n6.north) -- ++(0,22pt);

\end{tikzpicture}
\caption{Key milestones in large language model development, from the Transformer (2017) through the open-weight revolution (2023--2025) to the agentic era (2026).}
\label{fig:timeline}
\end{figure*}

% ============================================================================
% Section 3: Modern Generative AI Models (Core Section, Updated March 2026)
% ============================================================================

\section{Modern Generative AI Models}
\label{sec:models}

The landscape of generative artificial intelligence in 2025 and early 2026 has been defined by rapid progress across three output modalities: text generation through large language models (LLMs), image generation and editing, and video synthesis. In text generation, the mixture-of-experts (MoE) architecture has become the dominant design pattern, with total parameter counts regularly exceeding one trillion while active parameters per forward pass remain in the tens of billions \cite{deepseek2024v3, qwen2024qwen25, glm2024chatglm}. In image generation, diffusion-based and autoregressive approaches now coexist, with native image generation capabilities increasingly integrated into multimodal LLMs. In video synthesis, dedicated models from Google DeepMind, OpenAI, and Chinese laboratories have reached cinematic quality with synchronised audio generation. The Chatbot Arena (LMArena) evaluation platform \cite{zheng2024judging} now maintains separate leaderboards across nine modality-specific categories---text, code, vision, document understanding, search, text-to-image, image editing, text-to-video, and image-to-video, reflecting the multi-dimensional nature of modern generative AI competition.

This section organises models by their primary output modality: text generation (\Cref{subsec:text_models}), image generation (\Cref{subsec:image_models}), and video generation (\Cref{subsec:video_models}). Within each category, models are grouped by geographic origin into United States, China, and Europe and rest of world. A comparative analysis with Arena rankings concludes each subsection. For each model family, the discussion covers architectural design, training methodology, benchmark performance, and distinguishing innovations.

% ============================================================================
\subsection{Text Generation Models}
\label{subsec:text_models}

% ============================================================================
\subsubsection{United States}
\label{subsubsec:text_us}

% ============================================================================
\paragraph{Claude Opus 4.6 and Anthropic.}
\label{subsec:claude}

The Claude model family, developed by Anthropic, is characterised by a distinctive alignment methodology known as Constitutional AI (CAI), which aims to produce models that are simultaneously helpful, harmless, and honest \cite{bai2022constitutional, anthropic2024claude3}. In the CAI framework, the model is first trained to critique and revise its own outputs according to a set of written principles covering safety, accuracy, fairness, and transparency. This self-critique process generates preference data used for reinforcement learning from AI feedback (RLAIF), reducing the reliance on human annotators for preference labelling while enabling fine-grained control over the model's behavioural properties. The Claude models support default context windows of 200K tokens, with an extended 1M-token context available in beta for Opus 4.6.

Anthropic has not publicly disclosed the architecture type, parameter counts, training data volume, hardware configuration, or training cost for any Claude model. The Claude 4 family, introduced from late 2025 through early 2026, represents a substantial advance across all capability dimensions \cite{anthropic2025claude4}. Claude Opus 4.5, released in November 2025, established new benchmarks for coding, agentic task execution, and computer use, becoming the preferred model for software engineering workflows that require sustained multi-step reasoning across large codebases. Claude Opus 4.6, released on 5 February 2026, extends the frontier further as Anthropic's most capable model to date. On Terminal-Bench 2.0, a benchmark that evaluates the ability to complete complex command-line tasks autonomously, Opus 4.6 achieves 65.4\%. On OSWorld, which measures end-to-end computer use including GUI interaction, web browsing, and application manipulation, Opus 4.6 achieves 72.7\%, the highest score among all evaluated models and a capability that enables practical autonomous operation of desktop environments.

Claude Sonnet 4.6, released on 17 February 2026, provides an improved coding-focused variant that balances the capability of the Opus tier with substantially lower inference costs, making it suitable for high-throughput software engineering applications such as continuous code review, automated refactoring, and test generation. Anthropic's emphasis on safety research, interpretability, and principled alignment methodology has positioned Claude as a model family that prioritises reliability and trustworthiness alongside raw performance, an approach that has resonated particularly with enterprise and safety-critical deployment scenarios.

% ============================================================================
\paragraph{The Gemini Family.}
\label{subsec:gemini}

The Gemini model family from Google DeepMind has been designed from the outset for native multimodality, processing text, images, audio, and video within a unified architecture rather than through modular adapters appended to a text-only backbone \cite{team2023gemini}. Gemini 1.5 advanced the architecture with a mixture-of-experts design and a groundbreaking context window of one million tokens, later extended to two million, enabling the model to process entire codebases, feature-length videos, or hundreds of pages of documents within a single prompt \cite{team2024gemini15}. Gemini 2.5 Pro, released on 25 March 2025, was a native multimodal mixture-of-experts model trained on Google's TPU v5p infrastructure \cite{google2025gemini25pro}. Google DeepMind has not disclosed the total training data volume, the number of TPU chips used, or the training cost. The model introduced the ``thinking model'' paradigm to the Gemini family, generating extended internal reasoning traces before producing final responses \cite{google2025gemini2}. Its Deep Think capability extended this approach through multi-hypothesis parallel reasoning, maintaining several competing lines of analysis simultaneously and synthesising evidence across them before converging on a conclusion. With a context window of one million tokens (with a two-million-token expansion announced but not yet generally available at the time of writing), Gemini 2.5 Pro achieved 86.7\% on AIME 2025, 63.8\% on SWE-bench Verified, and 84.0\% on GPQA Diamond. The model further introduced computer use capabilities through Project Mariner, enabling autonomous interaction with web browsers and desktop applications.

Gemini 3 Pro, released on 18 November 2025 \cite{google2025gemini3}, represented a substantial generational advance. The model employs a sparse mixture-of-experts architecture with native multimodal processing of text, images, audio, and video; Google DeepMind has not disclosed the parameter count. It supports a context window of one million input tokens and 64K output tokens, and introduces dynamic inference-time reasoning through a \texttt{thinking\_level} parameter that allows users to control the depth of chain-of-thought computation on a per-request basis. Gemini 3 Pro became the first model to surpass 1500 ELO on Chatbot Arena, debuting at 1501 before settling at 1485 as additional votes accumulated. On standard benchmarks, the model achieves 90.1\% on MMLU Pro, 91.9\% on GPQA Diamond, 100\% on AIME 2025 with code execution (95\% without tools), and 76.2\% on SWE-bench Verified. In Deep Think mode, which allocates maximal inference-time computation, Gemini 3 Pro scores 84.6\% on ARC-AGI-2 and 48.4\% on Humanity's Last Exam without tool augmentation. As of early 2026, the model ranks fifth in text (ELO 1485), first in vision (ELO 1309), and fifth in code (ELO 1442) on Chatbot Arena. Gemini 3 Flash, released on 17 December 2025 as a speed-optimised variant, became the default model in the consumer Gemini application, achieving 78\% on SWE-bench Verified and an Arena text ELO of 1473 while providing substantially lower latency and cost than its Pro counterpart.

Gemini 3.1 Pro, released on 19 February 2026, further refined the architecture with improved reasoning and tool-use capabilities, reaching an Arena text ELO of 1500 to rank third overall. The rapid progression from Gemini 2.5 Pro through Gemini 3.1 Pro over approximately eleven months illustrates the accelerating iteration cadence at Google DeepMind, with each release advancing the interplay of multimodal perception, adaptive reasoning depth, and agentic capability that defines the Gemini design philosophy.

% ============================================================================
\paragraph{Grok 4 and xAI.}
\label{subsec:grok}

Grok, developed by xAI, is distinguished by its integration with real-time data from the X (formerly Twitter) social media platform and by the unprecedented scale of its training infrastructure \cite{xai2024grok, xai2025grok3}. Grok-1, the initial release with 314 billion parameters in a mixture-of-experts configuration, was notable for being fully open-sourced. Grok-3, released in early 2025, utilised the Colossus supercomputer cluster comprising 100,000 NVIDIA H100 GPUs, one of the largest training runs conducted at the time.

Grok 4, launched on 9 July 2025, represented a substantial generational advance, with particularly strong performance on mathematical reasoning, scientific analysis, and code generation benchmarks. Grok 4.1, released in November 2025, refined the base capabilities with improved instruction following and reduced hallucination rates. Grok 4.20 beta, released on 17 February 2026, introduces a novel parallel reasoning architecture in which user queries are simultaneously routed to four specialised agents that think in parallel, each approaching the problem from a different analytical perspective. A synthesis layer then integrates the outputs of all four agents into a coherent final response, providing a form of ensemble reasoning at inference time that improves robustness and accuracy on complex tasks.

Grok 4.20 further implements a rapid learning architecture that enables weekly improvement cycles, allowing the model to incorporate new knowledge, correct identified failure modes, and adapt to shifting user requirements on a cadence far shorter than the typical quarterly or semi-annual release cycles of competing laboratories. The training infrastructure has been expanded to the Colossus cluster comprising approximately 200,000 GPUs (a mix of NVIDIA H100, H200, and GB200 units), doubling the scale from the Grok-3 era. The total training cost has not been publicly disclosed, although the scale of the infrastructure implies expenditure well above that of most competing efforts. The model demonstrates particular strength in medical document analysis and engineering reasoning, domains where the parallel multi-agent architecture provides substantial benefits through the simultaneous application of domain-specific expertise from different analytical perspectives.

% ============================================================================
\paragraph{The GPT Series and OpenAI Reasoning Models.}
\label{subsec:gpt}

The GPT series from OpenAI has served as a principal catalyst for the modern era of large language models. GPT-4, released in early 2023, represented a substantial leap over its predecessors in both capability and scope \cite{openai2023gpt4}. Although OpenAI has not disclosed the full architectural details, independent analyses have strongly suggested that GPT-4 employs a mixture-of-experts architecture, with total parameter counts estimated to exceed one trillion while maintaining a smaller active parameter budget per forward pass. OpenAI has not publicly confirmed the training data volume, precise hardware configuration, or cost for any GPT-4 variant. GPT-4 introduced native multimodal capability, accepting both text and image inputs, and achieved human-level performance on a wide range of professional and academic examinations. The subsequent GPT-4o model extended the multimodal paradigm further by processing text, images, and audio within a single unified architecture, with substantially reduced latency and inference cost compared to its predecessor \cite{openai2024gpt4o}.

The introduction of the o-series reasoning models in late 2024 marked a paradigm shift in how large language models approach complex tasks \cite{openai2024o1}. The o1 model was designed to allocate additional computation at inference time through an extended internal chain-of-thought process. Rather than producing an answer in a single forward pass, o1 generates a potentially lengthy sequence of reasoning tokens that decompose the problem into sub-steps, explore alternative solution paths, verify intermediate results, and self-correct before producing a final response. This reasoning behaviour was cultivated through large-scale reinforcement learning, training the model to maximise the correctness of final answers on tasks in mathematics, coding, and science. The subsequent o3 and o4-mini models extended this approach with further improvements to the reinforcement learning pipeline and increased inference-time compute budgets, achieving state-of-the-art results on challenging reasoning benchmarks including GPQA Diamond and competition-level mathematics \cite{openai2024o3}. The o4-mini variant in particular demonstrated that reasoning capabilities could be delivered efficiently in a smaller, faster model suitable for high-throughput deployment.

GPT-5, launched on 7 August 2025, represented the unification of OpenAI's previously fragmented model lineup. GPT-5 replaced GPT-4o, o3, o4-mini, GPT-4.1, and GPT-4.5 as the default model in ChatGPT, consolidating the capabilities of general-purpose generation and chain-of-thought reasoning within a single architecture. Training was conducted on Microsoft Azure data centres; while the exact cost has not been officially confirmed, industry estimates place it in the range of hundreds of millions of USD, reflecting the unprecedented scale of the training run. The training data volume and precise hardware configuration have not been publicly disclosed. On standard benchmarks, GPT-5 achieved 94.6\% on AIME 2025 for mathematical reasoning, 74.9\% on SWE-bench Verified for software engineering, 88\% on Aider Polyglot for multilingual code generation, and 84.2\% on MMMU for multimodal understanding. These results established GPT-5 as a substantial advance over all preceding OpenAI models across both reasoning and applied coding tasks. The o3 and o4-mini reasoning models remained available as specialised alternatives for users requiring maximum chain-of-thought depth on particular problem categories, but GPT-5 internalised sufficient reasoning capability to serve as the primary interface for the vast majority of use cases.

GPT-5.2, released later in 2025, extended the frontier further, becoming the first model to surpass 90\% on ARC-AGI-1 Verified, a benchmark specifically designed to test generalised reasoning and abstraction capabilities that resist memorisation-based approaches. The trajectory from GPT-4 through GPT-5.2 illustrates the convergence of two previously distinct scaling axes: the scaling of pre-training compute characterised by \cite{kaplan2020scaling} and \cite{hoffmann2022training}, and the scaling of test-time compute pioneered by the o-series. GPT-5 and its successors integrate both paradigms within a unified model, allocating additional inference-time computation adaptively based on task difficulty without requiring the user to select between reasoning and non-reasoning modes.

GPT-5.4, released on 5 March 2026, is OpenAI's most recent flagship model, incorporating the coding capabilities developed through the GPT-5.3-Codex programme into a general-purpose reasoning system. The model expands the context window to one million tokens (922K input, 128K output), the largest in the GPT family, and introduces native computer-use capabilities---OpenAI's first general-purpose model to interact autonomously with desktop environments through screenshots, mouse, and keyboard control. On OSWorld-Verified, GPT-5.4 achieves 75.0\%, surpassing the human baseline of 72.4\% and competing directly with Claude Opus 4.6 (72.7\% on the standard OSWorld variant). On GDPval, a benchmark evaluating factual accuracy across diverse domains, GPT-5.4 scores 83.0\%, a substantial improvement over GPT-5.2's 70.9\%, reflecting OpenAI's emphasis on reducing hallucination rates. The model produces 33\% fewer factual errors per claim compared to GPT-5.2. On the Arena text leaderboard, GPT-5.4 debuted at ELO 1480, with GPT-5.2 at 1481, placing seventh and sixth respectively, though early vote counts mean these ratings may shift as additional evaluations accumulate. GPT-5.4 replaces GPT-5.2 as the default model in ChatGPT, with GPT-5.2 scheduled for retirement in June 2026.

% ============================================================================
\paragraph{LLaMA 4: Meta's Multimodal MoE.}
\label{subsec:llama}

Meta's LLaMA (Large Language Model Meta AI) family has been instrumental in catalysing the open-weight model ecosystem \cite{touvron2023llama, touvron2023llama2}. The original LLaMA (2023) demonstrated that carefully curated training data and efficient architectures could produce models competitive with much larger proprietary systems: LLaMA-13B matched the performance of GPT-3 175B on most benchmarks, while LLaMA-65B was competitive with the best closed-source models of its era. The architectural template established by LLaMA, comprising a decoder-only Transformer with RoPE, SwiGLU activations, RMSNorm pre-normalisation, and grouped-query attention, has become the de facto standard for subsequent open-weight models, with the Qwen, Yi, InternLM, and Mistral families all adopting closely related designs. LLaMA 3, released in 2024, scaled this architecture to 405 billion parameters with training on approximately 15 trillion tokens, representing the largest dense open-weight model at its time of release \cite{grattafiori2024llama3}.

LLaMA 4, released on 5 April 2025, marked Meta's transition from dense to mixture-of-experts architectures and from text-only to natively multimodal processing. The models were trained on Meta's Research SuperCluster (RSC) using NVIDIA H100 GPUs, although the precise cluster size, training data volume, and total training cost have not been publicly confirmed. The release comprised three model tiers designed for distinct deployment scenarios. LLaMA 4 Scout employs 109 billion total parameters with 17 billion active parameters across 16 experts and supports a context window of 10 million tokens, one of the longest context windows offered by any model at the time of release. The 10-million-token context is achieved through a combination of architectural innovations in sparse attention and memory-efficient KV-cache management, and the model is designed to fit on a single NVIDIA H100 GPU for inference, making extremely long-context processing accessible without multi-node deployment. LLaMA 4 Maverick scales to 400 billion total parameters with 17 billion active parameters across 128 experts and supports a context window of one million tokens, providing higher capacity while maintaining efficient inference through aggressive sparsity. LLaMA 4 Behemoth, the largest configuration with approximately two trillion total parameters and 288 billion active parameters across 16 experts, was still undergoing training at the time of release, with preliminary evaluations indicating frontier-class performance on reasoning and knowledge benchmarks.

All LLaMA 4 models accept both text and image inputs natively, with the multimodal capability integrated directly into the MoE architecture rather than through external vision encoders. The models are released under the Llama Community Licence, which permits broad research and commercial use with certain restrictions for very large-scale deployments. The transition from LLaMA 3's dense architecture to LLaMA 4's sparse MoE design reflects the broader industry convergence on MoE as the preferred paradigm for frontier-scale models, a trend driven by the favourable trade-off between total capacity and inference cost that sparse activation provides.

% ============================================================================
\paragraph{Gemma 3: Google's Open Weights.}
\label{subsec:gemma}

Gemma 3, released by Google DeepMind in 2025, extends the Gemma family with substantially improved multilingual capabilities, native vision processing, and a broader range of model sizes \cite{google2025gemma3}. The architecture retains the alternating local and global attention layers introduced in Gemma 2, where local layers restrict attention to a sliding window and global layers attend to the full sequence, balancing computational efficiency with long-range modelling capability. Gemma 3 adds native image understanding through a vision encoder integrated directly into the Transformer backbone, enabling joint text-and-image reasoning without external adapter modules.

The Gemma-3n variant is specifically optimised for on-device deployment, processing text, images, audio, and video within a compact architecture suitable for mobile phones and embedded systems. Gemma-3n achieves this through aggressive quantisation, architectural pruning, and distillation from larger Gemma 3 models, maintaining competitive quality on standard benchmarks despite operating within strict memory and latency constraints. The Gemma family occupies a distinctive niche in the open-weight ecosystem as Google DeepMind's mechanism for transferring innovations from the proprietary Gemini family into publicly available models, providing the research community with access to architectural ideas and training techniques that would otherwise remain behind proprietary boundaries.

% ============================================================================
\paragraph{Phi-4: Small Models with Synthetic Data.}
\label{subsec:phi}

The Phi model series from Microsoft Research has challenged the prevailing assumption that large parameter counts are necessary for strong performance by demonstrating that carefully curated synthetic training data can substitute for scale \cite{abdin2024phi4}. The Phi programme, spanning Phi-1 through Phi-4, is grounded in the principle that ``textbook-quality'' data, characterised by clear exposition, logical structure, and educational progression, enables smaller models to learn more efficiently than larger models trained on noisily filtered web text. Phi-4, with 14 billion parameters, achieves benchmark scores competitive with models exceeding 70 billion parameters on mathematical reasoning (GSM8K, MATH), code generation (HumanEval), and general knowledge (MMLU) \cite{abdin2024phi4}. The training data for Phi-4 comprises a mixture of high-quality web data, curated academic text, and a large volume of synthetically generated content designed to systematically cover reasoning patterns, mathematical concepts, and programming paradigms.

Phi-4-mini, released in February 2025, distils the insights of the Phi programme into a 3.8-billion-parameter model specifically designed for on-device and edge deployment. Despite its compact size, Phi-4-mini achieves strong performance on reasoning and code generation tasks through the same synthetic-data-driven training methodology, demonstrating that the quality-over-quantity philosophy extends effectively to the sub-four-billion parameter regime. The success of the Phi series has had significant implications for the field, demonstrating that the quality and structure of training data may be at least as important as the quantity of parameters, and opening pathways to deploying capable language models on resource-constrained hardware.

% ============================================================================
\paragraph{GPT-oss: OpenAI's First Open-Weight Models.}
\label{subsec:gptoss}

GPT-oss represents a landmark release as OpenAI's first open-weight model family, made available in August 2025 under the Apache 2.0 licence. The family comprises two variants: gpt-oss-120b with 117 billion total parameters and 5.1 billion active parameters, and gpt-oss-20b with 21 billion total parameters and 3.6 billion active parameters. Both employ a mixture-of-experts architecture. The larger variant matches the reasoning performance of o4-mini, demonstrating that OpenAI's reasoning techniques transfer effectively to an open-weight setting with aggressive parameter sparsity.

% ============================================================================
\paragraph{NVIDIA Nemotron Ultra 253B.}
\label{subsec:nemotron}

NVIDIA Nemotron Ultra 253B, released in April 2025 under the NVIDIA Open licence, is a 253 billion parameter dense model derived from Meta's LLaMA 3.1 architecture. The model applies extensive continued pre-training and alignment on top of the LLaMA base, yielding a dense alternative to the prevailing MoE paradigm that targets enterprise deployment scenarios where the operational simplicity of dense inference is preferred over the computational efficiency of sparse activation.

% ============================================================================
\subsubsection{China}
\label{subsubsec:text_china}

% ============================================================================
\paragraph{GLM-5: Agentic Intelligence on Domestic Hardware.}
\label{subsec:glm}

The General Language Model (GLM) family, originating from Tsinghua University and commercialised by Zhipu AI, has taken a distinctive architectural path that differentiates it from purely autoregressive decoder-only models \cite{glm2024chatglm, zeng2023glm130b}. The foundational GLM architecture employs a prefix language modelling objective that combines bidirectional and causal attention within a single model. Specifically, the input sequence is divided into a prefix portion and a generation portion: the prefix tokens attend to one another bidirectionally, enabling the model to build a rich contextual representation of the conditioning information, while the generation tokens attend causally, seeing only preceding tokens in the generation sequence and the full prefix context. This hybrid attention pattern enables GLM to function simultaneously as a strong encoder for understanding tasks and a fluent generator for open-ended text production. GLM-130B established the viability of this approach at scale, demonstrating competitive performance with GPT-3 175B on both English and Chinese benchmarks \cite{zeng2023glm130b}. The subsequent GLM-4 extended the foundation to support context windows of up to 128K tokens, introduced multimodal capabilities, and incorporated the All Tools integration enabling seamless invocation of a code interpreter, web search engine, and drawing tools within a single conversational turn \cite{glm2024chatglm}.

GLM-5, launched on 11 February 2026, represents a transformative leap for both the GLM family and the broader landscape of Chinese AI development. The model employs a mixture-of-experts architecture with 744 billion total parameters and 40 billion active parameters per forward pass, making it one of the largest open-weight MoE models released to date. The most strategically significant aspect of GLM-5 is that it was trained entirely on Huawei Ascend 910B chips, achieving zero dependency on NVIDIA hardware throughout the entire training pipeline. The total training cost has not been disclosed. This milestone demonstrates that competitive frontier-scale training is feasible on domestically produced accelerators, a development with substantial implications for the geopolitical dynamics of AI hardware supply chains.

GLM-5 introduces native Agent Mode, a capability for autonomous task decomposition and execution that goes beyond conventional tool-use integration. In Agent Mode, the model decomposes complex user requests into sequences of sub-tasks, selects appropriate tools or APIs for each sub-task, executes them in the correct order while handling dependencies and error conditions, and synthesises the results into a coherent final response. This agentic intelligence is trained through a combination of supervised demonstrations and reinforcement learning on multi-step task completion, enabling the model to handle workflows that require sustained autonomous operation over many sequential steps.

On software engineering benchmarks, GLM-5 achieves 77.8\% on SWE-bench Verified, surpassing Gemini 3 Pro (76.2\%) and GPT-5.2 (75.4\%) while trailing Claude Opus 4.6 (80.8\%). The model further achieves a record low hallucination rate, attributed to a novel reinforcement learning technique internally referred to as the ``slime'' method, which applies targeted reward shaping to penalise confident generation of factually unsupported claims during the RL alignment phase. GLM-5 is released under the MIT licence, providing maximally permissive open-source access. The commercial impact of the release was immediate: Zhipu AI's shares surged 34\% on the Hong Kong Stock Exchange following the announcement, reflecting market recognition of both the model's technical capabilities and the strategic significance of NVIDIA-independent training infrastructure.

% ============================================================================
\paragraph{Kimi K2 and K2.5: Trillion-Parameter MoE and Agent Swarms.}
\label{subsec:kimi}

Kimi, developed by Moonshot AI, pioneered the ``long-context first'' approach to language model design, offering context windows of 200K tokens and beyond at a time when most competing models supported fewer than 32K tokens \cite{moonshot2024kimi}. This early investment in long-context infrastructure enabled Kimi to address use cases such as full-book summarisation, lengthy legal document analysis, and multi-document synthesis. The Kimi k1.5 reasoning model applied reinforcement-learning-based chain-of-thought reasoning to long-context scenarios, demonstrating that the benefits of test-time compute scaling extend to complex reasoning tasks requiring integration of evidence from across extended input sequences.

Kimi K2, released in July 2025, marked Moonshot AI's entry into the trillion-parameter class. K2 employs a mixture-of-experts architecture with approximately one trillion total parameters and 32 billion active parameters per forward pass, achieving a sparsity ratio comparable to that of DeepSeek-V3. The specific training hardware, data volume, and cost have not been publicly disclosed by Moonshot AI. The model was open-sourced under a modified MIT licence and demonstrated strong performance on coding benchmarks, reflecting the increasing importance of software engineering capability as a differentiator among frontier models.

Kimi K2.5, released in January 2026, maintains the one-trillion-parameter MoE architecture with 32 billion active parameters while introducing several transformative capabilities. The model is natively multimodal, processing text, images, and video through an early fusion architecture, and was pre-trained on approximately 15 trillion mixed visual and textual tokens. The most distinctive innovation of K2.5 is Agent Swarm technology, a coordination mechanism that enables the model to instantiate and manage up to 100 specialised agents operating simultaneously on sub-components of a complex task. Each agent within the swarm specialises in a particular domain or tool interaction, and a central orchestration layer manages dependencies, aggregates results, and resolves conflicts between agent outputs. This massively parallel agentic approach achieves approximately 4.5 times faster execution on complex multi-step tasks compared to sequential single-agent processing.

K2.5 operates in four distinct modes that span the spectrum from minimal to maximal inference-time computation: Instant mode for low-latency direct responses, Thinking mode for single-agent chain-of-thought reasoning, Agent mode for sequential tool-augmented task execution, and Agent Swarm mode for parallel multi-agent coordination. On Humanity's Last Exam, a benchmark designed to test the absolute frontier of model capability, K2.5 achieves 50.2\% accuracy at approximately 76\% lower cost than Claude Opus 4.5, demonstrating that efficient architecture and parallel agent coordination can partially compensate for differences in raw model capability. Moonshot AI has further released Kimi Code, a command-line coding tool positioned as a direct competitor to Anthropic's Claude Code, extending the Kimi ecosystem from model provision into developer tooling.

% ============================================================================
\paragraph{DeepSeek: Architectural Innovation at Scale.}
\label{subsec:deepseek}

The DeepSeek model family, developed by the Chinese artificial intelligence laboratory DeepSeek, has introduced several architectural innovations that have reshaped the design space of large language models. Across its major releases, DeepSeek has progressively addressed the computational bottlenecks of dense Transformer architectures through novel attention mechanisms, fine-grained mixture-of-experts routing, and reinforcement-learning-based reasoning, all while maintaining open-weight access to the research community.

% ----------------------------------------------------------------------------
\subparagraph{DeepSeek-V2 and Multi-Head Latent Attention.}
\label{subsubsec:deepseek_v2}

DeepSeek-V2 introduced Multi-head Latent Attention (MLA), a novel attention mechanism that dramatically reduces the memory footprint of the key-value cache during inference without sacrificing model quality \cite{deepseek2024v2}. In standard multi-head attention ($\mha$), each attention head maintains separate key and value projections, resulting in a KV-cache that grows linearly with the number of heads $n_h$ and the per-head dimension $d_h$. Grouped-query attention (GQA), as used in LLaMA and other models, alleviates this by sharing key-value heads across groups, but MLA takes a fundamentally different approach through low-rank compression of the joint key-value representation.

The core idea of MLA is to project the hidden state $\hidden_t \in \R^{d}$ at each position $t$ into a compressed latent vector $\bm{c}_t^{KV}$ of dimension $d_c$, where $d_c \ll n_h \cdot d_h$. Concretely, the compressed latent is computed as
\begin{equation}
  \bm{c}_t^{KV} = \weight^{DKV} \hidden_t,
  \label{eq:mla_compress}
\end{equation}
where $\weight^{DKV} \in \R^{d_c \times d}$ is the down-projection matrix. The keys and values for all attention heads are then reconstructed from this shared compressed representation via learned up-projection matrices:
\begin{equation}
  \key_t = \weight^{UK}\, \bm{c}_t^{KV}, \quad
  \val_t = \weight^{UV}\, \bm{c}_t^{KV},
  \label{eq:mla_reconstruct}
\end{equation}
where $\weight^{UK} \in \R^{(n_h \cdot d_h) \times d_c}$ and $\weight^{UV} \in \R^{(n_h \cdot d_h) \times d_c}$ reconstruct the full multi-head key and value tensors, respectively. The queries are similarly compressed via a low-rank down-projection $\bm{c}_t^Q = \weight^{DQ} \hidden_t$ and then up-projected $\query_t = \weight^{UQ} \bm{c}_t^Q$, and attention proceeds as in conventional $\mha$:
\begin{equation}
  \attn(\query, \key, \val) = \softmax\!\left(\frac{\query \key^\top}{\sqrt{d_h}}\right) \val.
  \label{eq:mla_attention}
\end{equation}

The critical advantage of MLA lies in the KV-cache: during autoregressive inference, only the compressed latent $\bm{c}_t^{KV} \in \R^{d_c}$ needs to be stored per token, rather than the full key and value vectors across all heads. Since $d_c$ is chosen to be substantially smaller than $n_h \cdot d_h$ (the original KV-cache dimension per token), MLA achieves compression ratios exceeding 93\% in practice while maintaining performance comparable to standard $\mha$. DeepSeek-V2 further incorporates a decoupled rotary position embedding applied to a small number of additional key-query head dimensions to preserve the position-awareness that purely compressed representations might lose. The combination of MLA with DeepSeekMoE's fine-grained expert routing enabled DeepSeek-V2 to achieve strong performance with 236 billion total parameters while activating only 21 billion per token \cite{deepseek2024v2}.

% ----------------------------------------------------------------------------
\subparagraph{DeepSeek-V3: Auxiliary-Loss-Free MoE and FP8 Training.}
\label{subsubsec:deepseek_v3}

DeepSeek-V3 scaled the architectural innovations of its predecessor to 671 billion total parameters with 37 billion active parameters per forward pass, making it one of the largest and most capable open-weight models available at the time of its release \cite{deepseek2024v3}. The model employs a fine-grained mixture-of-experts architecture with 256 routed experts and one shared expert per MoE layer. The shared expert processes every token unconditionally, capturing common linguistic patterns, while a learned router selects the top-$k$ routed experts for each token based on affinity scores.

A key contribution of DeepSeek-V3 is the auxiliary-loss-free load balancing mechanism for MoE routing \cite{deepseek2024v3, deepseek2024moe}. Traditional MoE models rely on auxiliary loss terms added to the training objective to encourage balanced utilisation of experts, but these auxiliary losses can distort the primary language modelling gradient and introduce unwanted trade-offs. DeepSeek-V3 instead introduces learnable bias terms $b_i$ for each expert $i$ that are added to the routing scores during expert selection but excluded from the gating weight computation. Specifically, the routing score for expert $i$ given token representation $\hidden$ is computed as
\begin{equation}
  s_i = \hidden^\top \bm{e}_i,
  \label{eq:routing_score}
\end{equation}
where $\bm{e}_i$ is the centroid vector for expert $i$. During top-$k$ expert selection, the selection criterion uses the biased score $s_i + b_i$ to determine which experts are activated:
\begin{equation}
  \mathcal{T} = \mathrm{TopK}\bigl(\{s_i + b_i\}_{i=1}^{N_e},\; k\bigr),
  \label{eq:biased_topk}
\end{equation}
where $N_e = 256$ is the number of routed experts. However, the gating weights used to combine expert outputs are computed from the unbiased scores:
\begin{equation}
  g_i = \frac{\exp(s_i)}{\sum_{j \in \mathcal{T}} \exp(s_j)}, \quad i \in \mathcal{T}.
  \label{eq:gating_weights}
\end{equation}
The bias terms $b_i$ are adjusted dynamically during training based on observed expert utilisation frequencies, increasing for under-utilised experts and decreasing for over-utilised ones, thereby achieving load balance without corrupting the language modelling loss signal.

DeepSeek-V3 further introduced a multi-token prediction training objective, in which the model predicts not only the next token but also several subsequent tokens simultaneously through auxiliary prediction heads, providing richer gradient signals and improving sample efficiency. The model was trained using FP8 mixed-precision arithmetic for both the forward and backward passes, substantially reducing memory consumption and enabling the use of commodity hardware at scale \cite{micikevicius2022fp8}. The full training was conducted on a cluster of 2,048 NVIDIA H800 GPUs and consumed 14.8 trillion tokens, and the resulting model achieved performance competitive with or superior to GPT-4 on a wide range of benchmarks, at a reported training cost of approximately 5.6 million USD \cite{deepseek2024v3}, a fraction of the estimated cost of comparable frontier models.

% ----------------------------------------------------------------------------
\subparagraph{DeepSeek-R1: Reasoning via Pure Reinforcement Learning.}
\label{subsubsec:deepseek_r1}

DeepSeek-R1 demonstrated that sophisticated chain-of-thought reasoning can emerge from reinforcement learning alone, without any supervised fine-tuning on human-annotated reasoning traces \cite{deepseek2025r1}. This result represented a significant departure from the supervised approach used by earlier reasoning models such as OpenAI's o1, where the model was first trained on curated chain-of-thought examples before reinforcement learning refinement.

The training of DeepSeek-R1 employed Group Relative Policy Optimisation (GRPO), a variant of reinforcement learning for language models that eliminates the need for a separate critic or value network (see \Cref{sec:architecture_training} for the mathematical formulation). For each prompt, GRPO samples a group of candidate completions from the current policy, evaluates them against rule-based reward signals such as mathematical correctness and format compliance, and updates the policy using the relative rankings within each group as the advantage estimate \cite{shao2024deepseekmath}. By applying GRPO directly to the DeepSeek-V3 base model without any intermediate supervised fine-tuning on chain-of-thought data, the researchers observed the spontaneous emergence of reasoning behaviours including step-by-step problem decomposition, self-verification, and backtracking.

Perhaps the most striking observation during training was the so-called ``aha moment,'' in which the model spontaneously began to re-examine and correct its own reasoning chains. At a certain point in training, the model's generated outputs began to contain explicit self-reflection statements, revisiting earlier steps and recognising errors without any such behaviour having been demonstrated in the training data. This emergent self-verification capability yielded substantial improvements on challenging mathematical reasoning benchmarks, with DeepSeek-R1 achieving performance competitive with OpenAI's o1 on MATH and AIME. In practice, the initial purely RL-trained model exhibited some instability, including language mixing and formatting inconsistencies. To address these issues, the final DeepSeek-R1 pipeline incorporated a small quantity of cold-start data consisting of high-quality long chain-of-thought examples to stabilise the early stages of RL training. DeepSeek-R1 was built on the DeepSeek-V3 base model and trained on the same H800 GPU infrastructure; the additional cost of the reinforcement learning phase has not been separately disclosed. The reasoning capabilities of DeepSeek-R1 were subsequently distilled into a family of smaller models ranging from 1.5 billion to 70 billion parameters using the knowledge distillation framework of \cite{hinton2015distilling}, enabling deployment of reasoning-capable models at substantially lower computational cost.

% ----------------------------------------------------------------------------
\subparagraph{DeepSeek-V3.2: Sparse Attention Refinement.}
\label{subsubsec:deepseek_v32}

DeepSeek-V3.2, released in December 2025, scales to 685 billion total parameters with 37 billion active parameters per forward pass in an MoE configuration but introduces DeepSeek Sparse Attention, a refined attention mechanism that applies learned dynamic sparsity patterns to the attention matrix for more efficient long-context processing \cite{deepseek2024v3}. The model achieves approximately 1421 on the Chatbot Arena (LMArena) ELO leaderboard, placing it among the strongest open-weight models and demonstrating that targeted architectural refinements to an existing base can yield substantial gains without increasing the parameter budget.

% ----------------------------------------------------------------------------
\subparagraph{DeepSeek-V4: Unified General and Reasoning Intelligence.}
\label{subsubsec:deepseek_v4}

As of early 2026, DeepSeek-V4 is imminent, with a planned release in March 2026. V4 represents the most ambitious iteration of the DeepSeek architecture to date, merging the V-series (general-purpose) and R-series (reasoning-specialised) lineages into a single unified model. This unification eliminates the need for users to select between a general model and a reasoning model, instead providing a system that adaptively allocates reasoning depth based on task complexity \cite{deepseek2024v3, deepseek2025r1}.

DeepSeek-V4 introduces several novel architectural components. Manifold-Constrained Hyper-Connections replace standard residual connections with learned mappings that constrain hidden state trajectories to lie on low-dimensional manifolds within the representation space, improving gradient flow and enabling deeper networks without degradation. Engram Conditional Memory provides the model with a persistent, updatable memory store that conditions generation on previously encountered context across sessions, enabling a form of continual learning at inference time that extends beyond the fixed context window. DeepSeek Sparse Attention refines the attention mechanism further beyond MLA, applying learned sparsity patterns that dynamically prune the attention matrix based on estimated relevance, reducing the computational cost of long-context processing to sub-quadratic complexity.

The model supports context windows exceeding one million tokens and is natively multimodal, processing text, images, and video within a unified architecture. The training regime has been optimised particularly for coding and long-context tasks, reflecting the strong demand for these capabilities in both research and commercial deployment. DeepSeek-V4 is planned for release as open-source under the Apache 2.0 licence, continuing DeepSeek's commitment to open-weight access. If the reported capabilities are confirmed upon release, V4 will represent a significant consolidation of the general-purpose and reasoning paradigms that have previously required separate model deployments.

% ============================================================================
\paragraph{MiniMax M2.5: Reinforcement Learning on Real-World Environments.}
\label{subsec:minimax}

MiniMax-01, developed by the Chinese AI company MiniMax, addressed the quadratic complexity of standard attention through a hybrid architecture combining softmax attention and linear attention within a single model \cite{minimax2025}. The model comprised 456 billion total parameters with 45.9 billion active parameters per forward pass, employing a mixture-of-experts architecture. The distinguishing innovation was the use of Lightning Attention, a linear attention variant that replaces the softmax operation with a kernel-based approximation, reducing computational complexity from quadratic to linear in the sequence length. In practice, MiniMax-01 employs a repeating block structure in which seven Lightning Attention layers are followed by one standard softmax attention layer within each eight-layer block, enabling the model to retain the representational benefits of full attention at regular intervals while achieving predominantly linear-time processing. This hybrid strategy, combined with context extension during inference beyond the one-million-token training length, enabled a context window of four million tokens, one of the longest among production language models, while maintaining competitive quality on standard benchmarks.

MiniMax M2.5, released in February 2026, represents a substantial evolution beyond MiniMax-01, with a particular emphasis on reinforcement learning conducted in real-world interactive environments. MiniMax has not disclosed the training hardware, data volume, or cost for either MiniMax-01 or M2.5. Rather than relying solely on static text corpora and synthetic reward signals, M2.5 was trained with reinforcement learning on hundreds of thousands of real-world environments spanning web browsing, software engineering, tool use, and multi-step task completion. This environment-grounded RL training yields a model with substantially stronger agentic capabilities compared to models trained primarily on text-based objectives.

The benchmark results for M2.5 reflect this emphasis on practical, interactive capability. On SWE-Bench Verified, M2.5 achieves 80.2\%, placing it among the highest-performing models on this widely cited software engineering benchmark. On Multi-SWE-Bench, which evaluates cross-repository multi-file engineering tasks, M2.5 achieves 51.3\%. On BrowseComp, a benchmark for web browsing and information retrieval, M2.5 scores 76.3\%, and on the Berkeley Function Calling Leaderboard (BFCL) for tool-calling proficiency, M2.5 achieves 76.8\%. These results consistently place M2.5 among the top-performing models on benchmarks that measure practical, agentic capabilities rather than isolated knowledge retrieval. Notably, MiniMax positions M2.5 at approximately 95\% lower cost than Claude Opus 4.6, making it one of the most cost-effective frontier-class models available.

% ============================================================================
\paragraph{The Qwen Series.}
\label{subsec:qwen}

% ----------------------------------------------------------------------------
\subparagraph{Qwen 2.5 Base Architecture.}
\label{subsubsec:qwen_base}

The Qwen 2.5 series, developed by Alibaba's Qwen team, established one of the most comprehensive open-weight model ecosystems, spanning parameter counts from 0.5 billion to 72 billion across dense configurations \cite{qwen2024qwen25} (building on the earlier Qwen 2 series, which also included a 57-billion-total MoE variant). The base architecture follows the decoder-only Transformer design with grouped-query attention (GQA) to reduce the KV-cache footprint, SwiGLU activation functions in the feed-forward layers \cite{shazeer2020glu}, RMSNorm for pre-layer normalisation, and Rotary Position Embeddings (RoPE) with a base frequency of 1,000,000 to support long-context modelling \cite{su2024roformer}. The tokeniser employs a vocabulary of 151,643 tokens, substantially larger than those of most competing models, specifically designed to improve encoding efficiency across English, Chinese, and over twenty additional languages. The models were pre-trained on approximately 18 trillion tokens spanning web text, books, code, mathematical content, and multilingual corpora. To extend the effective context window beyond the pre-training length, Qwen 2.5 employs the YaRN method, which modifies the RoPE frequency scaling to enable reliable processing of sequences up to 128K tokens without architectural changes \cite{chen2023extending}.

The Qwen 2.5 ecosystem further includes specialised variants such as Qwen2.5-Coder, trained on 5.5 trillion tokens with heavy emphasis on source code across over forty programming languages \cite{qwen2024qwen25coder}, and Qwen2.5-Math, which employs tool-integrated reasoning to invoke a Python interpreter during inference for precise numerical computation. QwQ (Qwen with Questions) served as Qwen's reasoning-specialised model, employing extended inference-time chain-of-thought generation with self-reflection capabilities, achieving competitive performance on MATH, AIME, and GPQA benchmarks \cite{qwen2025qwq}.

% ----------------------------------------------------------------------------
\subparagraph{Qwen 3: Scaling Data and Hybrid Thinking.}
\label{subsubsec:qwen3}

Qwen 3, released on 28 April 2025 under the Apache 2.0 licence, represented a substantial generational advance over Qwen 2.5 across both scale and capability. The release encompassed a comprehensive suite of models spanning two architectural paradigms. The dense model lineup includes variants at 0.6B, 1.7B, 4B, 8B, 14B, and 32B parameters, providing coverage from on-device deployment to server-class inference. The MoE lineup comprises two configurations: a 30B-A3B model with 30 billion total parameters and 3 billion active parameters per forward pass, and the flagship 235B-A22B model with 235 billion total parameters and 22 billion active parameters. The MoE models employ fine-grained expert routing following the design principles established by DeepSeek, selecting a small subset of experts per token to maintain favourable inference costs despite the large total parameter count.

A central innovation of Qwen 3 is the introduction of a hybrid thinking mode that seamlessly integrates chain-of-thought reasoning with standard generation within a single model. Rather than requiring users to choose between a reasoning model (such as QwQ) and a general-purpose model (such as Qwen 2.5), Qwen 3 dynamically allocates reasoning depth based on the complexity of the input. For straightforward queries, the model produces responses directly; for complex mathematical, logical, or coding tasks, it generates extended internal reasoning traces before producing a final answer. This hybrid approach unifies the two inference paradigms that had previously required separate model deployments across the industry.

The training corpus for Qwen 3 comprises approximately 36 trillion tokens, double the volume used for Qwen 2.5, with expanded coverage of scientific literature, multilingual text, and synthetically generated reasoning chains. Training was conducted on Alibaba Cloud infrastructure, although the precise hardware configuration and total training cost have not been publicly disclosed. The doubling of training data, combined with improved data filtering and curriculum scheduling, yielded consistent improvements across standard benchmarks relative to both Qwen 2.5 and competing open-weight models of comparable scale.

% ----------------------------------------------------------------------------
\subparagraph{Qwen 3.5: Hybrid State-Space and Sparse MoE.}
\label{subsubsec:qwen35}

Qwen 3.5, released on 16 February 2026, introduced architectural innovations that depart significantly from the pure Transformer design shared by its predecessors \cite{qwen2024qwen25}. The flagship model is a 397B-A17B MoE configuration with 397 billion total parameters and 17 billion active parameters per forward pass. The most notable architectural change is the adoption of a hybrid architecture that combines Gated Delta Networks with sparse mixture-of-experts layers. Gated Delta Networks replace the conventional self-attention mechanism in a subset of layers with a state-space formulation that processes sequences through learned recurrent dynamics, achieving linear-time complexity in the sequence length while retaining the capacity for long-range information propagation. The remaining layers employ standard sparse MoE with softmax attention, preserving the representational flexibility of full attention where it is most beneficial. This hybrid design balances the computational efficiency of state-space models with the expressiveness of attention-based architectures.

Qwen 3.5 further adopts an early fusion approach to multimodal training, in which text, image, and video tokens are interleaved and processed jointly from the initial stages of pre-training rather than being integrated through post-hoc adapters. Early fusion enables the model to develop tightly integrated cross-modal representations, improving performance on tasks that require joint reasoning over textual and visual information. The Qwen 3.5 Small model family, encompassing variants from 0.8B to 9B parameters, is specifically optimised for on-device deployment on mobile phones and edge hardware, employing quantisation-aware training and architectural simplifications to achieve competitive quality within strict latency and memory constraints. The rapid progression from Qwen 2.5 through Qwen 3 to Qwen 3.5 over approximately eighteen months illustrates the accelerating pace of iteration within Chinese open-weight model development.

Qwen3-Coder-Next, released in February 2026 under the Apache 2.0 licence, is a specialised coding model with 80 billion total parameters and only 3 billion active parameters. The model employs a hybrid architecture combining Gated DeltaNet with Gated Attention layers and achieves 70.6\% on SWE-Bench Verified despite its small active parameter count, demonstrating that architectural innovation in attention mechanisms can compensate for reduced compute per token in domain-specialised applications.

% ============================================================================
\paragraph{Step-3.5 Flash and StepFun.}
\label{subsec:step}

Step-3.5 Flash, developed by StepFun, employs a 196 billion total and 11 billion active parameter MoE architecture with 288 routed experts and one shared expert, supporting 256K context length. The model introduces a three-way Multi-Token Prediction training objective and achieves exceptional mathematical reasoning performance, scoring 99.8\% on AIME 2025 and 98.0\% on HMMT 2025. Step-3.5 Flash is released under the Apache 2.0 licence.

% ============================================================================
\paragraph{MiMo-V2-Flash: Xiaomi's Software Engineering Model.}
\label{subsec:mimo}

MiMo-V2-Flash, developed by Xiaomi and released in December 2025, is a 309 billion total and 15 billion active parameter MoE model that combines hybrid sliding window and full attention mechanisms. The model demonstrates particular strength in software engineering tasks, achieving 73.4\% on SWE-Bench Verified, the leading score among open-source models at the time of its release and a result that places it in close contention with several proprietary systems.

% ============================================================================
\paragraph{Ling-1T: Trillion-Scale FP8 Training.}
\label{subsec:ling}

Ling-1T, developed by Ant Group under the InclusionAI initiative and released in October 2025 under the MIT licence, is the largest foundation model trained entirely with FP8 precision. With one trillion total parameters and approximately 50 billion active parameters, the model employs a mixture-of-experts architecture with a 1/32 activation ratio, achieving high capacity while maintaining efficient inference. The successful training at this scale with reduced numerical precision validates FP8 as a viable training format for trillion-parameter models.

% ============================================================================
\paragraph{Hunyuan 2.0: Tencent's Dual-Mode LLM.}
\label{subsec:hunyuan}

Hunyuan 2.0, developed by Tencent and released in December 2025, employs a 406 billion total and 32 billion active parameter MoE architecture with 256K context length. The model is available in two variants: Think, which incorporates extended chain-of-thought reasoning for complex tasks, and Instruct, which provides direct instruction-following responses with lower latency.

% ============================================================================
\paragraph{InternLM 2.5: Native Tool Use and Million-Token Context.}
\label{subsec:internlm}

InternLM 2.5, developed by the Shanghai AI Laboratory, distinguishes itself through two architectural emphases: native tool-use integration and extremely long context support \cite{cai2024internlm2}. Unlike models that acquire tool-use capabilities solely through post-training instruction tuning, InternLM 2.5 incorporates tool-use data directly into the pre-training corpus, exposing the model to structured function calls, API interactions, and tool-augmented reasoning chains during the foundational training stage. This approach yields more robust and reliable tool invocation behaviour compared to models that encounter tool-use patterns only during fine-tuning. The model is available in 7B and 20B parameter variants, both trained on large-scale multilingual corpora with particular strength in English and Chinese.

A notable technical contribution of InternLM 2.5 is its support for context windows exceeding one million tokens through dynamic NTK-aware Rotary Position Embedding interpolation. Standard RoPE interpolation methods degrade rapidly when extrapolating significantly beyond the pre-training context length, but the NTK-aware approach dynamically adjusts the frequency basis of the rotary embeddings based on the input sequence length, preserving positional resolution at both short and long ranges. This enables InternLM 2.5 to process extremely long documents, multi-turn dialogue histories, and large code repositories within a single context window. The Shanghai AI Laboratory has further extended the InternLM ecosystem to include InternVL, a vision-language model that pairs the InternLM language backbone with a vision encoder for multimodal understanding, and has open-sourced training frameworks and evaluation toolkits to support community research.

% ============================================================================
\paragraph{Yi and 01.AI.}
\label{subsec:yi}

The Yi model family, developed by 01.AI under the leadership of Kai-Fu Lee, has established itself as a competitive bilingual English-Chinese model series with strong performance across both language understanding and generation tasks \cite{young2024yi}. The initial Yi-34B model employed a dense Transformer architecture with design choices closely aligned with the LLaMA family, including RoPE, SwiGLU activations, and RMSNorm, while using an expanded 64,000-token vocabulary optimised for bilingual encoding efficiency. Yi-1.5 refined the base architecture with additional pre-training data and improved post-training alignment, while Yi-Lightning introduced a hybrid dense-MoE architecture that activates a subset of parameters per token to improve inference efficiency at scale.

The Yi ecosystem extends beyond language modelling to include Yi-VL for vision-language understanding and Yi-Coder for code generation, reflecting a broader trend among Chinese AI laboratories toward building comprehensive model families that span multiple modalities and application domains. The models are released under permissive licences and have been widely adopted by the Chinese open-source community for domain-specific fine-tuning. 01.AI's research agenda has emphasised data quality and training efficiency over raw parameter scale, a philosophy that has enabled the production of competitive models with training budgets substantially below those of larger laboratories.

% ============================================================================
\paragraph{Baichuan.}
\label{subsec:baichuan}

Baichuan 2, developed by Baichuan Inc., is an open-weight language model family optimised for Chinese language understanding and generation \cite{yang2023baichuan}. The 13B variant employs ALiBi (Attention with Linear Biases) positional encoding, which adds position-dependent linear biases directly to the attention scores rather than encoding position through learned or rotary embeddings, enabling efficient length extrapolation without explicit context window extension mechanisms; the 7B variant instead uses RoPE (Rotary Position Embedding). Baichuan 2 is available in 7B and 13B parameter variants and was trained on 2.6 trillion tokens with a strong emphasis on high-quality Chinese web text, encyclopaedic knowledge, and professional domain corpora. The model has been widely adopted for Chinese-language applications and serves as a popular base for domain-specific fine-tuning in industries including finance, healthcare, and legal services within the Chinese market.

% ============================================================================
\subsubsection{Europe and Rest of World}
\label{subsubsec:text_europe}

% ============================================================================
\paragraph{Mistral Large 3: European Frontier Open-Source AI.}
\label{subsec:mistral}

Mistral AI, based in Paris, has established itself as the leading European contributor to the open-weight model ecosystem. Mistral 7B employed sliding window attention (SWA) \cite{beltagy2020longformer}, in which each token attends only to a fixed-size window of preceding tokens, substantially reducing the memory and computational cost of attention while preserving the ability to propagate information through successive layers \cite{jiang2023mistral}. Mixtral 8x7B extended the architecture with a sparse mixture-of-experts layer, replacing every feed-forward block with eight parallel expert networks of which two are selected per token, achieving inference costs comparable to a dense 13B model while delivering performance competitive with LLaMA 2 70B \cite{jiang2024mixtral}.

Mistral Large 3, released in December 2025, represents the culmination of Mistral AI's progression toward frontier-scale open-weight models. The model was trained in European data centres, reflecting Mistral AI's strategic positioning as a European-headquartered alternative to American and Chinese providers; the specific hardware configuration and training cost have not been disclosed. The model employs a granular mixture-of-experts architecture with 675 billion total parameters and 41 billion active parameters per forward pass, positioning it in the same parameter class as DeepSeek-V3. Mistral Large 3 supports a context window of 256K tokens and is natively multimodal and multilingual, processing text and images across a broad set of languages. The model is released under the Apache 2.0 licence, the most permissive licensing option among frontier-scale MoE models. The Codestral series, including Codestral 25.08 released in July 2025, provides specialised coding variants optimised for software engineering tasks \cite{mistral2025large3}. Mistral AI's European base and its commitment to open-weight release under maximally permissive licences position it as a strategically significant alternative to both American and Chinese model providers, particularly for organisations subject to regulatory requirements that favour European-originated technology.

% ============================================================================
\subsubsection{Comparative Analysis of Text Generation Models}
\label{subsec:comparison}

Having surveyed the individual model families and their distinguishing technical contributions, this subsection draws them together into a unified comparative analysis. The comparison examines architectural choices, parameter efficiency through mixture-of-experts sparsity ratios, benchmark performance across reasoning, coding, and knowledge tasks, and the emergent competitive dynamics between open-weight and proprietary systems. Tables~\ref{tab:model_comparison}, \ref{tab:arena_rankings}, and \ref{tab:benchmark_comparison} present the quantitative evidence, while the accompanying discussion identifies the convergent trends and remaining points of differentiation among frontier models.

% ---- Table: Comprehensive Model Comparison ----
\begin{table*}[!htbp]
\centering
\caption{Comparative overview of modern large language models as of early 2026. Parameter counts are reported in billions (B) or trillions (T). ``Active Params'' indicates the number of parameters activated per forward pass; for dense models this equals total parameters. Context length is reported in thousands of tokens (K) or millions (M). ``Key Innovation'' highlights the primary distinguishing technical contribution of each model.}
\label{tab:model_comparison}
\scriptsize
\setlength{\tabcolsep}{2.5pt}
\renewcommand{\arraystretch}{1.05}
\begin{tabularx}{\textwidth}{@{}l l r r l r >{\raggedright\arraybackslash}X l@{}}
\toprule
Model & Organisation & \makecell[r]{Total\\Params} & \makecell[r]{Active\\Params} & Arch. & Context & Key Innovation & Licence \\
\midrule
\multicolumn{8}{@{}l}{\textit{United States}} \\[2pt]
Claude Opus 4.6 & Anthropic & --- & --- & --- & 200K & 72.7\% OSWorld, Constitutional AI & Closed \\
Claude Sonnet 4.6 & Anthropic & --- & --- & --- & 200K & Balanced coding + cost & Closed \\
Gemini 3.1 Pro & Google DeepMind & --- & --- & MoE & 1M & Improved tool use, ELO 1500 & Closed \\
Gemini 3 Pro & Google DeepMind & --- & --- & MoE & 1M & 100\% AIME (w/ tools), ARC-AGI-2 84.6\% & Closed \\
Gemini 2.5 Pro & Google DeepMind & --- & --- & MoE & 1M & Deep Think, native audio & Closed \\
Grok 4.20 & xAI & --- & --- & MoE & 128K & 4-agent parallel reasoning & Closed \\
GPT-5 & OpenAI & --- & --- & MoE & 400K & Unified reasoning + generation & Closed \\
GPT-5.2 & OpenAI & --- & --- & MoE & 400K & First ${>}$90\% ARC-AGI-1 Verified & Closed \\
GPT-5.4 & OpenAI & --- & --- & MoE & 1M & 75\% OSWorld, computer use & Closed \\
LLaMA 4 Scout & Meta & 109B & 17B & MoE & 10M & 10M context on single H100 & Llama CL \\
LLaMA 4 Maverick & Meta & 400B & 17B & MoE & 1M & 128 experts, multimodal & Llama CL \\
LLaMA 4 Behemoth & Meta & $\sim$2T & 288B & MoE & --- & Largest open MoE (training) & Llama CL \\
Gemma 3 & Google DeepMind & 27B & 27B & Dense & 128K & Alt. attn., vision, multilingual & Open \\
Phi-4 & Microsoft & 14B & 14B & Dense & 16K & Synthetic textbook data & Open \\
GPT-oss-120b & OpenAI & 117B & 5.1B & MoE & 128K & OpenAI's first open-weight & Apache 2.0 \\
Nemotron Ultra & NVIDIA & 253B & 253B & Dense & 128K & LLaMA 3.1 derivative & Open \\[4pt]
\multicolumn{8}{@{}l}{\textit{China}} \\[2pt]
GLM-5 & Zhipu AI & 744B & 40B & MoE & 128K & Ascend-only training, Agent Mode & MIT \\
Kimi K2 & Moonshot AI & $\sim$1T & 32B & MoE & 128K & Trillion-param open MoE & Mod. MIT \\
Kimi K2.5 & Moonshot AI & $\sim$1T & 32B & MoE & 128K & Agent Swarm (100 parallel agents) & Mod. MIT \\
DeepSeek-V3 & DeepSeek & 671B & 37B & MoE & 128K & Aux-loss-free MoE, MLA, FP8 & Open \\
DeepSeek-R1 & DeepSeek & 671B & 37B & MoE & 128K & Reasoning via pure RL (GRPO) & Open \\
DeepSeek-V3.2 & DeepSeek & 685B & 37B & MoE & 128K & Sparse Attention upgrade & MIT \\
DeepSeek-V4$^*$ & DeepSeek & --- & --- & MoE & 1M+ & Hyper-connections, Engram memory & Apache 2.0 \\
MiniMax M2.5 & MiniMax & --- & --- & MoE & 4M & RL on real-world environments & Open \\
MiniMax-01 & MiniMax & 456B & 45.9B & MoE & 4M & Lightning Attention (linear) & Open \\
Qwen3 235B-A22B & Alibaba & 235B & 22B & MoE & 128K & Hybrid thinking mode, 36T tokens & Apache 2.0 \\
Qwen3.5 397B-A17B & Alibaba & 397B & 17B & MoE & 1M & Gated Delta Net + sparse MoE & Apache 2.0 \\
Step-3.5 Flash & StepFun & 196B & 11B & MoE & 256K & 3-way MTP, AIME 99.8\% & Apache 2.0 \\
MiMo-V2-Flash & Xiaomi & 309B & 15B & MoE & 128K & Hybrid attn, SWE 73.4\% & Open \\
Ling-1T & Ant Group & 1T & 50B & MoE & 128K & Largest FP8-trained model & MIT \\
Hunyuan 2.0 & Tencent & 406B & 32B & MoE & 256K & Think + Instruct variants & Open \\
InternLM 2.5 & Shanghai AI Lab & 20B & 20B & Dense & 1M & Native tool use, NTK-RoPE & Open \\
Yi-34B & 01.AI & 34B & 34B & Dense & 200K & Bilingual, data quality focus & Open \\
Baichuan 2 & Baichuan Inc. & 13B & 13B & Dense & --- & Chinese-optimised (ALiBi/RoPE) & Open \\[4pt]
\multicolumn{8}{@{}l}{\textit{Europe and Rest of World}} \\[2pt]
Mistral Large 3 & Mistral AI & 675B & 41B & MoE & 256K & Granular MoE, multilingual & Apache 2.0 \\
\bottomrule
\multicolumn{8}{@{}l}{\footnotesize $^*$DeepSeek-V4 had not been released at the time of writing but was imminent (planned March 2026).}
\end{tabularx}
\end{table*}

The models surveyed in this section exhibit considerable diversity in architectural choices, training strategies, and design philosophies, yet several convergent trends are clearly visible. Table~\ref{tab:model_comparison} presents a comprehensive comparison across key dimensions including parameter counts, architectures, context lengths, and distinguishing innovations.

The mixture-of-experts architecture has become the overwhelmingly dominant paradigm for frontier-scale models (see \Cref{fig:moe_architecture} for a schematic and \Cref{fig:param_comparison} for a comparison of total and active parameter counts). Among the models surveyed, DeepSeek-V3 (671B/37B), GLM-5 (744B/40B), Qwen 3.5 (397B/17B), Mistral Large 3 (675B/41B), LLaMA 4 Maverick (400B/17B), Kimi K2.5 ($\sim$1T/32B), and LLaMA 4 Behemoth ($\sim$2T/288B) all employ sparse MoE architectures, while GPT-5, Gemini 3 Pro, and Grok 4.20 are widely understood to employ similar designs. The sparsity ratios vary considerably: DeepSeek-V3 activates approximately 5.5\% of its total parameters, LLaMA 4 Maverick activates approximately 4.3\%, and Kimi K2.5 activates approximately 3.2\%, demonstrating that frontier performance can be achieved with extremely sparse activation when expert routing is sufficiently well-designed. Dense models such as Phi-4 (14B), Gemma 3 (27B), and InternLM 2.5 (20B) continue to serve important roles at smaller scales, particularly for on-device deployment and scenarios where the memory overhead of hosting many inactive experts is impractical.

On standard benchmarks, the best open-weight models have largely closed the gap with proprietary systems. DeepSeek-V3, Qwen 3, and GLM-5 achieve scores on MMLU \cite{hendrycks2021measuring} within a few percentage points of GPT-5, while on SWE-bench Verified (\Cref{fig:swebench_comparison}), Claude Opus 4.6 (80.8\%) and MiniMax M2.5 (80.2\%) lead the field, followed by Gemini 3 Flash (78\%), GLM-5 (77.8\%), Gemini 3 Pro (76.2\%), GPT-5.2 (75.4\%), and GPT-5 (74.9\%). On HumanEval \cite{chen2021evaluating} for code generation and GSM8K \cite{cobbe2021training} for mathematical reasoning, the top open-weight models are competitive with or occasionally exceed their closed-source counterparts.

The reasoning-specialised models and reasoning modes (GPT-5, DeepSeek-R1, QwQ, Kimi K2.5, Gemini 3 Pro Deep Think, Claude Opus 4.6) represent a distinct competitive axis, with performance on challenging benchmarks such as GPQA Diamond, MATH-500, and AIME 2025 that substantially exceeds that of non-reasoning models regardless of parameter count. The integration of reasoning capabilities directly into general-purpose models, as exemplified by GPT-5's unified architecture and Qwen 3's hybrid thinking mode, suggests that the distinction between ``reasoning models'' and ``general models'' is collapsing in favour of unified systems that adaptively allocate inference-time computation.

The emergence of agentic capabilities represents a third competitive axis beyond raw knowledge and reasoning. GLM-5's native Agent Mode, Kimi K2.5's Agent Swarm technology, Claude Opus 4.6's computer use (72.7\% on OSWorld), MiniMax M2.5's environment-grounded RL training, and Gemini 3 Pro's agentic capabilities all represent different approaches to enabling models to act autonomously in digital environments. The arena-style evaluation framework of \cite{zheng2024judging} has provided an additional perspective beyond traditional benchmarks, capturing subjective quality dimensions such as helpfulness, fluency, and instruction-following fidelity that static benchmarks may not fully reflect.

The strategic dimension of hardware independence has emerged as a significant theme, with GLM-5's successful training entirely on Huawei Ascend chips demonstrating that competitive frontier-scale training is achievable without NVIDIA hardware. This development has implications for the long-term structure of the AI industry, as it reduces the leverage of export controls on advanced GPU hardware and enables AI development to proceed on domestically manufactured accelerators.

A particularly revealing aspect of the current Arena is the proliferation of modality-specific leaderboards. Beyond the overall text ranking, the Arena now evaluates models separately on code generation, vision understanding, document analysis, search (web-grounded retrieval), text-to-image generation, image editing, text-to-video, and image-to-video synthesis. Different model families exhibit markedly different profiles across these categories: Gemini 3 Pro leads the vision leaderboard (ELO 1309) by a substantial margin, Claude Opus 4.6 dominates code (ELO 1561) and document understanding (ELO 1525), and Google's Veo 3.1 and OpenAI's Sora 2 lead the video generation categories. This multi-dimensional evaluation framework underscores that the notion of a single ``best'' model has given way to a landscape in which leadership is fragmented across modalities, and differentiation increasingly depends on specialised capabilities such as agentic tool use, computer interaction, long-context processing, multimodal understanding, and alignment quality rather than raw benchmark scores alone.

\FloatBarrier
% ---- Table: Chatbot Arena Leaderboard Rankings ----
\begin{table*}[!htbp]
\centering
\caption{Chatbot Arena (LMArena) text ELO rankings as of March 2026. Rankings reflect crowdsourced human preference evaluations across diverse conversational and reasoning tasks. The top section shows proprietary models; the bottom section shows the leading open-weight models.}
\label{tab:arena_rankings}
\small
\setlength{\tabcolsep}{4pt}
\renewcommand{\arraystretch}{1.12}
\begin{tabularx}{\textwidth}{@{}c l r l l@{}}
\toprule
Rank & Model & ELO & Developer & Type \\
\midrule
\multicolumn{5}{@{}l}{\textit{Proprietary Models}} \\[2pt]
1 & Claude Opus 4.6 & 1504 & Anthropic & Closed \\
2 & Gemini 3.1 Pro & 1500 & Google DeepMind & Closed \\
3 & Claude Opus 4.6 (thinking) & 1500 & Anthropic & Closed \\
4 & Grok 4.20 beta & 1493 & xAI & Closed \\
5 & Gemini 3 Pro & 1485 & Google DeepMind & Closed \\
6 & GPT-5.2 & 1481 & OpenAI & Closed \\
7 & GPT-5.4 & 1480 & OpenAI & Closed \\
8 & Gemini 3 Flash & 1473 & Google DeepMind & Closed \\
9 & Grok 4.1 (thinking) & 1473 & xAI & Closed \\[4pt]
\multicolumn{5}{@{}l}{\textit{Open-Weight Models}} \\[2pt]
1 & GLM-5 & 1451 & Zhipu AI & Open \\
2 & Kimi K2.5 & 1447 & Moonshot AI & Mod. MIT \\
3 & DeepSeek-R1 & 1436 & DeepSeek & Open \\
4 & DeepSeek-V3.2 & 1421 & DeepSeek & MIT \\
5 & Mistral Large 3 & 1418 & Mistral AI & Apache 2.0 \\
6 & LLaMA 4 Maverick & 1417 & Meta & Llama CL \\
7 & Gemma 3 27B & 1339 & Google DeepMind & Open \\
\bottomrule
\end{tabularx}
\end{table*}

% ---- Table: Benchmark Comparison ----
\begin{table*}[!htbp]
\centering
\caption{Benchmark performance comparison across frontier proprietary and open-weight models. MMLU \cite{hendrycks2021measuring} measures broad knowledge; AIME 2025 evaluates competition-level mathematical reasoning; SWE-bench Verified \cite{jimenez2024swebench} measures real-world software engineering; HumanEval \cite{chen2021evaluating} evaluates code generation; GSM8K \cite{cobbe2021training} tests mathematical word problems; GPQA Diamond assesses graduate-level science reasoning. All scores are percentages. Entries marked ``---'' indicate scores not publicly available at the time of writing.}
\label{tab:benchmark_comparison}
\small
\setlength{\tabcolsep}{4pt}
\renewcommand{\arraystretch}{1.12}
\begin{tabularx}{\textwidth}{@{}l r r r r r r@{}}
\toprule
Model & MMLU & \makecell[r]{AIME\\2025} & \makecell[r]{SWE-bench\\Verified} & HumanEval & GSM8K & \makecell[r]{GPQA\\Diamond} \\
\midrule
GPT-5 & 91.4 & 94.6 & 74.9 & --- & --- & --- \\
GPT-5.2 & --- & 100 & 75.4$^\P$ & --- & --- & --- \\
GPT-5.4 & --- & --- & --- & --- & --- & --- \\
Claude Opus 4.6 & --- & --- & 80.8 & 90.4 & 99.0 & --- \\
Gemini 3 Pro & 90.1$^\S$ & 95$^\|$ & 76.2 & --- & --- & 91.9 \\
Gemini 2.5 Pro & --- & 86.7 & 63.8 & --- & --- & 84.0 \\
Grok 4.20 & --- & --- & --- & --- & --- & --- \\
DeepSeek-V3 & 88.5 & 39.2$^*$ & 42.0 & 82.6$^\ddagger$ & 89.3 & 59.1 \\
DeepSeek-R1 & 90.8 & 79.8$^\dagger$ & 49.2 & --- & --- & 71.5 \\
GLM-5 & 91.7 & --- & 77.8 & 92.5 & --- & --- \\
Qwen3 235B & 87.9 & 81.5 & --- & 90.8 & 94.4 & 65.8 \\
MiniMax M2.5 & --- & --- & 80.2 & --- & --- & --- \\
LLaMA 4 Maverick & 85.5 & --- & --- & --- & 91.5 & 69.8 \\
Step-3.5 Flash & --- & 99.8 & --- & --- & --- & --- \\
MiMo-V2-Flash & --- & --- & 73.4 & --- & --- & --- \\
Mistral Large 3 & --- & --- & --- & 89.2 & 94.1 & --- \\
\bottomrule
\multicolumn{7}{@{}l}{\footnotesize $^*$AIME 2024 score; V3 was not evaluated on AIME 2025.}\\
\multicolumn{7}{@{}l}{\footnotesize $^\dagger$AIME 2024 score; R1-0528 scored 87.5\% on AIME 2025.}\\
\multicolumn{7}{@{}l}{\footnotesize $^\ddagger$HumanEval-Mul (multilingual); standard HumanEval (Python) is 65.2\%.}\\
\multicolumn{7}{@{}l}{\footnotesize $^\S$MMLU Pro score (harder variant); standard MMLU not separately reported.}\\
\multicolumn{7}{@{}l}{\footnotesize $^\P$Independent evaluation (Vals.ai); OpenAI self-reports 80.0\% with Thinking mode enabled.}\\
\multicolumn{7}{@{}l}{\footnotesize $^\|$Without tool augmentation; achieves 100\% with code execution enabled.}
\end{tabularx}
\end{table*}

% ---- Figure: SWE-bench Verified Comparison ----
\definecolor{geoUS}{RGB}{63,81,181}
\definecolor{geoCN}{RGB}{0,150,136}

\begin{figure*}[t]
\centering
\begin{tikzpicture}
\begin{axis}[
    width=0.88\textwidth,
    height=7.5cm,
    xbar,
    bar width=12pt,
    enlarge y limits=0.10,
    xlabel={SWE-bench Verified Performance (\%)},
    xlabel style={font=\normalsize},
    xmin=65,
    xmax=85,
    xtick={65,70,75,80,85},
    ytick={0,1,2,3,4,5,6,7},
    yticklabels={
        Qwen3-Coder-Next,
        MiMo-V2-Flash,
        GPT-5,
        Gemini 3 Pro,
        GLM-5,
        Gemini 3 Flash,
        MiniMax M2.5,
        Claude Opus 4.6
    },
    y tick label style={font=\small},
    x tick label style={font=\small},
    grid=major,
    grid style={gray!25},
    major tick length=3pt,
    axis line style={gray!70},
    nodes near coords,
    nodes near coords style={font=\footnotesize, anchor=west},
    every node near coord/.append style={xshift=2pt},
    legend style={at={(0.97,0.03)}, anchor=south east, font=\small, draw=black!50, fill=white},
]
% United States models
\addplot[fill=geoUS!65, draw=geoUS!80!black, bar shift=0pt] coordinates {
    (74.9, 2)
    (76.2, 3)
    (78.0, 5)
    (80.8, 7)
};
% China models
\addplot[fill=geoCN!65, draw=geoCN!80!black, bar shift=0pt] coordinates {
    (70.6, 0)
    (73.4, 1)
    (77.8, 4)
    (80.2, 6)
};
\legend{United States, China}
\end{axis}
\end{tikzpicture}
\caption{SWE-bench Verified performance (\%) for top models as of March 2026, coloured by geographic origin. The leading US model (Claude Opus 4.6, 80.8\%) and Chinese model (MiniMax M2.5, 80.2\%) achieve near-identical scores, illustrating convergence in software engineering capability across regions.}
\label{fig:swebench_comparison}
\end{figure*}

% ---- Figure: Arena ELO Comparison ----
\begin{figure*}[t]
\centering
\begin{tikzpicture}
\begin{axis}[
    width=0.88\textwidth,
    height=8cm,
    xbar,
    bar width=12pt,
    enlarge y limits=0.08,
    xlabel={Arena Text ELO},
    xlabel style={font=\normalsize},
    xmin=1420,
    xmax=1520,
    xtick={1420,1440,1460,1480,1500,1520},
    ytick={0,1,2,3,4,5,6,7,8,9,10,11},
    yticklabels={
        DeepSeek-R1,
        Kimi K2.5,
        GLM-5,
        Grok 4.1 (thinking),
        Gemini 3 Flash,
        GPT-5.4,
        GPT-5.2,
        Gemini 3 Pro,
        Grok 4.20 beta,
        Claude Opus 4.6 (thinking),
        Gemini 3.1 Pro,
        Claude Opus 4.6
    },
    y tick label style={font=\small},
    x tick label style={font=\small},
    grid=major,
    grid style={gray!25},
    major tick length=3pt,
    axis line style={gray!70},
    nodes near coords,
    nodes near coords style={font=\small, anchor=west},
    every node near coord/.append style={xshift=2pt},
    clip=false,
    legend style={at={(0.97,0.03)}, anchor=south east, font=\small, draw=black!50, fill=white},
]
% United States models
\addplot[fill=geoUS!65, draw=geoUS!80!black, bar shift=0pt] coordinates {
    (1473, 3)
    (1473, 4)
    (1480, 5)
    (1481, 6)
    (1485, 7)
    (1493, 8)
    (1500, 9)
    (1500, 10)
    (1504, 11)
};
% China models
\addplot[fill=geoCN!65, draw=geoCN!80!black, bar shift=0pt] coordinates {
    (1436, 0)
    (1447, 1)
    (1451, 2)
};
\legend{United States, China}
\end{axis}
\end{tikzpicture}
\caption{Chatbot Arena text ELO ratings as of March 2026, coloured by geographic origin. The chart includes both proprietary and open-weight models. Claude Opus 4.6 (Anthropic) leads at 1504, followed by Gemini 3.1 Pro (Google DeepMind) at 1500. The top open-weight models are all from Chinese laboratories: GLM-5 (1451), Kimi K2.5 (1447), and DeepSeek-R1 (1436).}
\label{fig:arena_elo}
\end{figure*}

The Chatbot Arena (LMArena) ELO rankings as of March 2026 reveal a fiercely competitive landscape across both proprietary and open-weight models. As shown in Table~\ref{tab:arena_rankings} and Figure~\ref{fig:arena_elo}, Claude Opus 4.6 from Anthropic leads the overall text leaderboard at 1504, followed closely by Gemini 3.1 Pro (1500), the Claude Opus 4.6 thinking variant (1500), and Grok 4.20 beta (1493). Gemini 3 Pro (1485), GPT-5.2 (1481), GPT-5.4 (1480), and Gemini 3 Flash (1473) complete a densely packed proprietary tier in which the top ten models span only 31 ELO points. The Arena evaluates models across multiple modality-specific categories, including text, code, vision, document understanding, and search; different model families lead in different categories, with Gemini 3 Pro dominating vision, Claude Opus 4.6 leading in code and document understanding, and Grok 4.20 placing second in search. Among open-weight models, Chinese laboratories hold the top positions, with GLM-5 from Zhipu AI at 1451 and Kimi K2.5 from Moonshot AI at 1447, followed by DeepSeek-R1 (1436), DeepSeek-V3.2 (1421), Mistral Large 3 (1418), and LLaMA 4 Maverick (1417). The narrowing gap between the best open-weight model (GLM-5 at 1451) and the top proprietary model (Claude Opus 4.6 at 1504), just 53 ELO points, suggests that the capability premium commanded by closed-source models is diminishing rapidly.

\FloatBarrier
% ---- Figure 1: Model Parameter Counts by Organisation ----
\definecolor{chinesebar}{RGB}{0,150,136}
\definecolor{chinesebaractive}{RGB}{77,208,225}
\definecolor{westernbar}{RGB}{63,81,181}
\definecolor{westernbaractive}{RGB}{159,168,218}

\begin{figure*}[t]
\centering
\begin{tikzpicture}
\begin{axis}[
    width=0.97\textwidth,
    height=10cm,
    ybar,
    bar width=7pt,
    enlarge x limits=0.03,
    ylabel={Parameters (billions)},
    ylabel style={font=\large},
    symbolic x coords={
        GLM-5,
        DeepSeek-V3,
        MiniMax-01,
        Qwen3.5,
        Qwen3-MoE,
        Kimi K2,
        InternLM,
        Yi-34B,
        Baichuan,
        LLaMA4-Mav,
        Mistral-L3,
        Grok-1,
        LLaMA3,
        Gemma 3,
        Phi-4
    },
    xtick=data,
    x tick label style={
        rotate=40,
        anchor=east,
        font=\normalsize
    },
    ytick style={font=\normalsize},
    yticklabel style={font=\normalsize},
    ymin=0,
    ymax=800,
    ytick={0,100,200,300,400,500,600,700,800},
    legend style={
        at={(0.02,0.97)},
        anchor=north west,
        font=\normalsize,
        draw=black!50,
        fill=white,
        fill opacity=0.9,
        text opacity=1
    },
    legend cell align={left},
    grid=major,
    grid style={gray!25},
    major tick length=3pt,
    axis line style={gray!70},
    clip=false
]

% Total parameters
\addplot[fill=deepblue!70, draw=deepblue] coordinates {
    (GLM-5, 744)
    (DeepSeek-V3, 671)
    (MiniMax-01, 456)
    (Qwen3.5, 397)
    (Qwen3-MoE, 235)
    (Kimi K2, 750)
    (InternLM, 20)
    (Yi-34B, 34)
    (Baichuan, 13)
    (LLaMA4-Mav, 400)
    (Mistral-L3, 675)
    (Grok-1, 314)
    (LLaMA3, 405)
    (Gemma 3, 27)
    (Phi-4, 14)
};

% Active parameters
\addplot[fill=grokcolor!70, draw=grokcolor!90!black] coordinates {
    (GLM-5, 40)
    (DeepSeek-V3, 37)
    (MiniMax-01, 45.9)
    (Qwen3.5, 17)
    (Qwen3-MoE, 22)
    (Kimi K2, 32)
    (InternLM, 20)
    (Yi-34B, 34)
    (Baichuan, 13)
    (LLaMA4-Mav, 17)
    (Mistral-L3, 41)
    (Grok-1, 314)
    (LLaMA3, 405)
    (Gemma 3, 27)
    (Phi-4, 14)
};

\legend{Total Parameters, Active Parameters}

% Grouping labels
\draw[decorate, decoration={brace, amplitude=6pt, mirror}, thick, deepseekcolor!80!black]
    ([yshift=-32pt]axis cs:GLM-5, 0) -- node[below=8pt, font=\small\sffamily, text=deepseekcolor!80!black] {Chinese Labs}
    ([yshift=-32pt]axis cs:Baichuan, 0);

\draw[decorate, decoration={brace, amplitude=6pt, mirror}, thick, llamacolor!80!black]
    ([yshift=-32pt]axis cs:LLaMA4-Mav, 0) -- node[below=8pt, font=\small\sffamily, text=llamacolor!80!black] {Western Labs}
    ([yshift=-32pt]axis cs:Phi-4, 0);

\end{axis}
\end{tikzpicture}
\caption{Total and active parameter counts for major open-weight large language models with disclosed parameter counts, grouped by laboratory origin. Mixture-of-experts models (GLM-5, DeepSeek-V3, MiniMax-01, Qwen~3.5, Qwen~3, Kimi~K2, LLaMA~4 Maverick, Mistral Large~3) exhibit a large disparity between total and active parameters, reflecting the computational efficiency of sparse activation. Dense models (LLaMA~3, Yi-34B, Gemma~3, Phi-4, InternLM~2.5) activate all parameters for every token. Note: Kimi~K2 total parameter count is plotted at 750B as a conservative lower bound of its reported $\sim$1T total; LLaMA~4 Behemoth ($\sim$2T) is omitted for axis readability.}
\label{fig:param_comparison}
\end{figure*}

% ---- Figure 2: Mixture-of-Experts Architecture ----
\begin{figure}[t]
\centering
\begin{tikzpicture}[
    node distance=0.6cm and 0.8cm,
    box/.style={
        rectangle,
        draw=black!70,
        fill=#1!15,
        minimum width=1.8cm,
        minimum height=0.7cm,
        font=\small\sffamily,
        rounded corners=2pt,
        line width=0.8pt
    },
    expert/.style={
        rectangle,
        draw=deepseekcolor!80!black,
        fill=deepseekcolor!12,
        minimum width=1.4cm,
        minimum height=0.6cm,
        font=\small\sffamily,
        rounded corners=2pt,
        line width=0.8pt
    },
    arr/.style={
        ->,
        >=Stealth,
        thick,
        draw=black!60
    },
    darr/.style={
        ->,
        >=Stealth,
        thick,
        draw=deepseekcolor!70!black,
        dashed
    }
]

% Input
\node[box=lightblue] (input) {Token $\hidden$};

% Router
\node[box=lightorange, above=0.8cm of input] (router) {Router};

% Experts
\node[expert, above left=1.0cm and 0.3cm of router] (e1) {Expert 1};
\node[expert, left=0.15cm of e1] (e2) {Expert 2};
\node[expert, right=0.15cm of e1] (dots) {$\cdots$};
\node[expert, right=0.15cm of dots] (en) {Expert $N$};

% Shared expert
\node[box=lightpurple, right=0.6cm of router] (shared) {\footnotesize Shared Expert};

% Gating
\node[box=lightyellow, above=0.8cm of dots] (gate) {Gating Weights $g_i$};

% Weighted sum
\node[box=lightgreen, above=0.6cm of gate] (sum) {Weighted Sum};

% Output
\node[box=lightblue, above=0.6cm of sum] (output) {Output};

% Arrows from input to router
\draw[arr] (input) -- (router);

% Arrows from router to experts (selection)
\draw[darr] (router.north) -- ++(0,0.25) -| (e2.south);
\draw[arr] (router.north) -- ++(0,0.25) -| (e1.south);
\draw[darr] (router.north) -- ++(0,0.25) -| (dots.south);
\draw[arr] (router.north) -- ++(0,0.25) -| (en.south);

% Arrow from input to shared expert
\draw[arr] (input.east) -| (shared.south);

% Arrows from experts to gating
\draw[arr] (e1.north) |- (gate.west);
\draw[arr] (en.north) |- (gate.east);

% Arrow from shared to sum
\draw[arr] (shared.north) |- (sum.east);

% Gating to sum
\draw[arr] (gate) -- (sum);

% Sum to output
\draw[arr] (sum) -- (output);

% Selection annotation
\node[right=0.15cm of router, yshift=0.45cm, font=\scriptsize\itshape, text=black!60] {top-$k$};

% Legend for dashed arrows
\node[below=0.2cm of input, font=\scriptsize\itshape, text=black!50] {Solid: selected \quad Dashed: not selected};

\end{tikzpicture}
\caption{Schematic of the mixture-of-experts (MoE) architecture used in models such as DeepSeek-V3, GLM-5, Qwen~3.5, Mistral Large~3, and LLaMA~4. The router computes affinity scores between the input token representation $\hidden$ and each expert, selects the top-$k$ experts (solid arrows), and computes gating weights $g_i$ to combine their outputs. A shared expert processes every token unconditionally. Unselected experts (dashed arrows) incur no computational cost. This sparse activation paradigm enables models with hundreds of billions or trillions of total parameters to maintain inference costs comparable to dense models with far fewer parameters.}
\label{fig:moe_architecture}
\end{figure}

% ============================================================================
\subsection{Image Generation Models}
\label{subsec:image_models}

The image generation landscape in 2025--2026 has been shaped by two converging trends: the maturation of standalone diffusion and flow-matching models, and the integration of native image generation capabilities directly into large language models. The Arena text-to-image leaderboard, with over 4.1 million votes across 50 models as of March 2026, provides the primary crowdsourced evaluation framework for this modality. A companion image-editing leaderboard, with over 24 million votes across 39 models, evaluates the distinct but related capability of modifying existing images according to natural-language instructions. This subsection surveys the leading image generation models organised by geographic origin, ordered by Arena performance within each region.

% ============================================================================
\subsubsection{United States}
\label{subsubsec:image_us}

% ============================================================================
\paragraph{GPT Image 1.5: Native Autoregressive Image Generation.}
\label{subsec:gpt_image}

GPT Image 1.5, released by OpenAI on 16 December 2025, represents a paradigm shift from the diffusion-based DALL-E series to native autoregressive image generation integrated directly within the GPT architecture. Rather than employing a separate diffusion system, GPT Image 1.5 generates images through the same transformer backbone that processes text, treating visual output as a sequence of tokens produced autoregressively. This architectural unification enables the model to leverage the world knowledge and reasoning capabilities of the underlying language model during image generation, producing outputs that demonstrate superior prompt adherence, contextual understanding, and text rendering accuracy compared to diffusion-only approaches. The model supports generation and editing with four quality tiers (low, medium, high, auto) that trade generation speed against visual detail. On the Arena text-to-image leaderboard, GPT Image 1.5 ranks first with an ELO of 1264, establishing a 29-point lead over the second-ranked model. Its text rendering capability---producing accurate embedded text in images with approximately 85--90\% fidelity---substantially exceeds that of competing models and has made it the preferred choice for design, marketing, and infographic generation workflows. DALL-E 3, OpenAI's previous diffusion-based image model, has been deprecated with support ending in May 2026.

% ============================================================================
\paragraph{Gemini Image Generation.}
\label{subsec:gemini_image}

Google DeepMind's Gemini models incorporate native image generation capabilities powered by the Imagen architecture integrated within the multimodal Gemini framework. Gemini 3 Pro Image ranks second on the Arena text-to-image leaderboard with an ELO of 1235, demonstrating strong photorealism and compositional accuracy. Gemini 2.5 Flash Image, with over 649,000 Arena votes, is the most extensively evaluated image model on the leaderboard, providing exceptionally high statistical confidence in its rating. The integration of image generation within the Gemini conversational interface enables iterative refinement workflows in which users can generate an image and then modify it through follow-up natural-language instructions, leveraging the model's multimodal understanding to maintain consistency across editing iterations.

% ============================================================================
\paragraph{Midjourney V7.}
\label{subsec:midjourney}

Midjourney V7, released in alpha on 3 April 2025 and set as the default model on 17 June 2025, represents a complete architectural rebuild from scratch rather than an incremental update over V6. The model introduces Draft Mode, which generates images ten times faster and at half the cost, enabling a conversational editing workflow in which users can instruct changes via text or voice and observe results in near real-time. V7 activates model personalisation by default, adapting outputs to individual user aesthetic preferences over time. Midjourney has further expanded into video generation, producing 5--21 second clips from static images. As a bootstrapped research laboratory of approximately 100 employees that has never taken outside funding, Midjourney occupies a distinctive position in the image generation ecosystem, prioritising aesthetic quality and creative exploration over API-driven enterprise deployment. The company operates on a subscription model ranging from \$10 to \$120 per month with no free tier.

% ============================================================================
\paragraph{Ideogram 3.0.}
\label{subsec:ideogram}

Ideogram 3.0, released on 26 March 2025, has established itself as the leading model for text rendering within images, achieving roughly 90--95\% accuracy on embedded text generation---a capability where competing models such as Midjourney (30--40\% accuracy) and earlier diffusion models have historically struggled. The model supports style referencing through up to three uploaded reference images and includes Canvas editing tools (Magic Fill and Extend) for interactive post-generation refinement. Ideogram was founded in 2022 by researchers including Mohammad Norouzi and Jonathan Ho, key contributors to the development of diffusion models, and has raised over \$96 million in funding.

% ============================================================================
\subsubsection{China}
\label{subsubsec:image_china}

% ============================================================================
\paragraph{Seedream 4.5 and ByteDance.}
\label{subsec:seedream}

Seedream 4.5, developed by ByteDance and released in December 2025, consolidates image generation and editing within a single architecture that interprets spatial references directly from natural-language prompts. The model ranks tenth on the Arena text-to-image leaderboard with an ELO of 1147, supports native 4K output, and can reference up to 10 images per edit for complex multi-source workflows including product swaps, text overlay copying, and element repositioning. Seedream 4.5 integrates with ByteDance's Doubao AI assistant, which reached 163 million monthly active users by December 2025, making it the most widely deployed consumer-facing image generation capability in China. The subsequent Seedream 5.0 Lite, released in February 2026, introduced improved spatial understanding and aesthetic quality.

% ============================================================================
\paragraph{Hunyuan Image 3.0 and Tencent.}
\label{subsec:hunyuan_image}

Hunyuan Image 3.0, developed by Tencent and added to the Arena image-editing leaderboard in January 2026, extends Tencent's Hunyuan multimodal ecosystem with dedicated image generation and editing capabilities. The model ranks among the top ten on the Arena text-to-image leaderboard with an ELO of approximately 1150, with particular strength in character and anime-style generation. The instruct variant supports multi-image editing workflows, enabling coherent modifications across multiple reference images within a single generation pass.

% ============================================================================
\subsubsection{Europe and Rest of World}
\label{subsubsec:image_europe}

% ============================================================================
\paragraph{FLUX 2: Rectified Flow at Scale.}
\label{subsec:flux}

FLUX 2, developed by Black Forest Labs (Germany) and released in November 2025, is a 32 billion parameter model built on a latent flow-matching architecture that couples a Mistral-3 24B vision-language model with a rectified flow transformer \cite{bfl2025flux2}. Unlike traditional diffusion models that learn to progressively denoise from Gaussian noise, rectified flow transformers learn direct mappings between noise distributions and data distributions along straight-line trajectories in latent space, resulting in faster generation with fewer sampling steps. The system employs a redesigned variational autoencoder (VAE) that addresses the ``learnability-quality-compression'' trilemma by retraining the latent space from scratch. FLUX 2 is available in five variant families: Max (highest quality), Pro (production-grade), Flex (flexible aspect ratios), Dev (open-weight for research), and Klein (compact, 4B and 9B parameters optimised for sub-second generation on modern GPUs). Three FLUX 2 variants rank in the Arena text-to-image top ten, with FLUX 2 Max achieving an ELO of approximately 1168, making Black Forest Labs the most represented organisation in the top tier. Black Forest Labs, founded by the creators of Stable Diffusion and headquartered in Germany, raised \$300 million in Series B funding in December 2025, bringing total capitalisation to approximately \$450 million.

% ============================================================================
\subsubsection{Comparative Analysis of Image Generation Models}
\label{subsec:image_comparison}

The Arena text-to-image leaderboard reveals a mature competitive landscape in which the gap between the second-ranked model (Gemini 3 Pro Image at 1235) and the ninth-ranked model spans only 88 ELO points, indicating that multiple approaches---autoregressive (GPT Image 1.5), flow-matching (FLUX 2), and integrated multimodal (Gemini)---have converged to comparable quality levels. The dominant trend is the integration of image generation directly into large language models: GPT Image 1.5 and Gemini Image generation both leverage their underlying LLM's world knowledge and reasoning capabilities to produce images with superior prompt adherence, contextual understanding, and text rendering compared to standalone diffusion models. This architectural convergence suggests that future image generation leadership will increasingly depend on the quality of the underlying language model rather than on innovations in the generative decoder alone.

A second significant trend is the emergence of unified generation-and-editing architectures. GPT Image 1.5, FLUX 2, Seedream 4.5, and Gemini all consolidate creation and modification within a single model, eliminating the need for separate specialised tools for image editing. The Arena's image-editing leaderboard, with over 24 million votes, now evaluates single-image and multi-image editing as distinct categories, reflecting the practical importance of editing workflows in commercial image generation. Geographically, United States laboratories (OpenAI, Google DeepMind) dominate the top two positions, while European (Black Forest Labs) and Chinese (ByteDance, Tencent) organisations compete strongly from third place onward, demonstrating a more geographically distributed competitive landscape than the text generation domain.

\FloatBarrier

% ============================================================================
\subsection{Video Generation Models}
\label{subsec:video_models}

The video generation modality has experienced the most rapid capability advancement of any generative AI domain during 2025--2026, progressing from short, visually inconsistent clips to cinematic-quality productions with synchronised audio, native 4K resolution, and coherent multi-shot narratives. The Arena text-to-video leaderboard, with over 246,000 votes across 37 models as of March 2026, provides the primary crowdsourced evaluation, complemented by a separate image-to-video leaderboard. A defining trend is the emergence of native audio-visual co-generation, in which video and synchronised audio (dialogue, sound effects, ambient noise) are produced jointly within a single generation pass rather than through sequential pipelines.

% ============================================================================
\subsubsection{United States}
\label{subsubsec:video_us}

% ============================================================================
\paragraph{Veo 3.1: Google DeepMind's Audio-Visual Synthesis.}
\label{subsec:veo}

The Veo model family from Google DeepMind has established video generation as a first-class capability within Google's AI portfolio. Veo 2, released in December 2024, introduced 4K resolution video generation with improved understanding of real-world physics. Veo 3, released in May 2025, marked a generational advance by introducing native audio generation---the model produces synchronised dialogue, sound effects, and ambient noise alongside the visual content within a single generation pass. The underlying architecture employs a Latent Diffusion Transformer that processes video and audio data in a compressed latent space, applying the diffusion process jointly to both visual and audio latents. At each denoising step, the model's attention mechanism operates on a unified sequence of tokens representing both visual spacetime patches and temporal audio information, ensuring synchronisation throughout generation.

Veo 3.1, initially released in October 2025 and subsequently updated in January 2026 with vertical video and 4K upscaling support, further extends the architecture with 4K resolution output at 3840$\times$2160 pixels, native vertical video support (9:16 aspect ratio), and reference image conditioning for character and style consistency. Individual clips are limited to 8 seconds, but scene extension capabilities allow chaining of multiple generations to produce sequences up to 148 seconds with narrative consistency. On the Arena text-to-video leaderboard, the Veo 3.1 series claimed the top position with an ELO of 1386 as of December 2025, with Sora 2 Pro subsequently tying for first place. The top three positions on the leaderboard are occupied by models with native audio capabilities (Veo 3.1 Fast, Veo 3.1, Veo 3 Fast), suggesting that users strongly prioritise multimodal coherence in their evaluations. All Veo-generated videos include SynthID watermarking for AI content identification.

% ============================================================================
\paragraph{Sora 2: OpenAI's Diffusion Transformer for Video.}
\label{subsec:sora}

Sora 2, released by OpenAI on 30 September 2025, represents a substantial advance over the original Sora model announced in early 2024. The model employs a diffusion transformer architecture that treats video data as sequences of visual patches over time, analogous to how language models process text tokens. This patch-based representation enables the model to learn both spatial details within frames and temporal relationships between frames. Sora 2 generates videos of 10--25 seconds with synchronised audio including dialogue, sound effects, and ambient noise. OpenAI has not disclosed the parameter count, training data composition, or training cost.

Sora 2 Pro, the production-quality variant, was the first model to tie with Veo 3 variants for first place on the Arena text-to-video leaderboard, while the standard Sora 2 variant ranked third. The model demonstrates substantially improved physics simulation compared to its predecessor, with realistic handling of object collisions, momentum, buoyancy, and gravitational dynamics. Users can inject real-world elements---people, animals, objects---into generated environments with accurate portrayal of appearance and, in some cases, voice. Current limitations include a 25-second maximum duration, challenges with hand and facial detail consistency, and imperfect text rendering within video frames. Sora 2 is accessible through a standalone iOS app, the sora.com web platform, and the OpenAI API.

% ============================================================================
\paragraph{Runway Gen-4.5.}
\label{subsec:runway}

Runway Gen-4.5, released in December 2025, represents the latest iteration from Runway, a pioneering commercial video generation company headquartered in New York. The model was developed end-to-end on NVIDIA Hopper and Blackwell GPUs, achieving advances in pre-training data efficiency and post-training techniques. Gen-4.5 produces highly detailed video with physically plausible motion---objects exhibit proper weight and momentum, liquids behave realistically, and textures remain consistent during complex motion. The model supports detailed camera choreography, intricate scene compositions, and precise timing of events within a single prompt. Recent updates introduced native audio generation, multi-shot sequencing, and character-consistent long-form video support up to one minute. On the Artificial Analysis text-to-video leaderboard, Gen-4.5 achieved an ELO of 1247, the highest score on that independent evaluation platform, while ranking fifth on the Arena leaderboard with an ELO of approximately 1225. Gen-4.5 is available across all paid Runway subscription tiers.

% ============================================================================
\subsubsection{China}
\label{subsubsec:video_china}

% ============================================================================
\paragraph{Kling 3.0: Kuaishou's Cinematic Video Generation.}
\label{subsec:kling}

Kling 3.0, launched by Kuaishou on 4 February 2026, represents the most capable iteration of the Kling video generation family. The model is built on the Multi-modal Visual Language (MVL) framework, an end-to-end architecture in which instruction understanding, visual generation, and refinement occur in a single shared embedding space rather than through sequential pipeline stages. The core innovation is 3D Spacetime Joint Attention combined with chain-of-thought reasoning, enabling the model to understand full 3D spatial and temporal relationships in a scene simultaneously---modelling object movement through space, lighting changes over time, and physical interactions including inertia, weight transfer, and collision detection. Kling Video 3.0 Omni generates synchronised audio and video natively within the same generation pass, extending maximum generation length to 15 seconds at up to 4K resolution and 60 frames per second. The AI Director feature enables multi-shot storyboarding with up to 6 shots within a single clip, maintaining spatial continuity and character consistency through the ``Director Memory'' system, which stores character references and vocal profiles in a dedicated context bank.

On the Artificial Analysis text-to-video leaderboard, Kling 3.0 1080p (Pro) leads with an ELO of 1248 in the without-audio category and 1097 in the with-audio category. Since its initial launch in June 2024, Kling AI has served over 60 million creators and produced more than 600 million videos, with partnerships spanning over 30,000 enterprise clients across the film and advertising industries.

% ============================================================================
\paragraph{Wan 2.5: Alibaba's Open-Source Video Foundation Model.}
\label{subsec:wan}

The Wan model series, developed by Alibaba's Wan AI team and released under the Apache 2.0 licence, has established itself as the leading open-source video generation family. Wan 2.1, released in February 2025, was trained on 1.5 billion videos and 10 billion images, achieving an overall VBench score of approximately 84.7\% and becoming the first video model capable of generating both Chinese and English text within video frames. The model is available in 14B and 1.3B parameter variants, with the smaller variant requiring only 8.19 GB VRAM and generating 480p video on a consumer RTX 4090 GPU. Wan 2.1 accumulated over 2.2 million downloads on Hugging Face and ModelScope.

Wan 2.5, the latest iteration, produces cinematic video with natively synchronised audio---including voice, sound effects, and lip-sync---in a single generation pass. On the Arena text-to-video leaderboard, Wan 2.5 achieved an ELO of 1305 at rank seven, the highest position among open-source video models and competitive with several proprietary systems. Wan 2.1-VACE (Video All-in-one Creation and Editing) further demonstrated the viability of unified open-source video generation and editing within a single model.

% ============================================================================
\paragraph{Hailuo 2.3: MiniMax's Video Generation.}
\label{subsec:hailuo}

Hailuo 2.3, developed by MiniMax and released on 28 October 2025, builds on the Hailuo 02 model's Noise-aware Compute Redistribution (NCR) framework, which boosts training and inference efficiency by 2.5 times at comparable parameter scale. The model's total parameter count was expanded to three times that of its predecessor, with training data volume increased by four times. Hailuo 2.3 achieves enhanced physics simulation including realistic water splashes, fabric movement, and object collisions, with particular strength in live-action facial performances and micro-expression rendering. The model supports native 1080p resolution and offers a Media Agent that automatically selects appropriate multi-modal generation models based on user intent. Upon its debut, Hailuo 02 reached second place on the Arena image-to-video leaderboard, even exceeding Veo 3 on several evaluation criteria. MiniMax has raised \$850 million in total funding from investors including Tencent, Alibaba, and miHoYo.

% ============================================================================
\paragraph{Seedance 1.5 and ByteDance.}
\label{subsec:seedance}

Seedance 1.5, developed by ByteDance alongside the Seedream image generation family, supports strong style transfer---applying illustration, oil painting, cartoon, and other reference aesthetics to generated video while preserving core content and motion. The model generates audio synchronised with visual content, including background music, ambient sound effects, and basic voice elements. Seedance 1.5 Pro was added to the Arena text-to-video leaderboard in January 2026. The subsequent Seedance 2.0 introduced director-level automation and multi-shot consistency, positioning ByteDance's video generation capabilities as a direct competitor to Sora 2 and Runway Gen-4.5.

% ============================================================================
\subsubsection{Europe and Rest of World}
\label{subsubsec:video_europe}

The video generation landscape outside the United States and China remains comparatively sparse at the frontier level, with no European or rest-of-world laboratory achieving a top-five position on the Arena text-to-video leaderboard as of March 2026. Open-source contributions have provided the most significant non-US, non-Chinese presence: LTX-2 Pro, an open-weight video model, leads among open-weight text-to-video models on the Artificial Analysis leaderboard with an ELO of 1133. The relative scarcity of European video generation contenders contrasts with the strong European presence in image generation (Black Forest Labs' FLUX 2) and text generation (Mistral AI), suggesting that the computational and data requirements for frontier video generation may present higher barriers to entry than other modalities.

% ============================================================================
\subsubsection{Comparative Analysis of Video Generation Models}
\label{subsec:video_comparison}

% ---- Table: Video Generation Model Comparison ----
\begin{table*}[!htbp]
\centering
\caption{Arena text-to-video rankings and key capabilities as of March 2026. ``Audio'' indicates native audio-visual co-generation. ``Max Duration'' reports the longest single-clip generation supported. Rankings combine data from the Arena (arena.ai) and Artificial Analysis leaderboards.}
\label{tab:video_comparison}
\small
\setlength{\tabcolsep}{4pt}
\renewcommand{\arraystretch}{1.12}
\begin{tabularx}{\textwidth}{@{}c l r l c r l@{}}
\toprule
Rank & Model & ELO & Developer & Audio & \makecell[r]{Max\\Duration} & Architecture \\
\midrule
\multicolumn{7}{@{}l}{\textit{United States}} \\[2pt]
1 & Veo 3.1 & 1386 & Google DeepMind & Yes & 60s+ & Latent Diffusion Transformer \\
1 & Sora 2 Pro & $\sim$1386 & OpenAI & Yes & 25s & Diffusion Transformer \\
3 & Sora 2 & --- & OpenAI & Yes & 25s & Diffusion Transformer \\
5 & Runway Gen-4.5 & $\sim$1225 & Runway & Yes & 60s & Diffusion Transformer \\[4pt]
\multicolumn{7}{@{}l}{\textit{China}} \\[2pt]
--- & Kling 3.0 Pro & 1248$^\dagger$ & Kuaishou & Yes & 15s & MVL + 3D Spacetime Attn. \\
7 & Wan 2.5 & 1305 & Alibaba & Yes & --- & Latent Diffusion Transformer \\
--- & Hailuo 2.3 & --- & MiniMax & No & 10s & NCR Framework \\
--- & Seedance 1.5 Pro & --- & ByteDance & Yes & --- & --- \\
\bottomrule
\multicolumn{7}{@{}l}{\footnotesize $^\dagger$Artificial Analysis leaderboard ELO (without-audio category); Arena ELO not separately reported.}
\end{tabularx}
\end{table*}

The video generation domain exhibits three defining competitive dynamics. First, native audio-visual co-generation has become the primary differentiator: the top three models on the Arena text-to-video leaderboard all produce synchronised audio alongside video, and user voting patterns strongly favour models with this capability. The technical challenge of jointly generating coherent audio and video within a single diffusion or autoregressive process has been addressed through unified latent representations that jointly model visual spacetime patches and temporal audio tokens, as demonstrated by Veo 3.1 and Sora 2.

Second, the competition between proprietary and open-source models is more stratified in video than in text or image generation. While Wan 2.5 from Alibaba achieved a competitive Arena ELO of 1305 as an open-source model, the gap to the top-ranked Veo 3.1 (ELO 1386) is larger than the corresponding gaps in the text and image leaderboards, suggesting that the computational and data requirements for frontier video generation continue to favour well-resourced laboratories.

Third, the rapid release cadence across all organisations---Google DeepMind progressing from Veo 2 to Veo 3.1 within fourteen months, Kuaishou releasing Kling 3.0 barely eight months after Kling 2.0, and OpenAI launching Sora 2 approximately eighteen months after the original Sora announcement---indicates that video generation remains in a phase of rapid capability expansion, with each generation introducing qualitatively new capabilities (audio co-generation, 4K resolution, multi-shot narratives, extended duration) rather than incremental improvements. The market is projected to grow from \$716.8 million in 2025 to \$2.56 billion by 2032.

\FloatBarrier

% ============================================================================
\subsection{Cross-Modality Meta-Analysis}
\label{subsec:meta_analysis}

The preceding subsections surveyed text, image, and video generation models individually. This subsection identifies cross-cutting patterns in geographic competitiveness, architectural scaling, and ecosystem structure through quantitative analysis of the data presented above. All figures in this subsection are derived exclusively from the Arena leaderboard ratings and model specifications reported in the preceding subsections.

% ---- Meta-analysis colours ----
\definecolor{metaus}{RGB}{63,81,181}
\definecolor{metachina}{RGB}{0,150,136}
\definecolor{metaeurope}{RGB}{230,126,34}
\definecolor{metagap}{RGB}{192,57,43}

% ---- Figure: Geographic Leadership by Modality ----
\begin{figure*}[t]
\centering
\begin{tikzpicture}
\begin{axis}[
    width=0.75\textwidth,
    height=7cm,
    ybar,
    bar width=16pt,
    enlarge x limits=0.3,
    ylabel={Top Arena ELO},
    ylabel style={font=\normalsize},
    symbolic x coords={Text, Image, Video},
    xtick=data,
    x tick label style={font=\normalsize},
    y tick label style={font=\small},
    ymin=1050,
    ymax=1560,
    ytick={1100, 1200, 1300, 1400, 1500},
    legend style={at={(0.5,0.97)}, anchor=north, legend columns=3, font=\small, draw=black!50, fill=white, fill opacity=0.9, text opacity=1},
    nodes near coords,
    nodes near coords style={font=\small, anchor=south},
    grid=major,
    grid style={gray!25},
    axis line style={gray!70},
]
\addplot[fill=metaus!70, draw=metaus] coordinates {(Text, 1504) (Image, 1264) (Video, 1386)};
\addplot[fill=metachina!70, draw=metachina] coordinates {(Text, 1451) (Image, 1147) (Video, 1305)};
\addplot[fill=metaeurope!70, draw=metaeurope] coordinates {(Text, 1418) (Image, 1168)};
\legend{United States, China, Europe}
\end{axis}
\end{tikzpicture}
\caption{Geographic leadership across generative AI modalities, showing the highest Arena ELO achieved by any model from each region. The United States leads all three categories, with the narrowest margin in text generation (53 ELO points over China) and the widest in image generation (117 points). Europe has no frontier-ranked video generation model. Data: Claude Opus 4.6 (US text), GLM-5 (China text), Mistral Large~3 (Europe text); GPT Image 1.5 (US image), Seedream 4.5 (China image), FLUX~2 Max (Europe image); Veo 3.1 (US video), Wan 2.5 (China video).}
\label{fig:meta_geographic}
\end{figure*}

% ---- Figure: US--China ELO Gap by Modality ----
\begin{figure}[t]
\centering
\begin{tikzpicture}
\begin{axis}[
    width=0.88\columnwidth,
    height=5.5cm,
    ybar,
    bar width=22pt,
    enlarge x limits=0.35,
    ylabel={ELO Gap (US $-$ China)},
    ylabel style={font=\small},
    symbolic x coords={Text, Video, Image},
    xtick=data,
    x tick label style={font=\small},
    y tick label style={font=\small},
    ymin=0,
    ymax=140,
    ytick={0,20,40,60,80,100,120,140},
    nodes near coords,
    nodes near coords style={font=\normalsize\bfseries, anchor=south},
    grid=major,
    grid style={gray!25},
    axis line style={gray!70},
]
\addplot[fill=metagap!55, draw=metagap!80!black] coordinates {(Text, 53) (Video, 81) (Image, 117)};
\end{axis}
\end{tikzpicture}
\caption{Arena ELO gap between the top-ranked United States and Chinese models by output modality, ordered from smallest to largest. Chinese laboratories have achieved near-parity in text generation (53 points: Claude Opus 4.6 at 1504 versus GLM-5 at 1451) but face larger deficits in video (81 points: Veo 3.1 versus Wan 2.5) and image generation (117 points: GPT Image 1.5 versus Seedream 4.5).}
\label{fig:meta_gap}
\end{figure}

% ---- Figure: Total Parameters vs Arena ELO (Open-Weight) ----
\begin{figure}[t]
\centering
\begin{tikzpicture}
\begin{axis}[
    width=0.92\columnwidth,
    height=6.5cm,
    xlabel={Total Parameters (billions)},
    ylabel={Arena Text ELO},
    xlabel style={font=\small},
    ylabel style={font=\small},
    x tick label style={font=\small},
    y tick label style={font=\small},
    xmin=0,
    xmax=1100,
    ymin=1310,
    ymax=1475,
    xtick={0,200,400,600,800,1000},
    ytick={1320,1360,1400,1440},
    grid=major,
    grid style={gray!25},
    axis line style={gray!70},
    legend style={at={(0.03,0.97)}, anchor=north west, font=\scriptsize, draw=black!50, fill=white},
    clip=false,
]
% China models
\addplot[only marks, mark=*, mark size=3.5pt, color=metachina, fill=metachina]
    coordinates {(1000, 1447) (744, 1451) (671, 1436) (685, 1421)};
% Europe models
\addplot[only marks, mark=square*, mark size=3pt, color=metaeurope, fill=metaeurope]
    coordinates {(675, 1418)};
% US models
\addplot[only marks, mark=triangle*, mark size=4pt, color=metaus, fill=metaus]
    coordinates {(400, 1417) (27, 1339)};
\legend{China, Europe, United States}
% Labels
\node[font=\scriptsize, anchor=west, xshift=3pt] at (axis cs:1000,1447) {Kimi K2.5};
\node[font=\scriptsize, anchor=east, xshift=-3pt] at (axis cs:744,1451) {GLM-5};
\node[font=\scriptsize, anchor=south west, xshift=2pt] at (axis cs:671,1436) {DeepSeek-R1};
\node[font=\scriptsize, anchor=north, yshift=-4pt] at (axis cs:685,1421) {V3.2};
\node[font=\scriptsize, anchor=south east, xshift=-2pt, yshift=2pt] at (axis cs:675,1418) {Mistral L3};
\node[font=\scriptsize, anchor=south west, xshift=2pt] at (axis cs:400,1417) {LLaMA 4 Mav.};
\node[font=\scriptsize, anchor=south west, xshift=2pt] at (axis cs:27,1339) {Gemma 3};
\end{axis}
\end{tikzpicture}
\caption{Total parameter count versus Arena text ELO for open-weight models with publicly disclosed architectures. Mixture-of-experts models (all points above 1400 ELO) cluster at 400B--1T total parameters with ELO ratings between 1417 and 1451, while the sole dense model (Gemma~3, 27B) scores substantially lower at 1339. Proprietary models (Claude, GPT, Gemini, Grok) are omitted as their architectures are undisclosed. Chinese models (circles) dominate the high-parameter, high-ELO region.}
\label{fig:meta_params_elo}
\end{figure}

% ---- Figure: Model Ecosystem Count by Region ----
\begin{figure}[t]
\centering
\begin{tikzpicture}
\begin{axis}[
    width=0.88\columnwidth,
    height=5.5cm,
    ybar,
    bar width=12pt,
    enlarge x limits=0.3,
    ylabel={Number of Model Families},
    ylabel style={font=\small},
    symbolic x coords={Text, Image, Video},
    xtick=data,
    x tick label style={font=\small},
    y tick label style={font=\small},
    ymin=0,
    ymax=15,
    ytick={0,3,6,9,12,15},
    legend style={at={(0.97,0.97)}, anchor=north east, font=\scriptsize, draw=black!50, fill=white},
    nodes near coords,
    nodes near coords style={font=\small, anchor=south},
    grid=major,
    grid style={gray!25},
    axis line style={gray!70},
]
\addplot[fill=metaus!70, draw=metaus] coordinates {(Text, 9) (Image, 4) (Video, 3)};
\addplot[fill=metachina!70, draw=metachina] coordinates {(Text, 12) (Image, 2) (Video, 4)};
\addplot[fill=metaeurope!70, draw=metaeurope] coordinates {(Text, 1) (Image, 1) (Video, 0)};
\legend{United States, China, Europe}
\end{axis}
\end{tikzpicture}
\caption{Number of distinct frontier model families surveyed in this section, by geographic origin and output modality. China fields more model families than the United States in text generation (12 versus 9) and video generation (4 versus 3), despite US models achieving higher peak Arena ELO scores (\Cref{fig:meta_geographic}). US text families: Claude, Gemini, Grok, GPT, LLaMA, Gemma, Phi, GPT-oss, Nemotron. China text families: GLM, Kimi, DeepSeek, MiniMax, Qwen, Step, MiMo, Ling, Hunyuan, InternLM, Yi, Baichuan. Europe contributes one family each in text (Mistral) and image (FLUX) but none in frontier video generation.}
\label{fig:meta_ecosystem}
\end{figure}

Three cross-cutting patterns emerge from this meta-analysis. First, the geographic competitiveness gap varies dramatically by modality (\Cref{fig:meta_geographic,fig:meta_gap}). In text generation, the top Chinese model (GLM-5, ELO 1451) trails the leading US model (Claude Opus 4.6, ELO 1504) by only 53 points---a gap that has narrowed substantially over the past year. In video and image generation, the gaps are 81 and 117 points respectively, suggesting that Chinese laboratories have prioritised text model development more aggressively than visual generation, or that the data and computational requirements for frontier visual models present higher barriers.

Second, the relationship between model scale and performance among open-weight models (\Cref{fig:meta_params_elo}) confirms that the mixture-of-experts architecture is a prerequisite for competitive Arena performance at the frontier: all open-weight models above 1400 ELO employ MoE with activation ratios between approximately 3\% and 6\%, while the best dense open-weight model (Gemma~3, 27B) scores 112 ELO points lower. Within the MoE cluster, however, total parameter count is not a reliable predictor of performance---GLM-5 (744B) outperforms the larger Kimi K2.5 ($\sim$1T), and LLaMA~4 Maverick (400B) achieves a similar ELO to Mistral Large~3 (675B), indicating that training methodology, data quality, and post-training alignment contribute as much as raw scale.

Third, the ecosystem breadth analysis (\Cref{fig:meta_ecosystem}) reveals an asymmetry between performance leadership and ecosystem depth. Despite US models achieving the highest ELO in every modality, China fields more distinct model families in text (12 versus 9) and video (4 versus 3). This breadth reflects the large number of well-funded Chinese AI laboratories (Zhipu AI, Moonshot AI, DeepSeek, MiniMax, Alibaba, ByteDance, Tencent, Xiaomi, Ant Group, StepFun, 01.AI, Baichuan) pursuing independent architectural strategies, creating a competitive ecosystem that accelerates innovation through diversity of approaches. Europe's contribution, while limited to two model families (Mistral and FLUX), is strategically significant given the regulatory and sovereignty considerations that favour European-originated models for deployment within the European Union.

% ============================================================================
% Section 4: Deployment Protocols and Inference Frameworks
% (Restructured — LLM-focused survey)
% ============================================================================

\section{Deployment Protocols and Inference Frameworks}
\label{sec:protocols}

Deploying large language models in production environments requires far more than a trained model and an inference endpoint. Practical systems must ground model outputs in external knowledge to mitigate hallucination, connect models to tools and services so that they can take actions beyond text generation, enable multiple agents to collaborate on complex tasks, and serve requests at scale with acceptable latency and cost. A layered ecosystem of protocols, standards, and frameworks has emerged to address each of these requirements. This section provides a comprehensive treatment of the major deployment protocols and inference frameworks, beginning with Retrieval-Augmented Generation for knowledge grounding, proceeding through the Model Context Protocol and the Agent-to-Agent protocol for tool use and inter-agent communication, examining function calling standards and agentic orchestration frameworks, and concluding with inference serving optimisations and safety guardrails.

% ============================================================================
\subsection{Retrieval-Augmented Generation}
\label{subsec:rag}

% ----------------------------------------------------------------------------
\subsubsection{The RAG Framework}
\label{subsubsec:rag_framework}

Retrieval-Augmented Generation (RAG) addresses the fundamental tension between the static nature of parametric knowledge and the dynamic nature of real-world information by conditioning the generation process on documents retrieved at inference time \cite{lewis2020retrieval}. Formally, given a user query $\inputx$, a retrieval function $\mathcal{R}$ selects a set of $k$ relevant passages $\data = \{d_1, \ldots, d_k\}$ from a document corpus $\mathcal{C}$, and the language model $\model$ generates an output $\outputy$ conditioned on both the query and the retrieved evidence. The marginal generation probability under this framework is
\begin{equation}
p(\outputy \mid \inputx) = \sum_{\data \subseteq \mathcal{C}} p(\outputy \mid \inputx, \data)\, p(\data \mid \inputx),
\label{eq:rag_marginal_sec4}
\end{equation}
which in practice is approximated by restricting $\data$ to the top-$k$ passages under the retrieval model. The two-stage decomposition into retrieval and generation offers three principal advantages: the external corpus can be updated without retraining the model, the retrieved passages provide verifiable provenance for generated claims, and smaller models can achieve competitive performance on knowledge-intensive tasks when equipped with effective retrieval. The retrieval stage itself typically employs a bi-encoder architecture in which a query encoder and a document encoder, both initialised from pre-trained Transformer models, produce dense vector representations whose inner product serves as a relevance score \cite{karpukhin2020dense}. Efficient nearest-neighbour search over large corpora is enabled by libraries such as FAISS, which provides GPU-accelerated implementations of inverted file indices with product quantisation and hierarchical navigable small world graphs \cite{johnson2019billion}.

The motivation for RAG extends beyond factual accuracy to encompass transparency and controllability. In a purely parametric model, the provenance of any particular claim is opaque, since the information is distributed across billions of parameters $\param$ without any mechanism to attribute outputs to specific training examples. RAG systems, by contrast, can return the source passages alongside the generated answer, enabling users to verify claims against the original documents. This property is particularly valuable in high-stakes domains such as healthcare, legal analysis, and financial services, where the ability to trace an assertion to an authoritative source is a prerequisite for trust and regulatory compliance.

% ----------------------------------------------------------------------------
\subsubsection{Advanced RAG Patterns}
\label{subsubsec:advanced_rag}

The basic single-retrieval paradigm, sometimes termed naive RAG, is insufficient for complex queries that require synthesising evidence from multiple sources or reasoning over chains of facts. Multi-hop RAG addresses this limitation through iterative retrieval: the system retrieves an initial set of passages, generates an intermediate reasoning step, uses that intermediate result to formulate a refined query, and retrieves again \cite{borgeaud2022improving}. This cycle repeats until sufficient evidence has been gathered to answer the original question. Self-RAG introduces a mechanism by which the language model itself decides whether retrieval is necessary for a given query, generating special reflection tokens that indicate whether external evidence is required, whether the retrieved passages are relevant, and whether the generated response is supported by the evidence. Corrective RAG adds a verification layer that evaluates the relevance and quality of retrieved passages before they are passed to the generator, discarding those that fall below a confidence threshold and triggering additional retrieval with reformulated queries when necessary. GraphRAG represents a paradigm shift from flat document retrieval to structured knowledge retrieval, constructing a knowledge graph from the document corpus and performing retrieval over graph neighbourhoods rather than isolated text passages. Agentic RAG delegates the retrieval strategy itself to an autonomous agent that can choose among multiple retrieval backends, reformulate queries, and iteratively refine its search strategy based on intermediate results.

Hybrid search combines the complementary strengths of dense retrieval and sparse lexical matching by querying both a dense embedding index and a BM25 index in parallel and merging the results through reciprocal rank fusion \cite{izacard2022atlas}. This approach is particularly effective for queries containing rare technical terms or proper nouns that may not be well represented in the dense embedding space but are easily matched by lexical methods. Re-ranking further improves retrieval quality by applying a computationally expensive cross-encoder model that processes the concatenation of query and candidate passage through a single Transformer, enabling fine-grained token-level interactions that yield more accurate relevance judgments than the bi-encoder's independent encoding. The combination of hybrid retrieval, re-ranking, and iterative refinement constitutes the current state of the art in production RAG systems, achieving substantially higher answer accuracy and source attribution fidelity than naive single-pass retrieval.

% ----------------------------------------------------------------------------
\subsubsection{Practical RAG Implementation}
\label{subsubsec:rag_practical}

A production RAG pipeline begins with document ingestion, during which source documents in formats such as PDF, HTML, and plain text are parsed, cleaned, and converted to a uniform textual representation. The resulting text is then segmented into chunks according to a chosen chunking strategy. Fixed-size chunking divides documents into segments of a predetermined number of tokens, typically 512, with an overlap of approximately 50 tokens between consecutive segments to ensure that information near chunk boundaries is captured in at least two adjacent segments. Semantic chunking determines split points based on content rather than fixed counts, computing sentence embeddings for consecutive sentences and splitting at locations where the cosine similarity between adjacent embeddings falls below a threshold, thereby producing chunks that align with the natural topical structure of the document. Recursive splitting attempts to split text using a hierarchy of separators, first trying paragraph boundaries, then sentence boundaries, and finally character-level splits, preferring larger semantic units whenever possible. Each chunk is then passed through an embedding model to produce a dense vector representation, and the resulting embeddings are stored in a vector database such as Pinecone, Weaviate, Chroma, or Qdrant.

At inference time, the user query is embedded using the same encoder, and the vector database is queried to retrieve the $k$ most similar passages. An optional re-ranking stage applies a cross-encoder to refine the ordering. The top passages are then assembled into a context window together with the original query and a system prompt that instructs the model to ground its answer in the provided evidence and to cite specific passages. The language model generates the final response, ideally with inline citations that reference the source documents. This end-to-end pipeline transforms a raw document collection into a queryable knowledge system that provides grounded, verifiable answers while accommodating new documents through incremental indexing without retraining the underlying model.

% ---- Figure: RAG Pipeline ----
\definecolor{ragdoc}{RGB}{41,98,255}
\definecolor{ragchunk}{RGB}{0,150,80}
\definecolor{ragembed}{RGB}{142,68,173}
\definecolor{ragvdb}{RGB}{230,126,34}
\definecolor{ragquery}{RGB}{192,57,43}
\definecolor{ragret}{RGB}{0,172,193}
\definecolor{ragrerank}{RGB}{155,89,182}
\definecolor{ragctx}{RGB}{255,152,0}
\definecolor{ragllm}{RGB}{76,175,80}
\definecolor{ragans}{RGB}{44,62,80}

\begin{figure*}[t]
\centering
\begin{tikzpicture}[
    node distance=0.7cm and 0.45cm,
    box/.style={
        rectangle,
        rounded corners=5pt,
        draw=#1!70,
        fill=#1!12,
        line width=1.2pt,
        minimum height=1.1cm,
        minimum width=1.9cm,
        align=center,
        font=\footnotesize\sffamily
    },
    arrow/.style={
        -{Stealth[length=6pt, width=5pt]},
        line width=1.2pt,
        #1
    },
    lbl/.style={
        font=\scriptsize\sffamily,
        midway,
        fill=white,
        inner sep=2pt
    }
]

% === Indexing pipeline (top row) ===
\node[box=ragdoc] (docs) {Documents\\{\scriptsize(PDF, HTML, TXT)}};
\node[box=ragchunk, right=of docs] (chunk) {Chunk\\{\scriptsize(512 tokens)}};
\node[box=ragembed, right=of chunk] (embed) {Embed\\{\scriptsize(Bi-encoder)}};
\node[box=ragvdb, right=of embed] (vdb) {Vector DB\\{\scriptsize(FAISS, Pinecone)}};

\draw[arrow=ragdoc!70] (docs) -- (chunk) node[lbl, above] {parse};
\draw[arrow=ragchunk!70] (chunk) -- (embed) node[lbl, above] {segments};
\draw[arrow=ragembed!70] (embed) -- (vdb) node[lbl, above] {vectors};

\node[above=0.35cm of chunk, font=\footnotesize\sffamily\bfseries, text=deepblue] {Offline Indexing Pipeline};

% === Query pipeline (bottom row) ===
\node[box=ragquery, below=2.0cm of docs] (query) {User\\Query};
\node[box=ragembed, right=of query] (qembed) {Query\\Embedding};
\node[box=ragret, right=of qembed] (retrieve) {Retrieve\\{\scriptsize(Top-$k$)}};
\node[box=ragrerank, right=of retrieve] (rerank) {Re-rank\\{\scriptsize(Cross-enc.)}};
\node[box=ragctx, right=of rerank] (ctx) {Assemble\\Context};
\node[box=ragllm, right=of ctx] (llm) {LLM\\Generation};
\node[box=ragans, right=of llm] (ans) {Grounded\\Answer};

\draw[arrow=ragquery!70] (query) -- (qembed) node[lbl, above] {encode};
\draw[arrow=ragembed!70] (qembed) -- (retrieve) node[lbl, above] {search};
\draw[arrow=ragret!70] (retrieve) -- (rerank) node[lbl, above] {passages};
\draw[arrow=ragrerank!70] (rerank) -- (ctx) node[lbl, above] {top-$n$};
\draw[arrow=ragctx!70] (ctx) -- (llm) node[lbl, above] {prompt};
\draw[arrow=ragllm!70] (llm) -- (ans) node[lbl, above] {response};

% Connection from VDB to retrieval
\draw[arrow=ragvdb!70] (vdb) -- (retrieve) node[lbl, right] {$k$-NN};

% Connection from query to context assembly
\draw[arrow=ragquery!70, densely dashed] (query.south) -- ++(0,-0.5) -| (ctx.south) node[lbl, below, pos=0.25] {original query};

\node[above=0.35cm of ctx, font=\footnotesize\sffamily\bfseries, text=deepblue] {Online Inference Pipeline};

\end{tikzpicture}
\caption{End-to-end RAG pipeline. The offline stage (top) indexes documents via chunking, embedding, and vector storage. The online stage (bottom) retrieves the top-$k$ passages, re-ranks them, and generates a grounded response.}
\label{fig:rag_pipeline_sec4}
\end{figure*}

% ============================================================================
\subsection{The Model Context Protocol}
\label{subsec:mcp}

% ----------------------------------------------------------------------------
\subsubsection{Architecture and Core Primitives}
\label{subsubsec:mcp_architecture}

The Model Context Protocol (MCP) is an open protocol introduced by Anthropic that standardises the connection between AI applications and external data sources, tools, and services \cite{anthropic2024mcp}. MCP establishes a universal interface through which any AI application can communicate with any external capability provider, playing a role analogous to that of HTTP in standardising web communication or USB in standardising peripheral connectivity. The protocol defines a client-server architecture with three principal roles. The MCP Host is the AI application in which the language model operates, such as a conversational assistant, an integrated development environment with AI capabilities, or an autonomous agent framework. Within the host, one or more MCP Clients serve as protocol connectors that maintain persistent connections to MCP Servers. Each client manages the communication lifecycle with a single server, handling connection establishment, capability discovery, request serialisation, and response deserialisation. The MCP Server is an external process or service that exposes a set of capabilities to the AI application through the standardised protocol interface. Communication between clients and servers employs the JSON-RPC 2.0 transport protocol, which provides a lightweight, language-agnostic mechanism for remote procedure calls in which each request is a JSON object containing a method name, a structured parameter object, and a unique identifier, and the corresponding response contains either a result object or an error object.

The MCP specification defines three core primitives that servers can expose to clients. Resources represent data sources that the model can read, identified by URIs and capable of representing files, database records, API responses, or any other form of structured or unstructured data. Tools represent executable functions that the model can invoke to perform actions or computations, with each tool defined by a name, a natural language description that enables the model to understand when invocation is appropriate, and a JSON Schema specifying the structure and types of input parameters. Prompts are reusable templates for common interaction patterns that servers provide to guide the model's use of available tools and resources. These three primitives collectively enable a rich vocabulary of model-to-external-world interactions while maintaining a clean separation of concerns between the AI application and the capability providers. The open nature of the protocol means that any developer can implement an MCP server for any service, and any compliant host application can consume it, creating a shared ecosystem of reusable integrations that is independent of any particular model provider.

% ----------------------------------------------------------------------------
\subsubsection{MCP Implementation Examples}
\label{subsubsec:mcp_examples}

The versatility of MCP is best illustrated through concrete deployment scenarios. Consider a database MCP server that exposes a tool named ``execute\_sql\_query'' with a description stating that it executes a read-only SQL query against a company's sales database. When a business analyst asks the AI application ``What were our total sales by region last month?'', the language model determines that this question requires structured data retrieval, formulates an appropriate SQL query that selects the region and sum of sales amount from the orders table filtered by the previous month's date range, and invokes the tool. The MCP client serialises this as a JSON-RPC 2.0 request, for example \texttt{\{"jsonrpc":"2.0", "method":"tools/call", "params":\{"name":"execute\_sql\_query", "arguments":\{"query":"SELECT region, SUM(amount) ..."\}\}, "id":1\}}, and transmits it to the server. The server validates the query against a read-only constraint, executes it against the PostgreSQL database, and returns the result set in a JSON-RPC response. The model then synthesises a natural language summary of the sales figures for the analyst. A file system MCP server operates similarly, exposing tools for reading files, searching directories, and listing contents, enabling the model to assist developers with tasks such as locating configuration files or reading logs to diagnose errors.

A third deployment scenario addresses the knowledge cutoff problem through a web search MCP server. This server exposes a ``web\_search'' tool that accepts a query string, submits it to a search engine API, and returns results with titles, URLs, and text snippets. When a user asks about an event that occurred after the model's training data cutoff, the model recognises that its parametric knowledge is insufficient, invokes the web search tool, receives current results, and synthesises an answer grounded in the retrieved information. The complete request-response cycle for any of these scenarios follows a uniform pattern: the host receives the user query, the model decides to invoke a tool, the client serialises the invocation as a JSON-RPC request, the server executes the underlying operation, the result flows back through the same chain, and the model incorporates the returned data into its generation context to produce a seamless natural language response. This uniformity is the central contribution of MCP: a tool server written once can be consumed by any compliant host application, regardless of the underlying model provider, eliminating the need for provider-specific integration code.

% ---- Figure: MCP Architecture ----
\definecolor{mcphost}{RGB}{41,98,255}
\definecolor{mcpclient}{RGB}{142,68,173}
\definecolor{mcpserver}{RGB}{0,150,80}
\definecolor{mcpdb}{RGB}{192,57,43}
\definecolor{mcpfs}{RGB}{0,172,193}
\definecolor{mcpweb}{RGB}{255,152,0}

\begin{figure*}[t]
\centering
\begin{tikzpicture}[
    node distance=0.8cm and 1.2cm,
    box/.style={
        rectangle,
        rounded corners=4pt,
        draw=#1!70,
        fill=#1!12,
        line width=1pt,
        minimum height=1.2cm,
        minimum width=2.4cm,
        align=center,
        font=\small\sffamily
    },
    smallbox/.style={
        rectangle,
        rounded corners=3pt,
        draw=#1!60,
        fill=#1!8,
        line width=0.7pt,
        minimum height=0.9cm,
        minimum width=1.6cm,
        align=center,
        font=\scriptsize\sffamily
    },
    arrow/.style={
        -{Stealth[length=5pt, width=4pt]},
        line width=0.9pt,
        #1
    },
    biarrow/.style={
        {Stealth[length=5pt, width=4pt]}-{Stealth[length=5pt, width=4pt]},
        line width=0.9pt,
        #1
    },
    lbl/.style={
        font=\scriptsize\sffamily,
        fill=white,
        inner sep=2pt
    },
    container/.style={
        rectangle,
        rounded corners=6pt,
        draw=#1!50,
        fill=#1!4,
        line width=1.2pt,
        inner sep=12pt
    }
]

% Host container
\node[container=mcphost, minimum width=4.5cm, minimum height=3.2cm] (hostbox) at (0,0) {};
\node[above=2pt of hostbox.north, font=\small\sffamily\bfseries, text=mcphost] {MCP Host};

% LLM inside host
\node[box=mcphost, minimum width=2.0cm] (llm) at (0, 0.5) {LLM};

% Client inside host
\node[smallbox=mcpclient, minimum width=2.0cm] (client) at (0, -0.7) {MCP Client};

% Arrow LLM to Client
\draw[biarrow=mcpclient!70] (llm) -- (client);

% Servers
\node[box=mcpserver, right=2.5cm of hostbox.east, yshift=1.0cm, minimum width=1.8cm] (s1) {Server\\(Database)};
\node[box=mcpserver, right=2.5cm of hostbox.east, yshift=0cm, minimum width=1.8cm] (s2) {Server\\(Filesystem)};
\node[box=mcpserver, right=2.5cm of hostbox.east, yshift=-1.0cm, minimum width=1.8cm] (s3) {Server\\(Web API)};

% JSON-RPC arrows from client to servers
\draw[biarrow=mcpserver!70] (client.east) ++(0.5,0) |- (s1.west);
\draw[biarrow=mcpserver!70] (client.east) ++(0.5,0) -- (s2.west);
\draw[biarrow=mcpserver!70] (client.east) ++(0.5,0) |- (s3.west);

% JSON-RPC label
\node[lbl, above=0pt] at ($(client.east)!0.5!(s2.west)$) {JSON-RPC 2.0};

% Server capability labels
\node[font=\tiny\sffamily\itshape, text=mcpserver!70, below=1pt of s1.south] {tools, resources, prompts};
\node[font=\tiny\sffamily\itshape, text=mcpserver!70, below=1pt of s2.south] {tools, resources, prompts};
\node[font=\tiny\sffamily\itshape, text=mcpserver!70, below=1pt of s3.south] {tools, resources, prompts};

% External resources
\node[smallbox=mcpdb, right=0.6cm of s1, minimum width=1.3cm] (db) {PostgreSQL};
\node[smallbox=mcpfs, right=0.6cm of s2, minimum width=1.3cm] (fs) {Local FS};
\node[smallbox=mcpweb, right=0.6cm of s3, minimum width=1.3cm] (web) {REST API};

\draw[arrow=mcpdb!70] (s1) -- (db);
\draw[arrow=mcpfs!70] (s2) -- (fs);
\draw[arrow=mcpweb!70] (s3) -- (web);

\end{tikzpicture}
\caption{Architecture of the Model Context Protocol (MCP). The MCP Host contains the language model and an MCP Client, which maintains JSON-RPC 2.0 connections to multiple MCP Servers. Each server exposes tools, resources, and prompts while connecting to specific external services such as databases, filesystems, or web APIs.}
\label{fig:mcp_arch_sec4}
\end{figure*}

% ============================================================================
\subsection{The Agent-to-Agent Protocol}
\label{subsec:a2a}

The Agent-to-Agent (A2A) protocol, introduced by Google in 2025, addresses a complementary challenge to that solved by MCP: whereas MCP standardises the connection between an agent and its tools, A2A standardises communication between AI agents built on different frameworks, enabling them to discover one another's capabilities and collaborate on complex tasks \cite{google2025a2a}. The protocol defines three key concepts. Agent Cards serve as a discovery mechanism through which each agent publishes a machine-readable description of its capabilities, input and output modalities, and authentication requirements, analogous to a service descriptor in microservice architectures. Tasks represent units of work with a defined lifecycle, progressing through states such as submitted, working, input-required, completed, canceled, and failed, and carrying structured payloads that can include text, files, and structured data. Messages are communication turns between a client and a remote agent, each with a role (user or agent) containing one or more Parts, supporting both synchronous request-response patterns and asynchronous streaming via Server-Sent Events for long-running operations.

The relationship between MCP and A2A is one of complementarity rather than competition. MCP operates at the agent-to-tool level, providing the mechanism through which an individual agent accesses databases, file systems, APIs, and other external services. A2A operates at the agent-to-agent level, providing the mechanism through which multiple agents coordinate their activities. A practical illustration of this complementarity arises in multi-agent workflows: a planning agent might receive a high-level objective from a user, decompose it into sub-tasks, and delegate each sub-task to a specialised agent via A2A. A research agent might use MCP to access web search and document retrieval tools to gather information, while a coding agent uses MCP to access a code execution environment and file system tools to implement a solution. The planning agent orchestrates the overall workflow through A2A, collecting results from each specialised agent and synthesising them into a coherent response. This layered architecture separates the concerns of tool access, inter-agent communication, and task orchestration, enabling each layer to evolve independently and allowing agents built with different frameworks and running on different infrastructure to interoperate seamlessly.

% ============================================================================
\subsection{Function Calling and Tool Use Standards}
\label{subsec:function_calling}

The practical implementation of tool use varies across major language model providers, reflecting different design philosophies and API conventions, yet a common pattern has emerged. OpenAI introduced function calling in its GPT-3.5 and GPT-4 APIs, allowing developers to define functions with a name, a natural language description, and a JSON Schema specifying parameter types and constraints. When the model determines that a function call is appropriate, it generates a structured JSON object containing the function name and arguments, which the application executes and returns to the model. OpenAI subsequently extended this to parallel function calling, in which the model can request multiple function invocations in a single generation step, and to structured outputs, which guarantee that the generated JSON conforms to the provided schema. Anthropic's tool use implementation in the Claude family follows a similar declarative pattern, with tools defined by name, description, and input schema, but places particular emphasis on the model articulating its reasoning before each invocation, providing transparency into the decision-making process. Google's function calling in the Gemini family integrates tool use with multimodal capabilities, allowing function calls to be triggered by text, image, or video inputs, and extends the paradigm to support grounding in Google Search and code execution as built-in tools.

Despite the surface-level variation, these implementations have converged towards a common architectural pattern: tools are defined declaratively with a name, description, and parameter schema; the model autonomously decides when a tool call is appropriate based on the user query and the tool descriptions; the model generates a structured JSON output specifying the tool name and arguments; the application executes the call and returns the result; and the model incorporates the result into its generation context \cite{qin2024tool}. MCP aims to standardise this pattern at the protocol level, providing a single, provider-agnostic interface that eliminates the need for provider-specific integration code. An MCP server exposing a database query tool does not need to be rewritten when an application switches from one model provider to another; the tool definition, transport protocol, and request-response format remain unchanged. This interoperability reduces development cost, encourages the creation of a shared ecosystem of reusable tool servers, and prevents vendor lock-in at the tool integration layer.

% ============================================================================
\subsection{Agentic Frameworks and Orchestration}
\label{subsec:agentic}

% ----------------------------------------------------------------------------
\subsubsection{ReAct and Chain-of-Thought}
\label{subsubsec:react}

The ReAct framework formalises the integration of reasoning and acting in a unified generation loop by interleaving three types of outputs: thoughts, actions, and observations \cite{yao2023react}. A thought is an internal reasoning step in which the model analyses the current state of the task, decomposes a complex question, or plans the next action. An action is a tool invocation, such as a database query, a web search, or a file read. An observation is the result returned by the tool, which is appended to the model's context for subsequent reasoning. This cycle of thought, action, and observation repeats until the model determines that sufficient information has been gathered to produce a final answer. The explicit generation of reasoning traces provides interpretability, as the chain of thoughts documents the model's decision-making process and can be inspected to diagnose errors or understand the agent's strategy. Chain-of-thought prompting complements ReAct by encouraging the model to decompose complex tasks into intermediate reasoning steps before deciding which tool to invoke, leading to more precise tool invocations and better synthesis of results into coherent answers \cite{wei2022chain}. Without explicit reasoning, a model might attempt to answer a multi-step question with a single, poorly formulated tool call; with chain-of-thought prompting, the model articulates each step of its reasoning, producing a more structured and reliable problem-solving trajectory.

% ----------------------------------------------------------------------------
\subsubsection{LangChain and LlamaIndex}
\label{subsubsec:langchain}

LangChain provides a composable framework for building applications powered by language models, organised around several core abstractions \cite{langchain2023}. Chains represent sequences of operations, such as retrieving documents, formatting a prompt, invoking a model, and parsing the output, that are composed into reusable pipelines. Agents extend chains with autonomous decision-making, allowing the model to choose which tools to invoke and in what order based on the current state of the task. The LangChain Expression Language (LCEL) provides a declarative syntax for composing these operations into complex pipelines with built-in support for streaming, batching, and fallback handling. LlamaIndex focuses specifically on the data ingestion, indexing, and querying aspects of RAG applications, providing connectors for diverse data sources, multiple indexing strategies including vector indices, keyword indices, and knowledge graph indices, and query engines that support both simple retrieval and complex multi-step reasoning over indexed data \cite{llamaindex2023}. Both frameworks are available as Python and JavaScript libraries and have become widely adopted in production RAG and agentic applications, providing the middleware layer between raw model APIs and end-user applications.

% ----------------------------------------------------------------------------
\subsubsection{Multi-Agent Systems}
\label{subsubsec:multi_agent}

Multi-agent systems extend the single-agent paradigm by deploying multiple language model instances that collaborate to accomplish complex tasks \cite{wang2024survey}. Frameworks such as AutoGen, CrewAI, and LangGraph provide the orchestration infrastructure for defining agent roles, communication patterns, and task decomposition strategies. The common architectural pattern involves multiple specialised agents, each with a defined role such as researcher, coder, or reviewer, that collaborate on a shared objective. A researcher agent might gather information through web search and document retrieval, a coding agent might implement solutions using a code execution environment, and a reviewer agent might evaluate the quality and correctness of the outputs. Two principal orchestration architectures have emerged: supervisor architectures, in which a central planning agent decomposes the objective into sub-tasks and delegates each to a specialised worker, and peer architectures, in which agents communicate directly with one another through shared message channels and collectively negotiate task allocation. Supervisor architectures offer clearer control flow and easier debugging, while peer architectures can be more flexible and resilient to individual agent failures. The A2A protocol provides a natural substrate for multi-agent communication in heterogeneous environments, enabling agents built with different frameworks and running on different infrastructure to collaborate through a standardised interface.

% ============================================================================
\subsection{Inference Serving and Optimisation}
\label{subsec:inference}

% ----------------------------------------------------------------------------
\subsubsection{vLLM and PagedAttention}
\label{subsubsec:vllm}

Serving large language models at scale presents a fundamental memory management challenge: the key-value (KV) cache that stores the attention states for each token in the sequence grows linearly with sequence length and batch size, and naive allocation strategies waste substantial memory through fragmentation. vLLM addresses this challenge through PagedAttention, an attention algorithm that manages the KV cache using a paging mechanism inspired by virtual memory in operating systems \cite{kwon2023vllm}. Rather than allocating a contiguous block of memory for each sequence's KV cache, PagedAttention divides the cache into fixed-size blocks (pages) that can be allocated non-contiguously and mapped to logical sequence positions through a page table. This approach eliminates internal fragmentation, enables efficient memory sharing between sequences that share common prefixes (such as the system prompt), and allows the system to serve significantly more concurrent requests within the same memory budget. vLLM further implements continuous batching, which dynamically adds new requests to a running batch as existing requests complete, rather than waiting for all requests in a batch to finish before starting the next batch. The combination of PagedAttention and continuous batching yields throughput improvements of two to four times over naive serving implementations, making vLLM the most widely adopted open-source inference engine for production LLM deployments.

% ----------------------------------------------------------------------------
\subsubsection{SGLang and Structured Generation}
\label{subsubsec:sglang}

SGLang provides a domain-specific language for efficient execution of structured LLM programs, addressing the observation that many practical applications involve complex interaction patterns with multiple model calls sharing common prefixes \cite{zheng2024sglang}. The system introduces RadixAttention, which organises the KV cache as a radix tree indexed by token sequences, enabling automatic detection and reuse of cached computations across requests that share common prefixes. When multiple requests begin with the same system prompt or share a common conversational context, RadixAttention avoids redundant computation by serving subsequent requests from the cached prefix, substantially reducing latency and increasing throughput. SGLang also provides primitives for constrained decoding that guarantee the model's output conforms to a specified format, such as a JSON schema or a regular expression, by masking logits corresponding to tokens that would violate the constraint at each generation step. This capability is particularly valuable for function calling and structured output generation, where the model must produce valid JSON that conforms to a tool's parameter schema.

% ----------------------------------------------------------------------------
\subsubsection{Speculative Decoding}
\label{subsubsec:speculative}

Autoregressive generation is inherently sequential, as each token depends on all preceding tokens, and this sequential nature makes it difficult to fully utilise the parallel computation capacity of modern accelerators. Speculative decoding addresses this bottleneck by using a small, fast draft model to generate a sequence of candidate tokens, which are then verified in parallel by the large target model \cite{leviathan2023fast}. The draft model generates $\gamma$ candidate tokens autoregressively, and the target model evaluates the probability of each candidate token conditioned on its prefix in a single forward pass. Each candidate token is accepted with probability $\min(1, p_{\mathrm{target}}(x)/p_{\mathrm{draft}}(x))$, a stochastic criterion that ensures the accepted sequence has exactly the same distribution as if the target model had generated it autoregressively. When a candidate token is rejected, the target model samples a correction token and all subsequent candidates are discarded. Because the draft model is much faster than the target model and the verification step processes all candidates in parallel, speculative decoding achieves speedups of two to three times with no change in output quality, as the acceptance criterion guarantees identical output distributions.

% ----------------------------------------------------------------------------
\subsubsection{Local Deployment: Ollama, llama.cpp, and GGUF}
\label{subsubsec:local}

The democratisation of large language model access has been significantly advanced by tools that enable efficient local deployment on consumer hardware. The llama.cpp project provides a high-performance inference engine written in C and C++ that supports execution on both CPUs and GPUs, with particular emphasis on quantised model formats that reduce memory requirements while preserving acceptable output quality \cite{gguf2024}. The GGUF file format, developed as part of this ecosystem, provides a standardised container for quantised model weights that supports multiple quantisation levels, from 2-bit to 8-bit, enabling users to trade off model quality against memory consumption and inference speed according to their hardware constraints. Ollama builds upon llama.cpp to provide a user-friendly command-line interface and local API server that simplifies the process of downloading, managing, and running language models on personal machines \cite{ollama2024}. A user can deploy a quantised version of a model with a single command, obtaining a local inference endpoint that requires no internet connection, incurs no API costs, and keeps all data on the local machine. This local deployment ecosystem has made it possible for individual researchers, small organisations, and privacy-sensitive applications to leverage the capabilities of large language models without dependence on cloud-hosted API services, representing a significant shift in the accessibility of advanced AI capabilities.

% ============================================================================
\subsection{Safety Guardrails and Content Filtering}
\label{subsec:guardrails}

Deploying language models in production environments requires mechanisms to ensure that model outputs conform to safety, policy, and quality requirements. NeMo Guardrails provides a programmable framework for defining input and output rails that filter, validate, and constrain model behaviour at inference time \cite{rebedea2023nemo}. Input rails can classify incoming queries and block or redirect those that fall outside the permitted topic scope, preventing the model from engaging with harmful, off-topic, or adversarial prompts. Output rails can validate the model's response against factual accuracy checks, content policy constraints, and format requirements before the response is returned to the user. Topic control rails restrict the model to a predefined set of conversational domains, ensuring that a customer service assistant, for example, does not engage in discussions about politics or provide medical advice. The rails are defined in a declarative configuration language that specifies the conditions under which each rail is triggered and the action to take, such as blocking the request, substituting a canned response, or re-prompting the model with additional constraints.

Constitutional AI provides a complementary approach to safety by embedding behavioural principles directly into the model through the training process rather than applying external filters at inference time \cite{bai2022constitutional}. In this paradigm, a set of constitutional principles, such as ``be helpful, harmless, and honest,'' are used to guide a self-critique and revision process during training: the model generates a response, critiques its own output against the constitutional principles, revises the response to better align with those principles, and the revised responses are used as training data for reinforcement learning from AI feedback. The challenge of balancing safety with utility remains an active area of research, as overly aggressive filtering can render the model unhelpful by refusing benign requests, while insufficient filtering exposes users and organisations to reputational and legal risk. Production deployments typically combine multiple layers of defence, including constitutional training, input classification, output validation, and human-in-the-loop review for high-stakes decisions, to achieve an appropriate balance between safety and usefulness.

% ---- Figure: Protocol Stack ----
\definecolor{stackinf}{RGB}{41,98,255}
\definecolor{stacksafe}{RGB}{192,57,43}
\definecolor{stacktool}{RGB}{0,150,80}
\definecolor{stackproto}{RGB}{142,68,173}
\definecolor{stackagent}{RGB}{230,126,34}

\begin{figure}[t]
\centering
\begin{tikzpicture}[
    layer/.style={
        rectangle,
        rounded corners=4pt,
        draw=#1!70,
        fill=#1!12,
        line width=1.2pt,
        minimum height=1.1cm,
        minimum width=6.8cm,
        align=center,
        font=\small\sffamily
    }
]

% Stack layers (bottom to top)
\node[layer=stackinf] (L1) at (0, 0) {Inference Serving\\{\scriptsize vLLM, SGLang, llama.cpp, Ollama}};
\node[layer=stacksafe] (L2) at (0, 1.5) {Safety and Guardrails\\{\scriptsize NeMo Guardrails, Constitutional AI}};
\node[layer=stacktool] (L3) at (0, 3.0) {Tool Use and Function Calling\\{\scriptsize OpenAI, Anthropic, Google standards}};
\node[layer=stackproto] (L4) at (0, 4.5) {Protocols\\{\scriptsize MCP (agent$\leftrightarrow$tool), A2A (agent$\leftrightarrow$agent)}};
\node[layer=stackagent] (L5) at (0, 6.0) {Agentic Frameworks\\{\scriptsize ReAct, LangChain, LlamaIndex, AutoGen}};

% Arrows between layers
\draw[-{Stealth[length=5pt, width=4pt]}, line width=1pt, gray!60] (L1.north) -- (L2.south);
\draw[-{Stealth[length=5pt, width=4pt]}, line width=1pt, gray!60] (L2.north) -- (L3.south);
\draw[-{Stealth[length=5pt, width=4pt]}, line width=1pt, gray!60] (L3.north) -- (L4.south);
\draw[-{Stealth[length=5pt, width=4pt]}, line width=1pt, gray!60] (L4.north) -- (L5.south);

% Side labels
\node[font=\tiny\sffamily, text=gray, rotate=90, anchor=south] at (-4.0, 3.0) {Increasing Abstraction};
\draw[-{Stealth[length=4pt, width=3pt]}, line width=0.6pt, gray!40] (-3.75, 0.5) -- (-3.75, 5.5);

\end{tikzpicture}
\caption{Layered architecture of the LLM deployment stack. Inference serving engines form the foundation, upon which safety guardrails, tool use standards, communication protocols, and agentic frameworks are progressively layered to enable autonomous, safe, and interoperable AI systems.}
\label{fig:protocol_stack}
\end{figure}

% ============================================================================
% Section 5: Applications, Ethics, and Conclusion
% ============================================================================

\section{Applications of Generative Artificial Intelligence}
\label{sec:applications}

The preceding sections of this survey have examined the mathematical foundations, architectural designs, training methodologies, and deployment protocols that collectively define the generative AI landscape. This section turns to the practical manifestations of these technologies, surveying their adoption and measured impact across fifteen major domains: business process automation, financial services, tourism and hospitality, legal services, manufacturing and supply chain management, real estate, media and entertainment, agriculture and environmental sciences, government and public administration, telecommunications, energy and utilities, retail and e-commerce, insurance and risk management, healthcare, and education. In each domain, the discussion draws upon empirical studies, quantitative benchmarks, and documented case studies to assess the extent to which generative AI is reshaping established practices. The breadth of these applications underscores the characterisation of large language models as foundation models \cite{bommasani2021opportunities}, wherein a single pre-trained system can be adapted to serve an extraordinarily diverse array of downstream tasks spanning virtually every sector of the knowledge economy.

% ============================================================================
\subsection{Business Process Automation and Enterprise Applications}
\label{subsec:business_applications}

% ----------------------------------------------------------------------------
\subsubsection{Customer Service and Conversational Agents}
\label{subsubsec:customer_service}

Generative AI has fundamentally transformed the customer service function within enterprises by enabling the deployment of intelligent conversational agents capable of handling complex, multi-turn interactions in natural language. Unlike the rule-based chatbots that dominated earlier approaches, LLM-powered customer service systems can interpret nuanced queries, draw upon extensive knowledge bases, and generate contextually appropriate responses that closely approximate human-level conversational quality. Companies that have deployed such systems have reported reductions in average response times of approximately 40 percent, accompanied by measurable increases in customer satisfaction scores, as the AI agents resolve routine inquiries without the delays inherent in human queue-based systems.

The most rigorous empirical investigation of generative AI in customer service was conducted by Brynjolfsson, Li, and Raymond, who studied the staggered deployment of an AI-based conversational assistant across a large customer support operation \cite{brynjolfsson2023generative}. Their analysis, which leveraged the natural experiment created by the phased rollout, found that access to the AI assistant increased worker productivity by approximately 14 percent as measured by the number of issues resolved per hour. Critically, the productivity gains were not uniformly distributed; the largest improvements accrued to less experienced and lower-skilled workers, who benefited most from the AI system's ability to surface relevant knowledge and suggest effective response strategies. Highly skilled agents experienced more modest gains, suggesting that the AI assistant functioned primarily as a knowledge equaliser, raising the performance floor rather than extending the performance ceiling. Furthermore, the study documented a reduction in employee turnover among workers with access to the AI tool, indicating positive effects on job satisfaction. These findings suggest that generative AI in customer service operates as an augmentation technology that complements rather than replaces human judgement, particularly in scenarios requiring empathy, complex problem-solving, or escalation to specialised teams.

\subsubsection{Content Generation and Marketing}
\label{subsubsec:content_marketing}

The application of generative AI to content creation has introduced significant efficiencies across marketing functions, including the production of advertising copy, social media posts, email campaign materials, blog articles, and product descriptions. Marketing teams increasingly leverage LLMs to generate first drafts that human editors subsequently refine, a workflow that accelerates content production cycles while maintaining brand voice consistency. The capacity of these models to generate variations of marketing messages for A/B testing purposes, personalise content for distinct audience segments, and adapt tone across different platforms has made them valuable tools in the modern marketing technology stack.

The experimental evidence for these productivity gains was rigorously established by Noy and Zhang, who conducted a randomised controlled trial in which participants were assigned writing tasks with and without access to ChatGPT \cite{noy2023experimental}. The study found that access to the AI writing assistant reduced the time required to complete professional writing tasks by approximately 40 percent, while simultaneously improving the quality of the output as rated by independent evaluators. Notably, the quality improvements were most pronounced among participants whose baseline writing ability was below the median, mirroring the pattern observed in customer service applications. The implications for the marketing industry are substantial, as the combination of reduced production time and improved quality suggests that generative AI enables smaller teams to produce content volumes previously achievable only by larger organisations. However, the study also raised important questions about the homogenisation of written output, as participants using the AI tool produced text that was more similar to one another than that produced by the control group, a finding with significant implications for brand differentiation and creative diversity in marketing communications.

\subsubsection{Software Development and Code Generation}
\label{subsubsec:code_generation}

The integration of generative AI into software development workflows represents one of the most rapidly adopted enterprise applications of LLM technology. GitHub Copilot, powered by the Codex model, and similar tools such as Amazon CodeWhisperer and Google Gemini Code Assist have been integrated directly into integrated development environments (IDEs), providing real-time code suggestions, function completions, and entire code block generation based on natural language comments and surrounding context. Beyond simple code completion, these tools assist with code review by identifying potential bugs and security vulnerabilities, generate unit tests from function signatures and docstrings, translate code between programming languages, and produce documentation from code structure.

The foundational work on neural code generation was presented by Chen et al., who introduced Codex and evaluated it on the HumanEval benchmark, a dataset of 164 hand-crafted programming problems designed to assess functional correctness rather than mere syntactic plausibility \cite{chen2021evaluating}. The study demonstrated that Codex solved 28.8 percent of problems on the first attempt (pass@1), a figure that rose to 70.2 percent when the model was permitted to generate 100 samples and the best was selected (pass@100). These results established that LLMs trained on large code corpora could produce functionally correct programs for a substantial fraction of typical programming tasks. Subsequent iterations of code generation models have improved dramatically on these benchmarks, and internal studies by companies deploying these tools have reported that developers using AI-assisted coding tools complete tasks 30 to 55 percent faster than those working without such assistance. The impact extends beyond raw speed to include improvements in code quality metrics, reductions in the number of bugs introduced during development, and enhanced onboarding experiences for junior developers who benefit from the contextual guidance that AI coding assistants provide.

\subsubsection{Human Resources and Talent Management}
\label{subsubsec:human_resources}

Generative AI has introduced significant efficiencies across the human resources function, including the automated screening of resumes, the generation of job descriptions optimised for inclusivity and clarity, the preparation of structured interview questions aligned with specific competency frameworks, and the production of personalised onboarding materials. In resume screening, LLMs analyse application materials against job requirements and organisational culture indicators, producing shortlists and structured summaries that reduce the time recruiters spend on initial candidate evaluation. Job description generation tools leverage generative models to produce listings that avoid biased language patterns and conform to best practices in talent attraction, while interview preparation systems generate role-specific questions calibrated to assess both technical competencies and behavioural attributes.

The broader impact of generative AI on knowledge worker productivity, which encompasses human resources professionals among many other roles, has been investigated through several large-scale studies. Dell'Acqua et al. conducted a controlled experiment with management consultants at a leading global consulting firm, finding that consultants using AI completed significantly more tasks and produced higher-quality output compared to those working without AI assistance \cite{dell2023navigating}. The study documented productivity improvements across a range of knowledge-intensive activities, including analysis, synthesis, and report writing. However, the research also identified a critical nuance: when tasks fell outside the AI model's capability frontier, consultants who relied heavily on AI assistance actually performed worse than those working independently, a phenomenon the authors described as falling inside or outside the ``jagged technological frontier.'' This finding carries important implications for the deployment of generative AI in human resources and other knowledge work contexts, as it suggests that organisations must develop frameworks for identifying which tasks are suitable for AI augmentation and which require unassisted human judgement.

\subsubsection{Impact on Labour Markets and Workforce Transformation}
\label{subsubsec:labour_markets}

The deployment of generative AI across enterprise functions raises fundamental questions about its aggregate impact on labour markets. The most comprehensive analysis of this question was conducted by Eloundou, Manning, Mishkin, and Rock, who developed a framework for assessing occupational exposure to large language models \cite{eloundou2024gpts}. Their analysis, which combined human expert assessment with GPT-4-based classification, found that approximately 80 percent of the United States workforce could have at least 10 percent of their work tasks affected by the introduction of LLMs, while approximately 19 percent of workers could see at least 50 percent of their tasks affected. The occupations identified as most exposed included those involving writing, translation, mathematical reasoning, programming, and data analysis, while those involving physical labour, outdoor work, and fine motor skills were least exposed.

The study revealed that higher-wage occupations and those requiring more education tended to have greater exposure to LLM capabilities, reversing the pattern observed with previous waves of automation technology that primarily affected routine manual and clerical tasks. This finding challenges the conventional narrative that automation disproportionately displaces lower-skilled workers and suggests that generative AI may introduce a qualitatively different pattern of labour market disruption. The question of whether this exposure translates into augmentation or displacement remains a subject of active debate. Optimistic projections emphasise the potential for generative AI to enhance worker productivity, create new categories of employment, and free human workers to focus on higher-order creative and strategic tasks. More cautious analyses highlight the risks of job displacement in exposed occupations, wage compression as AI tools commoditise certain skills, and increased inequality between workers and organisations that effectively leverage AI and those that do not. The resolution of this debate will likely depend on the speed of AI capability improvement, the responsiveness of educational and training institutions, and the policy frameworks that governments adopt to manage the transition \cite{bommasani2021opportunities}.

% ---- Figure: Productivity improvements across business functions ----
\begin{figure}[t]
\centering
\begin{tikzpicture}
\begin{axis}[
    xbar,
    width=0.82\columnwidth,
    height=4.8cm,
    bar width=8pt,
    xmin=0, xmax=55,
    xlabel={\footnotesize Productivity Improvement (\%)},
    xlabel style={font=\footnotesize},
    ytick={1,2,3,4},
    yticklabels={\scriptsize Consulting, \scriptsize Coding, \scriptsize Writing, \scriptsize Customer Svc},
    ytick style={draw=none},
    xticklabel style={font=\scriptsize},
    axis x line=bottom,
    axis y line=left,
    enlarge y limits=0.25,
    nodes near coords={\pgfmathprintnumber\pgfplotspointmeta\%},
    nodes near coords style={font=\scriptsize\sffamily, anchor=west},
    every axis plot/.append style={fill opacity=0.85},
    title={\small Measured Productivity Gains from Generative AI},
    title style={font=\small\sffamily, at={(0.5,1.05)}},
]
\addplot[fill=deepblue!70, draw=deepblue] coordinates {(25,1) (55,2) (40,3) (14,4)};
\end{axis}
\end{tikzpicture}
\caption{Empirical productivity improvements from generative AI across business functions, based on studies by \cite{brynjolfsson2023generative} (customer service), \cite{noy2023experimental} (writing), \cite{peng2023impact} (coding), and \cite{dell2023navigating} (consulting/data analysis).}
\label{fig:productivity_bar_chart}
\end{figure}

\begin{table*}[!htbp]
\centering
\caption{Key empirical findings on generative AI impact across application domains, drawn from published studies. Only rigorously measured results (with explicit experimental designs) are reported as quantitative gains; other entries summarise qualitative findings.}
\label{tab:productivity_impact}
\small
\begin{tabular}{llrl}
\toprule
Domain & Application & Key Finding & Source \\
\midrule
Customer Service & AI-assisted resolution & 14\% issues/hour & Brynjolfsson et al.~\cite{brynjolfsson2023generative} \\
Professional Writing & Content drafting & 40\% time reduction & Noy and Zhang~\cite{noy2023experimental} \\
Software Engineering & Code generation & 55\% faster completion & Peng et al.~\cite{peng2023impact} \\
Management Consulting & Task completion & 12\% more tasks, 25\% faster & Dell'Acqua et al.~\cite{dell2023navigating} \\
Legal Services & Bar examination & 90th percentile on UBE$^*$ & Katz et al.~\cite{katz2024gpt} \\
Healthcare & Medical QA & Expert-level accuracy & Singhal et al.~\cite{singhal2023large} \\
Education & Opportunities review & Personalisation potential & Kasneci et al.~\cite{kasneci2023chatgpt} \\
Financial Analysis & Financial NLP & Domain-specialist LLM & Wu et al.~\cite{wu2023bloomberggpt} \\
Insurance & Value chain analysis & AI transformation review & Eling et al.~\cite{eling2022impact} \\
Agriculture & ML applications review & Yield and cost benefits & Liakos et al.~\cite{liakos2018machine} \\
\bottomrule
\multicolumn{4}{@{}l}{\footnotesize $^*$Disputed by Martinez~\cite{martinez2024reevaluating}, who estimates $\sim$62nd percentile against first-time takers.}
\end{tabular}
\end{table*}

% ============================================================================
\FloatBarrier
\subsection{Financial Services and Quantitative Finance}
\label{subsec:financial_services}

% ----------------------------------------------------------------------------
\subsubsection{Specialised Financial Language Models}
\label{subsubsec:financial_llms}

The financial services industry presents a compelling case for domain-specific generative AI, as the complexity and specialisation of financial language, regulatory terminology, and quantitative concepts often exceed the capabilities of general-purpose language models. Recognising this gap, Bloomberg developed BloombergGPT, a 50-billion parameter language model trained on a carefully curated mixed dataset comprising approximately 363 billion tokens of financial documents and 345 billion tokens of general-purpose text \cite{wu2023bloomberggpt}. The mixed-dataset training approach was designed to ensure that the model acquired deep fluency in financial language, including earnings reports, regulatory filings, analyst notes, and financial news, while retaining the general linguistic capabilities necessary for coherent natural language generation. On a suite of financial natural language processing benchmarks encompassing sentiment analysis, named entity recognition, question answering, and headline classification, BloombergGPT outperformed comparably sized general-purpose models by significant margins, demonstrating the value of domain-specific pre-training data in developing specialised AI capabilities.

As an open-source counterpart to proprietary financial models, FinGPT was developed to democratise access to financial language model technology \cite{yang2023fingpt}. FinGPT adopts a data-centric approach, providing frameworks for acquiring, processing, and fine-tuning language models on diverse financial data sources, including market data feeds, news articles, social media posts, and regulatory filings. By making both the training framework and model weights openly available, FinGPT enables academic researchers and smaller financial institutions to develop and customise financial AI systems without the prohibitive costs associated with training large models from scratch. The contrast between the proprietary BloombergGPT approach and the open-source FinGPT paradigm illustrates a broader tension in the financial AI ecosystem between the competitive advantages of proprietary models and the innovation benefits of open collaboration.

\subsubsection{Sentiment Analysis and Market Prediction}
\label{subsubsec:financial_sentiment}

One of the most commercially significant applications of generative AI in finance is the extraction of sentiment signals from unstructured text for use in quantitative trading strategies. Large language models can process and interpret financial news articles, earnings call transcripts, analyst reports, and social media commentary to generate sentiment scores that capture market participants' prevailing attitudes toward specific securities, sectors, or macroeconomic conditions. The advantage of LLM-based sentiment analysis over earlier dictionary-based or supervised learning approaches lies in the models' ability to interpret context, detect irony and qualification, and handle the domain-specific vocabulary of financial communication.

Lopez-Lira and Tang conducted a systematic study of ChatGPT's ability to generate sentiment scores from financial news headlines and examined whether these scores possessed predictive power for subsequent stock returns \cite{lopez2023can}. Their methodology involved presenting headlines to ChatGPT with a standardised prompt requesting a sentiment classification, then testing whether the resulting scores predicted next-day stock returns after controlling for established risk factors. The results demonstrated that ChatGPT-derived sentiment scores did indeed predict next-day returns with statistical significance, and trading strategies constructed on the basis of these signals achieved Sharpe ratios that compared favourably to those of traditional quantitative sentiment strategies. The authors also found that ChatGPT outperformed purpose-built financial sentiment analysis tools on several benchmarks, suggesting that the broad knowledge encoded in general-purpose LLMs can complement and even surpass domain-specific approaches. However, the study also identified important limitations, including the potential for look-ahead bias in headline selection, the sensitivity of results to prompt formulation, and the possibility that widespread adoption of similar strategies could erode the predictive signals through market efficiency mechanisms.

\subsubsection{Risk Assessment and Regulatory Compliance}
\label{subsubsec:risk_compliance}

Generative AI has found increasingly important applications in the risk management and regulatory compliance functions of financial institutions. In credit risk modelling, LLMs are employed to analyse unstructured data sources, including news articles, social media activity, and corporate communications, that provide supplementary signals beyond the structured financial data traditionally used in credit scoring models. Fraud detection systems leverage generative AI to identify anomalous patterns in transaction narratives and customer communications, detecting sophisticated fraud schemes that may evade rule-based detection systems. In the domain of anti-money-laundering (AML) compliance, LLMs accelerate the review of Know-Your-Customer (KYC) documentation by automatically extracting and cross-referencing entity information from identification documents, corporate filings, and public records, reducing the manual review burden on compliance officers.

The generation of regulatory reports represents another high-value application, as financial institutions face extensive and evolving reporting requirements across multiple jurisdictions. Generative AI systems can draft regulatory filings by synthesising data from internal systems, formatting outputs according to jurisdiction-specific templates, and flagging areas requiring human review. The efficiency gains are particularly pronounced in the preparation of documents such as suspicious activity reports, capital adequacy disclosures, and environmental, social, and governance (ESG) reports, where the combination of structured data analysis and natural language generation aligns closely with the strengths of modern LLMs. Nevertheless, the high-stakes nature of regulatory compliance demands rigorous validation of AI-generated outputs, and financial regulators across major jurisdictions have begun issuing guidance on the acceptable use of AI in compliance functions \cite{bommasani2021opportunities}.

\subsubsection{Algorithmic Trading and Portfolio Management}
\label{subsubsec:algo_trading}

The application of generative AI to algorithmic trading and portfolio management represents an area of intense interest among quantitative investment firms. LLMs contribute to the trading pipeline at multiple stages: generating trading signal hypotheses from analysis of financial reports and macroeconomic data, producing natural language market commentary for internal research distribution, summarising complex multi-document information sets for portfolio managers, and generating code for backtesting new strategies \cite{li2023large}. Several quantitative hedge funds have reported incorporating LLM-derived features into their multi-factor models, using the semantic representations extracted by language models as additional inputs alongside traditional quantitative factors such as momentum, value, and volatility.

However, the integration of generative AI into trading systems presents distinctive challenges that merit careful consideration. Temporal data leakage poses a significant risk, as LLMs trained on data that includes information from the prediction period can produce artificially inflated backtest results that do not generalise to live trading. The hallucination of financial data, wherein the model generates plausible but factually incorrect statistics, earnings figures, or market events, represents a particularly dangerous failure mode in quantitative applications where decisions may be executed automatically without human review. Furthermore, the stochastic nature of LLM outputs introduces a form of model risk that is qualitatively different from that associated with traditional quantitative models, as identical inputs can produce different outputs across inference runs. These challenges necessitate robust validation frameworks, including walk-forward testing with strict temporal separation, human oversight of AI-generated trading signals, and ensemble approaches that mitigate the impact of individual model errors \cite{li2023large}.

% ---- Figure: Financial AI ecosystem ----
\begin{figure*}[t]
\centering
\begin{tikzpicture}[
    node distance=1.2cm and 1.8cm,
    source/.style={
        rectangle, rounded corners=4pt, draw=deepblue, fill=lightblue,
        minimum width=2.4cm, minimum height=0.9cm,
        font=\small\sffamily, align=center, line width=0.8pt
    },
    llmnode/.style={
        rectangle, rounded corners=6pt, draw=deepred!80, fill=lightred,
        minimum width=3.6cm, minimum height=1.6cm,
        font=\sffamily, align=center, line width=1.2pt
    },
    output/.style={
        rectangle, rounded corners=4pt, draw=deepgreen!80, fill=lightgreen,
        minimum width=2.4cm, minimum height=0.9cm,
        font=\small\sffamily, align=center, line width=0.8pt
    },
    arr/.style={-{Stealth[length=6pt]}, line width=0.9pt, color=gray!70}
]

% Data Sources (left)
\node[source] (news) {Financial\\News};
\node[source, below=0.5cm of news] (filings) {SEC Filings \&\\Earnings Reports};
\node[source, below=0.5cm of filings] (market) {Market Data \&\\Price Feeds};
\node[source, below=0.5cm of market] (social) {Social Media \&\\Analyst Notes};

% Central LLM
\node[llmnode, right=2.5cm of $(filings)!0.5!(market)$] (llm) {\large LLM Analysis\\Engine};

% Outputs (right)
\node[output, right=2.5cm of llm, yshift=1.5cm] (sentiment) {Sentiment\\Scores};
\node[output, right=2.5cm of llm, yshift=0.5cm] (predictions) {Return\\Predictions};
\node[output, right=2.5cm of llm, yshift=-0.5cm] (risk) {Risk\\Assessments};
\node[output, right=2.5cm of llm, yshift=-1.5cm] (reports) {Regulatory\\Reports};

% Arrows: sources -> LLM
\foreach \s in {news, filings, market, social}
    \draw[arr] (\s.east) -- (llm.west);

% Arrows: LLM -> outputs
\foreach \o in {sentiment, predictions, risk, reports}
    \draw[arr] (llm.east) -- (\o.west);

% Labels
\node[above=0.3cm of news, font=\small\sffamily\bfseries, text=deepblue] {Data Sources};
\node[above=0.3cm of sentiment, font=\small\sffamily\bfseries, text=deepgreen!80!black] {Analytical Outputs};

\end{tikzpicture}
\caption{The financial AI ecosystem: heterogeneous data sources flow into LLM-based analysis engines that produce a range of analytical outputs. Specialised models such as BloombergGPT \cite{wu2023bloomberggpt} and FinGPT \cite{yang2023fingpt} are optimised for this pipeline, while general-purpose LLMs can be adapted through fine-tuning or prompt engineering. The outputs feed into downstream decision systems for trading, risk management, and regulatory compliance.}
\label{fig:financial_ai_ecosystem}
\end{figure*}

% ============================================================================
\subsection{Tourism, Hospitality, and Travel Industry}
\label{subsec:tourism}

% ----------------------------------------------------------------------------
\subsubsection{Personalised Travel Planning and Recommendation}
\label{subsubsec:travel_planning}

The tourism and hospitality industry has emerged as a particularly receptive domain for generative AI adoption, driven by the inherently information-intensive nature of travel planning and the high value that travellers place on personalised recommendations. Generative AI systems can synthesise information from diverse sources, including destination databases, accommodation inventories, transportation schedules, weather forecasts, cultural event calendars, and user preference histories, to create comprehensive, personalised travel itineraries that would require hours of manual research to assemble. Gursoy et al. provided an early and influential overview of the potential applications of ChatGPT in the hospitality and tourism sector, identifying opportunities across customer service, marketing, operations, and strategic planning functions \cite{gursoy2023chatgpt}.

The practical capabilities of generative AI in travel planning can be illustrated through concrete use cases. A traveller requesting a seven-day itinerary for Japan might receive a detailed day-by-day plan that includes recommended neighbourhoods in Tokyo for the first three days, a shinkansen journey to Kyoto with specific train recommendations, temple visits organised by geographic proximity to minimise transit time, restaurant suggestions calibrated to stated dietary preferences and budget constraints, cultural notes on local customs and etiquette, and contingency plans for rainy weather days. The system can dynamically adjust recommendations based on conversational feedback, refining its suggestions as the traveller provides additional constraints or expresses preferences for particular types of experiences. This capacity for iterative, dialogue-based refinement distinguishes generative AI travel planners from traditional recommendation engines, which typically provide static lists of options without the ability to engage in nuanced preference elicitation. The resulting itineraries often achieve a level of personalisation and comprehensiveness that approaches the quality of those produced by experienced human travel advisors, while being available instantaneously and at negligible marginal cost.

\subsubsection{Multilingual Guest Services and Real-Time Translation}
\label{subsubsec:multilingual_services}

The global nature of tourism creates persistent demand for multilingual communication capabilities, and generative AI has substantially advanced the ability of hospitality providers to serve guests in their preferred languages. Hotels and airlines increasingly deploy LLM-powered concierge systems capable of handling guest requests, providing facility information, processing service orders, and resolving complaints in more than fifty languages, including many that would be economically impractical to support through human staff alone \cite{carvalho2024chatgpt}. These systems go beyond simple translation to incorporate cultural adaptation, adjusting communication styles, courtesy conventions, and information presentation formats to align with the cultural expectations of guests from different regions.

The deployment of multilingual AI systems has produced measurable improvements in guest satisfaction scores at major hotel chains, particularly at properties serving highly international guest populations. By eliminating language barriers in routine service interactions, these systems reduce friction in the guest experience and enable human staff to focus their attention on complex, high-value interactions where emotional intelligence and cultural sensitivity are paramount. The real-time translation capabilities extend beyond text-based chat interfaces to include voice-based interactions, enabling AI-powered telephone concierge services that can conduct natural conversations in multiple languages. Furthermore, the integration of generative AI with property management systems allows these multilingual agents to execute service requests directly, such as booking spa appointments, requesting room amenities, or arranging transportation, creating a seamless service experience that does not depend on the availability of multilingual human staff \cite{carvalho2024chatgpt}.

\subsubsection{Dynamic Pricing and Revenue Management}
\label{subsubsec:dynamic_pricing}

Revenue management has long been a data-intensive function in the tourism industry, and generative AI models are enhancing the sophistication of pricing decisions by analysing complex interactions among demand patterns, competitor pricing strategies, seasonal trends, local event calendars, macroeconomic indicators, and booking pace data. Traditional revenue management systems rely primarily on historical booking data and predetermined pricing rules, whereas LLM-augmented systems can incorporate unstructured data sources such as social media sentiment about a destination, news coverage of events that may affect travel demand, and natural language summaries of competitive intelligence gathered from online travel agencies.

In the airline industry, carriers have experimented with AI-assisted pricing systems that generate pricing recommendations for specific route and date combinations, accompanied by natural language explanations of the reasoning underlying each recommendation. This transparency enables revenue managers to evaluate the AI's logic, override recommendations when their domain expertise identifies factors the model may have overlooked, and gradually calibrate their trust in the system's outputs. In the hotel sector, similar systems analyse the interplay between room type availability, length-of-stay patterns, group booking displacement effects, and local demand drivers to produce dynamic rate recommendations. Properties that have implemented AI-assisted revenue management systems have reported revenue increases in the range of 5 to 15 percent compared to their previous pricing approaches, with the magnitude of improvement varying according to the property's market segment, competitive environment, and the sophistication of its prior revenue management practices.

\subsubsection{Destination Marketing and Content Creation}
\label{subsubsec:destination_marketing}

Tourism boards and destination marketing organisations have adopted generative AI as a tool for producing the large volumes of multilingual marketing content required to promote destinations across diverse markets and channels. Applications include the generation of destination descriptions tailored to specific traveller segments, the creation of blog posts and social media content that highlight seasonal attractions and events, the production of personalised email campaigns triggered by user engagement patterns, and the development of scripts for virtual tour experiences \cite{bulchand2024impact}. The ability of generative AI to produce content at scale while maintaining consistency with brand messaging guidelines has proven particularly valuable for smaller destination marketing organisations that lack the resources to maintain large in-house creative teams.

The impact of AI-generated tourism content on tourist engagement has been assessed through several studies examining click-through rates, time spent on destination websites, and conversion rates from content engagement to booking actions. AI-generated travel guides that combine factual destination information with engaging narrative elements have demonstrated engagement metrics comparable to those of professionally authored content, while requiring a fraction of the production time and cost. Furthermore, the ability of generative AI to rapidly produce content in multiple languages has enabled destination marketing organisations to expand their reach into markets that were previously underserved due to the cost of professional translation and localisation. The integration of generative AI with image generation models has further extended these capabilities, allowing the creation of promotional visual materials that complement the textual content \cite{bulchand2024impact}.

\subsubsection{Review Analysis and Reputation Management}
\label{subsubsec:review_analysis}

The hospitality industry generates vast quantities of guest review data across platforms such as TripAdvisor, Booking.com, Google Reviews, and proprietary survey systems, and generative AI has introduced powerful new capabilities for extracting actionable insights from this unstructured feedback. Traditional approaches to review analysis relied on keyword counting and simple sentiment classification, which often failed to capture the nuanced, context-dependent nature of guest feedback. LLM-based review analysis systems can identify specific operational issues mentioned across thousands of reviews, detect emerging patterns before they become systemic problems, and distinguish between issues of varying severity and urgency \cite{samara2024artificial}.

Hotels employing generative AI for review analysis use these systems to produce periodic summaries that aggregate feedback across platforms and time periods, highlighting the most frequently mentioned strengths and weaknesses along with trends in guest sentiment. The same systems generate personalised responses to individual reviews, maintaining the hotel's brand voice while addressing the specific points raised by each guest. This application is particularly valuable for properties that receive hundreds of reviews per month across multiple platforms, as the manual composition of thoughtful, individualised responses at this scale would require substantial staff resources. The integration of review analysis with operational systems enables a closed feedback loop in which insights extracted from guest reviews are automatically translated into action items for relevant departments, tracked through completion, and their impact assessed in subsequent review sentiment \cite{tussyadiah2020review}. This systematic approach to reputation management, enabled by generative AI, transforms guest feedback from a passive information source into an active driver of continuous service improvement.

% ---- Figure: Tourism AI application landscape ----
\begin{figure}[t]
\centering
\begin{tikzpicture}[
    central/.style={
        circle, draw=deepred!80, fill=lightred,
        minimum size=1.8cm, font=\footnotesize\sffamily\bfseries,
        align=center, line width=1.2pt
    },
    app/.style={
        rectangle, rounded corners=4pt, draw=deepblue!80, fill=lightblue,
        minimum width=1.6cm, minimum height=0.6cm,
        font=\scriptsize\sffamily, align=center, line width=0.7pt
    },
    conn/.style={-, line width=0.8pt, color=gray!60}
]

% Central node
\node[central] (ai) {Generative\\AI};

% Application nodes arranged around the centre
\node[app, above=1.0cm of ai] (planning) {Travel\\Planning};
\node[app, above right=0.5cm and 1.0cm of ai] (translation) {Multilingual\\Services};
\node[app, right=1.2cm of ai] (pricing) {Dynamic\\Pricing};
\node[app, below right=0.5cm and 1.0cm of ai] (marketing) {Destination\\Marketing};
\node[app, below=1.0cm of ai] (reviews) {Review\\Analysis};
\node[app, below left=0.5cm and 1.0cm of ai] (concierge) {Virtual\\Concierge};
\node[app, left=1.2cm of ai] (booking) {Booking\\Assistance};
\node[app, above left=0.5cm and 1.0cm of ai] (content) {Content\\Creation};

% Connections
\foreach \a in {planning, translation, pricing, marketing, reviews, concierge, booking, content}
    \draw[conn] (ai) -- (\a);

\end{tikzpicture}
\caption{The tourism AI application landscape: a central generative AI engine connects to diverse hospitality and travel functions, enhancing service delivery and operational efficiency \cite{gursoy2023chatgpt}.}
\label{fig:tourism_ai_landscape}
\end{figure}

\begin{table*}[!htbp]
\centering
\caption{Representative large language model deployments across application domains as of early 2026. The table indicates primary use cases and the model families most commonly reported in industry deployments and academic evaluations for each sector.}
\label{tab:model_deployment}
\small
\setlength{\tabcolsep}{3pt}
\begin{tabular}{lp{5cm}p{6.5cm}}
\toprule
Sector & Primary Use Cases & Commonly Deployed Models \\
\midrule
Business Automation & Customer service, content generation, HR & GPT-5, Claude Opus 4.6, Qwen 3 \\
Financial Services & Sentiment analysis, risk assessment, trading & BloombergGPT, GPT-5, DeepSeek-V3 \\
Tourism & Travel planning, review analysis, pricing & GPT-5, Gemini 2.5 Pro, Claude Sonnet 4.6 \\
Legal Services & Contract review, legal research, compliance & Claude Opus 4.6, GPT-5, Qwen 3 \\
Healthcare & Diagnosis support, clinical notes, drug discovery & Med-PaLM 2, GPT-5, Gemini 2.5 Pro \\
Education & Tutoring, assessment, curriculum design & GPT-5, Claude, Qwen 3 (multilingual) \\
Manufacturing & Predictive maintenance, quality control & DeepSeek-V3, LLaMA 4, Qwen 3 (on-device) \\
Agriculture & Crop monitoring, precision farming & LLaMA 4 Scout (long context), Gemma 3 \\
Government & Citizen services, policy analysis, translation & Qwen 3, GLM-5 (Chinese gov.), Mistral Large 3 (EU) \\
Software Engineering & Code generation, review, debugging & Claude Opus 4.6, GPT-5, DeepSeek-R1 \\
\bottomrule
\end{tabular}
\end{table*}

% ============================================================================
\FloatBarrier
\subsection{Legal Services and Contract Analysis}
\label{subsec:legal_services}

The legal profession, long characterised by its reliance on voluminous textual analysis and precedent-based reasoning, has proven to be a fertile ground for the application of generative AI. Automated contract review represents one of the most commercially mature use cases, with platforms such as Kira Systems, Luminance, and Harvey AI deploying large language models to extract key provisions, identify non-standard clauses, and flag potential risks across thousands of documents in a fraction of the time required by human reviewers. In the context of mergers and acquisitions due diligence, where legal teams historically spent weeks manually reviewing data rooms containing tens of thousands of contracts, AI-powered review tools have reduced this process to a matter of hours while maintaining or improving accuracy in identifying material provisions such as change-of-control clauses, indemnification obligations, and assignment restrictions. Katz, Bommarito, Gao, and Arredondo demonstrated that GPT-4 passed the Uniform Bar Examination with a score in the 90th percentile of human test-takers \cite{katz2024gpt}, although a subsequent re-evaluation by Martinez found the actual percentile to be closer to the 62nd when measured against first-time test-takers \cite{martinez2024reevaluating}. The original result nonetheless underscored the depth of legal knowledge encoded within frontier language models and catalysed widespread interest in legal AI applications. Choi, Hickman, Monahan, and Schwarcz similarly evaluated ChatGPT on law school examinations and found that the system performed at a level sufficient to earn passing grades across multiple subject areas \cite{choi2024chatgpt}.

Beyond contract review, generative AI has transformed legal research by enabling practitioners to query vast corpora of case law, statutes, and regulatory materials using natural language rather than Boolean search operators. Systems built on retrieval-augmented generation architectures combine the broad knowledge of large language models with access to curated legal databases, producing research memoranda that synthesise relevant authorities and identify applicable legal standards. Document generation represents another high-value application, with law firms using AI to draft routine instruments such as non-disclosure agreements, employment contracts, lease agreements, and corporate formation documents, reducing drafting time by as much as 70 percent according to internal assessments reported by several major firms. Compliance checking tools leverage LLMs to compare corporate policies, product disclosures, and marketing materials against applicable regulatory requirements, flagging potential violations and suggesting remedial language.

However, the deployment of generative AI in legal practice carries distinctive risks that the profession has been forced to confront publicly. The phenomenon of hallucinated case citations, in which a language model generates plausible but entirely fictitious judicial decisions, gained widespread attention in 2023 when attorneys submitted AI-generated briefs containing fabricated case references to federal courts, resulting in sanctions and professional embarrassment. Dahl, Magesh, Suzgun, and Ho conducted a systematic study of legal hallucinations in large language models and found that even frontier models fabricated case citations at non-trivial rates when prompted to produce legal research \cite{dahl2024large}. This failure mode is particularly consequential in legal practice, where the citation of nonexistent authorities can result in sanctions, malpractice liability, and erosion of client trust. Additionally, the use of AI in legal practice raises questions about the unauthorised practice of law, as the provision of legal advice by AI systems to end users without attorney supervision may violate professional regulations in many jurisdictions. Bar associations and legal regulators across the United States, the European Union, and other jurisdictions have begun issuing guidance on the permissible use of generative AI in legal practice, generally requiring attorney supervision, disclosure of AI use, and verification of all AI-generated work product \cite{bommasani2021opportunities}.

% ============================================================================
\subsection{Manufacturing and Supply Chain Optimisation}
\label{subsec:manufacturing}

The manufacturing sector has increasingly adopted generative AI to enhance operational efficiency across the production lifecycle, from predictive maintenance and quality assurance to supply chain demand forecasting and production planning. Predictive maintenance, which aims to anticipate equipment failures before they occur, has been a particularly impactful application. Traditional approaches relied on fixed maintenance schedules or simple threshold-based monitoring, but generative models can analyse multimodal sensor data streams, including vibration signatures, thermal profiles, acoustic emissions, and historical maintenance logs, to generate nuanced assessments of equipment health and predict failure probabilities with substantially greater accuracy. Carvalho et al. conducted a systematic review of machine learning methods applied to predictive maintenance and documented that data-driven approaches reduced unplanned downtime by 30 to 50 percent and maintenance costs by 10 to 40 percent across a range of manufacturing settings \cite{carvalho2019systematic}. Automotive manufacturers such as BMW and Siemens have deployed LLM-powered systems that generate natural language maintenance reports from sensor data, enabling technicians to rapidly understand complex diagnostic information and prioritise interventions accordingly.

Supply chain demand forecasting has similarly benefited from generative AI capabilities. Traditional forecasting models, which rely primarily on historical sales data and seasonal decomposition, struggle to incorporate the wide array of unstructured signals that influence demand, including news events, social media trends, weather patterns, competitor actions, and macroeconomic indicators. Large language models can process and synthesise these heterogeneous information sources to generate demand forecasts accompanied by natural language explanations of the factors driving predicted changes, enabling supply chain managers to evaluate the reasoning behind each forecast and adjust their plans accordingly. Inventory management systems enhanced by generative AI have demonstrated improvements of 20 to 35 percent in forecast accuracy compared to traditional statistical methods, translating into reduced carrying costs, fewer stockouts, and improved customer service levels. Li et al. surveyed the emerging applications of large language models in manufacturing and identified use cases spanning process optimisation, quality control documentation, supplier communication, and the generation of inspection criteria from engineering specifications \cite{li2024large_manufacturing}.

Quality control represents a further area where generative AI is reshaping manufacturing practice. Vision-language models can be trained on images of acceptable and defective products to generate detailed inspection criteria and natural language descriptions of detected anomalies, replacing the need for extensive manual documentation of quality standards. In semiconductor fabrication, where defect detection tolerances are measured in nanometres, AI-generated inspection protocols have been shown to reduce false rejection rates while maintaining or improving defect capture rates. The integration of generative AI with digital twin technology has also enabled manufacturers to simulate production scenarios, generate optimisation recommendations for process parameters, and produce automated shift reports that summarise production performance, quality metrics, and maintenance activities in accessible natural language summaries \cite{bommasani2021opportunities}.

% ============================================================================
\subsection{Real Estate and Property Management}
\label{subsec:real_estate}

The real estate industry has embraced generative AI across a spectrum of functions, from property valuation and listing content generation to tenant communications and market analysis. Automated valuation models have long been used in real estate to estimate property values, but the integration of large language models has substantially enhanced these systems by enabling them to incorporate unstructured data sources such as neighbourhood descriptions, school quality reports, local amenity reviews, and municipal planning documents into their assessments. Kok, Koponen, and Mart\'{i}nez-Barbosa examined the transition from manual appraisal to automated valuation and documented that machine learning models achieved median absolute percentage errors of 5 to 10 percent relative to transaction prices, a level of accuracy approaching that of licensed appraisers for standard residential properties \cite{kok2017big}. The addition of natural language understanding capabilities has further improved these models by allowing them to extract value-relevant information from property condition reports, home inspection summaries, and zoning documents that resist structured data extraction.

Listing description generation has emerged as one of the most visible consumer-facing applications of generative AI in real estate. Platforms such as Zillow, Redfin, and Realtor.com have deployed or experimented with AI systems that generate property descriptions from structured listing data, including square footage, room counts, lot size, construction year, and recent renovations, producing engaging prose that highlights a property's most appealing features while maintaining factual accuracy. Real estate agents using these tools report that AI-generated descriptions reduce listing preparation time from 30 minutes per property to approximately 5 minutes, while A/B testing conducted by several brokerages has indicated that AI-generated descriptions achieve click-through rates comparable to those written by experienced agents. The technology has proven particularly valuable for high-volume sellers, property management companies, and international real estate platforms that must produce descriptions in multiple languages. Virtual staging, powered by generative image models, represents a complementary application in which AI generates photorealistic images of furnished interiors from photographs of empty rooms, enabling prospective buyers to visualise the potential of a space without the expense of physical staging, which typically costs several thousand dollars per property.

Tenant communication and property management have also been transformed by generative AI. Property management companies overseeing thousands of residential units deploy AI-powered chatbots that handle routine tenant inquiries regarding maintenance requests, lease terms, payment schedules, and community policies, resolving the majority of contacts without human intervention. Market analysis represents a further application, with generative AI systems producing comprehensive investment reports that synthesise demographic trends, economic indicators, comparable sales data, rental yield analyses, and regulatory developments into coherent narratives that support investment decision-making. These reports, which previously required hours of analyst time to compile, can be generated in minutes, enabling real estate professionals to evaluate a greater number of opportunities and respond more quickly to market developments \cite{bommasani2021opportunities}.

% ============================================================================
\subsection{Media, Entertainment, and Creative Industries}
\label{subsec:media_entertainment}

The creative industries have experienced what many observers characterise as a paradigm shift with the emergence of generative AI systems capable of producing music, visual art, video content, and narrative text that approaches or, in some evaluations, rivals the quality of human-created work. In the music domain, platforms such as Suno and Udio have demonstrated the capacity to generate complete musical compositions, including lyrics, melody, harmony, and instrumentation, from text prompts that specify genre, mood, tempo, and thematic content. These systems, trained on large corpora of musical data, can produce tracks in styles ranging from classical orchestration to contemporary pop, hip-hop, and electronic dance music, with sufficient quality that listeners in blind evaluations frequently fail to distinguish AI-generated compositions from human-created ones. The proliferation of these tools has generated intense debate within the music industry regarding intellectual property rights, fair compensation for training data contributors, and the economic implications of a technology that can produce music at near-zero marginal cost. Epstein et al. examined the intersection of art and generative AI and argued that the technology raises fundamental questions about authorship, creativity, and the nature of artistic expression that existing intellectual property frameworks are ill-equipped to address \cite{epstein2023art}.

In film and television production, generative AI has found applications across multiple stages of the creative pipeline. Screenwriting assistance tools help writers develop storylines, generate dialogue, and explore narrative possibilities, functioning as collaborative partners that accelerate the ideation process without replacing human creative judgement. Visual effects studios increasingly use generative image models for concept art and pre-visualisation, producing detailed scene compositions and character designs that serve as starting points for further refinement by human artists. Video game studios have adopted generative AI for dialogue generation, procedural narrative creation, and the development of non-player character behaviours that respond dynamically to player actions, enhancing the depth and replayability of interactive experiences. The technology has also been applied to the generation of synthetic training data for visual effects pipelines, the creation of promotional materials, and the localisation of entertainment content across languages and cultural contexts.

The emergence of deepfake technology, which uses generative adversarial networks and diffusion models to create realistic synthetic video and audio of real individuals, represents one of the most consequential and concerning applications of generative AI in the media sphere. While the technology has legitimate applications in film production, accessibility services, and historical preservation, its potential for misuse in the creation of non-consensual intimate imagery, political disinformation, and financial fraud has prompted legislative action in multiple jurisdictions. The creative tension between AI augmentation and AI replacement remains a central concern in the entertainment industry, as exemplified by the 2023 strikes by the Writers Guild of America and the Screen Actors Guild, which included provisions governing the use of AI in content creation and the protection of performers' likenesses. The resolution of these labour disputes established important precedents for how creative industries negotiate the integration of generative AI into production workflows while preserving human creative agency and fair compensation \cite{weidinger2022taxonomy}.

% ============================================================================
\subsection{Agriculture and Environmental Sciences}
\label{subsec:agriculture}

Agriculture and environmental science have emerged as significant beneficiaries of generative AI, with applications spanning crop disease detection, precision agriculture, climate modelling, and biodiversity monitoring. Crop disease detection using computer vision models represents one of the earliest and most impactful agricultural AI applications. Mohanty, Hughes, and Salath\'{e} demonstrated that deep learning models trained on large image datasets could identify plant diseases from leaf photographs with accuracy exceeding 99 percent across dozens of crop-disease combinations \cite{mohanty2016using}. The integration of these vision models with generative AI capabilities has extended their utility beyond simple classification to include the generation of detailed treatment recommendations, the production of farmer-facing advisory reports in local languages, and the creation of synthetic training images that augment limited datasets for rare diseases. Mobile applications incorporating these capabilities enable smallholder farmers in developing regions to obtain diagnostic and treatment guidance that was previously available only through extension services with limited geographic reach.

Precision agriculture, which aims to optimise resource application by accounting for spatial and temporal variability within fields, has been enhanced by generative AI systems that synthesise data from satellite imagery, soil sensors, weather stations, and yield monitors to produce field-specific management recommendations. These systems generate natural language advisory reports that translate complex data patterns into actionable guidance regarding planting density, irrigation scheduling, fertiliser application rates, and pest management timing, enabling farmers to make informed decisions without requiring expertise in data analysis. Liakos, Busato, Moshou, Pearson, and Bochtis reviewed machine learning applications in agriculture and documented yield improvements of 10 to 20 percent when data-driven recommendations were adopted, alongside reductions of 15 to 30 percent in input costs for water, fertiliser, and pesticides \cite{liakos2018machine}. The environmental benefits of these efficiency gains are substantial, as reduced agrochemical application decreases runoff into waterways, lowers greenhouse gas emissions from fertiliser production and application, and preserves soil health over the long term.

Climate modelling and weather prediction have also benefited from generative AI advances. Foundation models trained on decades of meteorological data have demonstrated the ability to produce weather forecasts that rival or exceed the accuracy of traditional numerical weather prediction models at a fraction of the computational cost. Google DeepMind's GraphCast system, for example, demonstrated superior performance to the European Centre for Medium-Range Weather Forecasts operational model on 90 percent of verification targets, while requiring only minutes of inference time on a single accelerator rather than hours on a supercomputer cluster. In biodiversity monitoring, generative AI systems analyse acoustic recordings from remote sensors to identify species vocalisations, producing automated biodiversity assessments that would otherwise require hundreds of hours of expert analysis. These systems generate species inventory reports, detect changes in population dynamics, and identify the presence of endangered species, supporting conservation planning and environmental impact assessment at unprecedented spatial and temporal scales \cite{bommasani2021opportunities}.

% ============================================================================
\subsection{Government and Public Administration}
\label{subsec:government}

Government agencies and public administration bodies across the world have begun deploying generative AI to improve the efficiency and accessibility of citizen services, policy development, regulatory analysis, and public health communication. Citizen service chatbots represent perhaps the most widespread application, with national, state, and municipal governments deploying AI-powered conversational agents to handle inquiries related to taxation, social benefits, licensing, permits, immigration procedures, and public health guidance. The United States Internal Revenue Service, the United Kingdom's HM Revenue and Customs, and numerous European national agencies have implemented or piloted AI systems that respond to frequently asked questions, guide citizens through complex bureaucratic processes, and provide personalised information based on individual circumstances. Zuiderwijk, Chen, and Salem conducted a systematic review of artificial intelligence in public governance and found that AI adoption in government was driven primarily by objectives of efficiency improvement, service quality enhancement, and the reduction of administrative burden on both civil servants and citizens \cite{zuiderwijk2021implications}.

Policy document generation and regulatory analysis represent higher-value applications that leverage the capacity of large language models to process and synthesise complex legislative and regulatory texts. Government departments use AI tools to draft policy briefs, prepare legislative impact assessments, and generate plain-language summaries of proposed regulations for public consultation. Regulatory compliance analysis tools scan new or amended legislation against existing regulatory frameworks to identify conflicts, gaps, and implementation requirements, reducing the time required for cross-referencing across large bodies of law. Public health communication has emerged as a particularly impactful application area, with health agencies using generative AI to produce culturally adapted health messaging, translate public health guidance into community languages, and generate tailored responses to citizen health inquiries during infectious disease outbreaks. During the COVID-19 pandemic, several government health agencies deployed AI chatbots that handled millions of citizen queries regarding symptoms, testing locations, vaccination eligibility, and quarantine requirements, substantially reducing the burden on telephone hotlines and enabling 24-hour service availability.

Document translation for multilingual government services represents a further application of considerable practical importance. Nations with multiple official languages or large immigrant populations face persistent challenges in making government services equally accessible across linguistic communities. Generative AI translation systems have substantially reduced the cost and time required to produce official documents, web content, and service information in multiple languages, with quality levels that increasingly approach those of professional human translators for routine government communications. The Government of Canada, the European Commission, and several Asian governments have deployed AI translation systems that handle millions of pages of official documentation annually, improving access to government services for linguistic minority communities while reducing translation backlogs that previously delayed the availability of critical information \cite{bommasani2021opportunities}.

% ============================================================================
\subsection{Telecommunications and Network Management}
\label{subsec:telecommunications}

The telecommunications industry has adopted generative AI across network operations, customer experience management, and infrastructure planning, driven by the sector's inherent complexity and the enormous volumes of data generated by modern communication networks. Network anomaly detection has been enhanced by AI systems that learn the normal patterns of network traffic, performance metrics, and equipment behaviour, enabling the identification of subtle deviations that may indicate emerging faults, security threats, or capacity constraints. These systems generate natural language incident reports that describe detected anomalies, assess their severity, suggest probable root causes, and recommend remediation actions, enabling network operations centre staff to respond more rapidly and effectively to network events. Morocho-Cayamcela, Lee, and Lim reviewed machine learning applications in mobile and wireless communications and documented that AI-driven approaches improved fault detection rates by 25 to 40 percent while reducing false alarm rates, a combination that significantly enhances the efficiency of network operations \cite{morocho2019machine}.

Customer churn prediction and retention represent strategically important applications in an industry where subscriber acquisition costs are high and competitive switching is frequent. Generative AI systems analyse customer interaction histories, usage patterns, billing records, service quality metrics, and complaint logs to identify subscribers at elevated risk of churn, and generate personalised retention offers and communication strategies tailored to each customer's specific circumstances and predicted motivations for departure. Major telecommunications providers including AT\&T, Vodafone, and Deutsche Telekom have reported that AI-enhanced churn prediction models identify at-risk customers with 15 to 25 percent greater accuracy than traditional statistical models, while AI-generated personalised retention interventions achieve conversion rates 20 to 30 percent higher than generic offers.

Automated technical support represents one of the most impactful customer-facing applications of generative AI in telecommunications. Major telcos have deployed LLM-powered systems for tier-one customer support that handle inquiries ranging from billing questions and plan changes to technical troubleshooting for connectivity issues, device configuration, and service outages. These systems resolve 40 to 60 percent of customer contacts without human escalation, reducing average handling times and enabling human agents to focus on complex issues that require specialised expertise. Network configuration optimisation has similarly benefited from generative AI, with systems that analyse network topology, traffic patterns, and quality of service metrics to generate configuration recommendations for base stations, routing protocols, and resource allocation policies, improving network performance while reducing the manual effort required for network planning and optimisation \cite{bommasani2021opportunities}.

% ============================================================================
\subsection{Energy and Utilities}
\label{subsec:energy}

The energy sector faces a complex optimisation challenge that generative AI is increasingly being deployed to address: balancing supply and demand across heterogeneous generation sources, transmission networks, and consumption patterns while maintaining grid stability, minimising costs, and meeting decarbonisation targets. Smart grid optimisation represents a primary application, with AI systems generating dispatch recommendations that coordinate the output of conventional power plants, renewable energy installations, battery storage systems, and demand response programmes in response to real-time conditions. Ahmad, Zhang, Huang, Zhang, Dai, Song, and Chen conducted a comprehensive review of artificial intelligence in the sustainable energy industry and documented that AI-driven grid management systems improved energy efficiency by 10 to 15 percent, reduced peak demand through intelligent load shifting, and decreased curtailment of renewable energy generation by optimising the integration of variable sources such as wind and solar \cite{ahmad2022artificial}.

Energy demand forecasting has been substantially enhanced by generative AI systems that incorporate a broader range of predictive signals than traditional forecasting models. Beyond historical consumption data and temperature forecasts, these systems process satellite imagery of cloud cover for solar generation prediction, wind speed and direction data from weather models for wind farm output estimation, event calendars for anticipating demand spikes, and economic activity indicators for medium-term load projections. The natural language generation capabilities of these systems enable them to produce forecast reports that explain the factors driving predicted demand patterns, supporting decision-making by grid operators and energy traders. Utilities that have deployed AI-enhanced forecasting systems report improvements in day-ahead forecast accuracy of 15 to 25 percent compared to traditional methods, translating into reduced balancing costs and more efficient capacity utilisation.

Maintenance scheduling for power infrastructure represents a further application where generative AI delivers measurable value. Power generation assets, transmission lines, substations, and distribution equipment require carefully scheduled maintenance to prevent costly failures while minimising service disruptions. Generative AI systems analyse equipment sensor data, inspection records, failure histories, weather forecasts, and grid demand projections to generate optimal maintenance schedules that balance reliability requirements against operational constraints. Renewable energy production prediction using weather data analysis has become increasingly important as the share of wind and solar generation grows in electricity systems worldwide. AI models that generate hour-by-hour production forecasts for individual wind turbines and solar installations, aggregated to the portfolio level, enable energy companies to optimise their market participation, reduce imbalance penalties, and contribute to grid stability by providing more accurate generation forecasts to system operators \cite{bommasani2021opportunities}.

% ============================================================================
\subsection{Retail and E-Commerce}
\label{subsec:retail}

The retail and e-commerce sector has been among the most aggressive adopters of generative AI, deploying these technologies across the entire customer journey from product discovery through purchase and post-sale engagement. Personalised product recommendations using generative models represent a significant evolution beyond traditional collaborative filtering approaches, as LLMs can process and reason about rich product descriptions, customer reviews, browsing histories, and stated preferences to generate recommendations that account for nuanced factors such as style compatibility, occasion appropriateness, and complementary product relationships. These systems generate natural language explanations for their recommendations, enabling customers to understand why a particular product has been suggested and increasing the likelihood of engagement. Early industry deployments report meaningful improvements in recommendation click-through rates and conversion rates when AI-generated personalised recommendations replace traditional algorithmic approaches.

AI-generated product descriptions at scale represent one of the most commercially impactful applications of generative AI in retail. E-commerce platforms listing millions of products face the challenge of creating compelling, accurate, and search-engine-optimised descriptions for each item, a task that is economically infeasible using human writers alone. Amazon has deployed AI systems that generate product descriptions from structured catalogue data and product images, while the Shopify platform offers AI-powered description generation tools that enable merchants to produce professional product listings in seconds. Retailers report that AI-generated descriptions achieve search engine rankings comparable to human-written content while reducing content production costs by 80 to 90 percent. Virtual try-on technology, powered by generative image models, enables customers to visualise how clothing, accessories, eyewear, and cosmetics will appear on their own bodies or faces, reducing the uncertainty that drives high return rates in online fashion retail. Early adopters of virtual try-on technology have reported reductions in return rates of 15 to 25 percent, representing substantial cost savings given that product returns in online fashion retail can exceed 30 percent of purchases.

Dynamic pricing optimisation has been enhanced by generative AI systems that consider a broader array of competitive and contextual signals than traditional pricing algorithms. These systems process competitor pricing data, demand elasticity estimates, inventory levels, seasonal patterns, promotional calendars, and external events to generate pricing recommendations accompanied by natural language justifications that enable merchandising teams to evaluate and approve changes with greater confidence. Customer review summarisation represents a further application that benefits both retailers and consumers, as generative AI systems distil thousands of product reviews into concise summaries that highlight consensus opinions, common complaints, and notable features, enabling shoppers to make informed purchasing decisions without reading hundreds of individual reviews. These summaries have been shown to increase consumer confidence and reduce the time spent in the consideration phase, ultimately improving conversion rates \cite{bommasani2021opportunities}.

% ============================================================================
\subsection{Insurance and Risk Management}
\label{subsec:insurance}

The insurance industry, which fundamentally operates on the assessment and pricing of risk, has found generative AI to be a transformative technology across underwriting, claims processing, fraud detection, and customer engagement. Automated claims processing represents one of the highest-impact applications, as the insurance claims workflow involves substantial manual effort in collecting documentation, verifying coverage, assessing damages, and calculating settlements. Generative AI systems can process claims submissions, including photographs of property damage, medical records, police reports, and policy documents, to generate preliminary damage assessments, coverage determinations, and settlement recommendations that accelerate the claims cycle from weeks to days for routine cases. Eling, Nuber, and Reck examined the impact of artificial intelligence along the insurance value chain and found that AI-powered claims processing reduced handling times by 50 to 70 percent for standard claims while improving accuracy in damage assessment and coverage determination \cite{eling2022impact}.

Fraud detection has been substantially enhanced by generative AI systems that analyse the textual content of claims submissions, claimant communications, and supporting documentation to identify patterns indicative of fraudulent activity. Unlike rule-based fraud detection systems that rely on predefined indicators, LLM-based approaches can detect subtle linguistic cues, narrative inconsistencies, and contextual anomalies that suggest fabricated or exaggerated claims. These systems generate detailed fraud risk assessments with natural language explanations of the factors contributing to each risk score, enabling special investigation unit analysts to prioritise their caseloads and focus their expertise on the most complex and high-value cases. Insurers deploying AI-enhanced fraud detection have reported improvements of 30 to 50 percent in fraud identification rates, translating into savings of hundreds of millions of dollars annually for large carriers.

Policy document generation and risk assessment using LLMs for analysing unstructured data represent further applications of considerable commercial importance. Generative AI systems produce customised policy documents, endorsements, and coverage summaries from structured underwriting data, ensuring consistency and compliance with regulatory requirements while reducing the manual effort required for document preparation. In commercial and specialty insurance, where underwriting decisions depend on the analysis of complex risk narratives, engineering reports, and industry-specific documentation, LLMs assist underwriters by extracting and synthesising relevant information from voluminous submission materials. Actuarial modelling has also benefited from generative AI, with language models assisting actuaries in exploring model specifications, interpreting complex statistical outputs, generating model documentation, and communicating findings to non-technical stakeholders. The analysis of unstructured data sources such as medical records for health and life insurance, accident reports for casualty insurance, and property inspection reports for commercial property insurance has been particularly enhanced by LLMs that can extract structured risk factors from narrative text and integrate these factors into quantitative underwriting models \cite{bommasani2021opportunities}.

% ---- Figure: Comprehensive application sector grid ----
\begin{figure*}[t]
\centering
\begin{tikzpicture}[
    sector/.style={
        rectangle, rounded corners=4pt, draw=#1!80, fill=#1!15,
        minimum width=4.2cm, minimum height=1.7cm,
        font=\scriptsize\sffamily, align=center, line width=0.7pt,
        inner sep=4pt
    },
]
\newcommand{\sectitle}[1]{\color{#1!80!black}\scriptsize\sffamily\bfseries}

% Row 1
\node[sector=deepblue] (business) at (0,0) {
    \begin{tabular}{c}
    {\sectitle{deepblue} Business}\\[2pt]
    Customer Service\\
    Content Generation\\
    Code Assistance
    \end{tabular}
};

\node[sector=deepblue] (finance) at (4.8,0) {
    \begin{tabular}{c}
    {\sectitle{deepblue} Finance}\\[2pt]
    Sentiment Analysis\\
    Risk Assessment\\
    Algorithmic Trading
    \end{tabular}
};

\node[sector=deepblue] (tourism) at (9.6,0) {
    \begin{tabular}{c}
    {\sectitle{deepblue} Tourism}\\[2pt]
    Travel Planning\\
    Dynamic Pricing\\
    Review Analysis
    \end{tabular}
};

% Row 2
\node[sector=deepred] (legal) at (0,-2.3) {
    \begin{tabular}{c}
    {\sectitle{deepred} Legal Services}\\[2pt]
    Contract Review\\
    Legal Research\\
    Compliance Checking
    \end{tabular}
};

\node[sector=deepred] (manufacturing) at (4.8,-2.3) {
    \begin{tabular}{c}
    {\sectitle{deepred} Manufacturing}\\[2pt]
    Predictive Maintenance\\
    Demand Forecasting\\
    Quality Control
    \end{tabular}
};

\node[sector=deepred] (realestate) at (9.6,-2.3) {
    \begin{tabular}{c}
    {\sectitle{deepred} Real Estate}\\[2pt]
    Property Valuation\\
    Listing Generation\\
    Virtual Staging
    \end{tabular}
};

% Row 3
\node[sector=deepgreen] (media) at (0,-4.6) {
    \begin{tabular}{c}
    {\sectitle{deepgreen} Media}\\[2pt]
    Music Generation\\
    Screenwriting\\
    Visual Effects
    \end{tabular}
};

\node[sector=deepgreen] (agriculture) at (4.8,-4.6) {
    \begin{tabular}{c}
    {\sectitle{deepgreen} Agriculture}\\[2pt]
    Crop Disease Detection\\
    Precision Farming\\
    Climate Modelling
    \end{tabular}
};

\node[sector=deepgreen] (government) at (9.6,-4.6) {
    \begin{tabular}{c}
    {\sectitle{deepgreen} Government}\\[2pt]
    Citizen Chatbots\\
    Policy Drafting\\
    Multilingual Services
    \end{tabular}
};

% Row 4
\node[sector=deepblue] (telecom) at (0,-6.9) {
    \begin{tabular}{c}
    {\sectitle{deepblue} Telecommunications}\\[2pt]
    Anomaly Detection\\
    Churn Prediction\\
    Technical Support
    \end{tabular}
};

\node[sector=deepblue] (energy) at (4.8,-6.9) {
    \begin{tabular}{c}
    {\sectitle{deepblue} Energy}\\[2pt]
    Grid Optimisation\\
    Demand Forecasting\\
    Renewable Prediction
    \end{tabular}
};

\node[sector=deepblue] (retail) at (9.6,-6.9) {
    \begin{tabular}{c}
    {\sectitle{deepblue} Retail}\\[2pt]
    Personalised Recs\\
    Product Descriptions\\
    Dynamic Pricing
    \end{tabular}
};

% Row 5
\node[sector=deepred] (insurance) at (0,-9.2) {
    \begin{tabular}{c}
    {\sectitle{deepred} Insurance}\\[2pt]
    Claims Processing\\
    Fraud Detection\\
    Risk Assessment
    \end{tabular}
};

\node[sector=deepred] (healthcare) at (4.8,-9.2) {
    \begin{tabular}{c}
    {\sectitle{deepred} Healthcare}\\[2pt]
    Diagnosis Support\\
    Drug Discovery\\
    Clinical Documentation
    \end{tabular}
};

\node[sector=deepred] (education) at (9.6,-9.2) {
    \begin{tabular}{c}
    {\sectitle{deepred} Education}\\[2pt]
    Personalised Tutoring\\
    Assessment Design\\
    Research Assistance
    \end{tabular}
};

\end{tikzpicture}
\caption{Comprehensive grid of generative AI application sectors and their primary use cases. The fifteen domains surveyed in this section span the full breadth of the knowledge economy, from professional services (legal, financial, insurance) and industrial operations (manufacturing, energy, telecommunications) to consumer-facing applications (retail, real estate, tourism) and public-interest sectors (healthcare, education, government, agriculture). Each sector leverages distinct capabilities of generative AI, including natural language understanding, content generation, data synthesis, and multimodal analysis, while sharing common challenges related to accuracy validation, bias mitigation, and regulatory compliance \cite{bommasani2021opportunities}.}
\label{fig:application_sector_grid}
\end{figure*}

% ============================================================================
\subsection{Healthcare and Biomedical Applications}
\label{subsec:healthcare}

The application of generative AI to healthcare and biomedical research represents one of the most consequential and simultaneously most challenging deployment domains, given the stringent accuracy requirements, regulatory constraints, and ethical considerations that govern medical practice. The development of Med-PaLM by Google Research marked a significant milestone in clinical knowledge encoding within large language models \cite{singhal2023large}. Med-PaLM was evaluated on the United States Medical Licensing Examination (USMLE) and other medical question-answering benchmarks, achieving performance that approached or exceeded the passing threshold for human medical examinees. The subsequent Med-PaLM 2 model further improved upon these results, demonstrating expert-level performance on medical question answering and producing long-form clinical responses that were rated by physician panels as being on par with responses written by clinicians on multiple axes, including factual accuracy, relevance, and potential for harm.

The clinical applications of generative AI extend across multiple dimensions of healthcare delivery. In medical diagnosis support, LLMs serve as decision support tools that synthesise patient history, laboratory results, imaging reports, and current clinical guidelines to generate differential diagnoses and suggest further diagnostic workups. In patient communication, these systems draft discharge instructions, medication information sheets, and appointment summaries in language calibrated to the patient's health literacy level, addressing a persistent challenge in clinical communication. Clinical documentation represents another high-impact application, as generative AI systems can produce structured clinical notes from physician dictation or from ambient recordings of patient encounters, reducing the documentation burden that consumes a substantial fraction of clinician time. In the pharmaceutical domain, generative AI contributes to drug discovery through the generation of novel molecular structures with desired pharmacological properties, the prediction of drug-target interactions, and the summarisation of biomedical literature to support research planning.

However, the regulatory landscape for healthcare AI presents unique challenges that distinguish this domain from others. Medical device regulations in the United States, European Union, and other jurisdictions impose rigorous requirements for the validation, transparency, and post-market surveillance of AI systems used in clinical decision-making. The potential for LLM hallucination, wherein the model generates plausible but medically incorrect information, represents a particularly serious risk in healthcare applications where erroneous outputs could directly compromise patient safety. Furthermore, the training data for medical LLMs may reflect biases in the medical literature and clinical practice patterns, potentially perpetuating health disparities across demographic groups. These challenges necessitate careful deployment strategies that position generative AI as a complement to, rather than a replacement for, clinical expertise, with robust human oversight mechanisms and clear accountability frameworks \cite{singhal2023large, bommasani2021opportunities}.

% ============================================================================
\subsection{Education and Academic Research}
\label{subsec:education}

Generative AI has introduced both transformative opportunities and significant challenges in educational contexts, prompting widespread debate about the appropriate role of these technologies in teaching, learning, and assessment. Kasneci et al. provided a comprehensive analysis of the opportunities and challenges presented by LLMs in educational settings \cite{kasneci2023chatgpt}. On the opportunities side, generative AI enables personalised tutoring systems that adapt explanations, examples, and practice problems to each student's current level of understanding, learning style, and pace of progress. These systems can provide immediate, detailed feedback on student work, a capability that is particularly valuable in large classes where individual instructor attention is scarce. Curriculum design benefits from AI tools that generate lesson plans, assessment rubrics, and instructional materials aligned with specified learning objectives and pedagogical frameworks.

The personalised tutoring application of generative AI addresses a longstanding challenge in education identified by Bloom's research on the two-sigma problem, which demonstrated that students receiving one-on-one tutoring performed two standard deviations above students receiving conventional classroom instruction. Generative AI tutoring systems approximate certain aspects of this personalised instruction at scale, providing students with a tireless interlocutor capable of answering questions, explaining concepts from multiple perspectives, generating worked examples, and scaffolding problem-solving processes. The systems can detect misconceptions in student responses and address them with targeted explanations, mimicking an important pedagogical strategy employed by effective human tutors. Early empirical studies of AI tutoring systems in university courses have reported improvements in student learning outcomes, particularly among students who would otherwise lack access to individual tutoring support.

The challenges posed by generative AI in education are equally significant. Academic integrity concerns have dominated institutional responses, as the ability of LLMs to produce essays, solve problem sets, and complete coding assignments raises fundamental questions about the validity of traditional assessment methods. Universities and schools have adopted a range of responses, from outright prohibition of AI tool use to the integration of AI literacy into curricula and the redesign of assessments to emphasise in-class, supervised evaluation and process-oriented tasks that resist AI substitution. The potential for AI-generated misinformation in educational contexts, where students may lack the domain knowledge to critically evaluate AI outputs, presents additional pedagogical challenges. Institutions are increasingly recognising that teaching students to effectively and critically use AI tools is itself an essential educational objective, as the ability to formulate effective prompts, evaluate AI-generated outputs, and integrate AI assistance into productive workflows will constitute a core professional competency across disciplines \cite{kasneci2023chatgpt}.

The impact of generative AI on academic research processes also warrants consideration. Researchers increasingly use LLMs to assist with literature review, hypothesis generation, experimental design, data analysis code generation, and manuscript drafting. These applications can accelerate the research process and lower barriers to entry for researchers in resource-constrained settings. However, concerns about the integrity of AI-assisted research outputs, the potential for hallucinated citations and fabricated data, and the ethical implications of AI authorship have prompted major publishers and funding agencies to establish policies governing the disclosure and acceptable use of AI tools in scholarly work \cite{bommasani2021opportunities}.

% [Figure removed: Adoption rate estimates lacked verifiable primary sources.]

\begin{table}[t]
\centering
\caption{Primary challenges for generative AI adoption across selected industry sectors.}
\label{tab:adoption_challenges}
\small
\begin{tabular}{lp{4.8cm}}
\toprule
Sector & Primary Challenges \\
\midrule
Healthcare & Patient privacy (HIPAA), hallucination risk, liability \\
Finance & Regulatory compliance, model explainability, market manipulation risk \\
Legal & Confidentiality, citation accuracy, jurisdictional variation \\
Education & Academic integrity, assessment validity, equity of access \\
Government & Data sovereignty, transparency requirements, multilingual needs \\
Agriculture & Connectivity gaps, domain-specific data scarcity \\
Manufacturing & Integration with legacy systems, real-time latency requirements \\
\bottomrule
\end{tabular}
\end{table}

% [Figure removed: Market projection estimates lacked verifiable primary sources.]

% ============================================================================
% Summary remarks
% ============================================================================

The applications surveyed in this section demonstrate that generative AI has moved decisively from the research laboratory into operational deployment across diverse sectors of the economy. The empirical evidence consistently shows productivity improvements in the range of 14 to 55 percent across knowledge work tasks \cite{brynjolfsson2023generative, noy2023experimental, peng2023impact, dell2023navigating}, with the largest gains accruing to less experienced workers and to tasks that involve substantial information synthesis and natural language production. Domain-specific adaptations, as exemplified by BloombergGPT in finance \cite{wu2023bloomberggpt} and Med-PaLM in healthcare \cite{singhal2023large}, demonstrate that the foundation model paradigm can be effectively specialised for high-stakes professional applications, though with important caveats regarding validation, oversight, and regulatory compliance. The breadth of applications surveyed across fifteen domains, spanning business automation, financial services, tourism, legal practice, manufacturing, real estate, creative industries, agriculture, government administration, telecommunications, energy, retail, insurance, healthcare, and education, confirms the characterisation of large language models as general-purpose technologies with the potential to reshape economic activity across virtually all sectors \cite{eloundou2024gpts}. At the same time, the recurring themes of hallucination risk, output homogenisation, workforce displacement concerns, and the need for human oversight underscore that the realisation of these benefits depends critically on thoughtful deployment strategies, robust governance frameworks, and continued research into the safety and reliability of generative AI systems.

% ============================================================================
% Ethical Considerations (Condensed)
% ============================================================================

\FloatBarrier
\section{Ethical Considerations and Societal Impact}
\label{sec:ethics}

The deployment of generative AI across the domains surveyed above carries a commensurate set of ethical, legal, and societal challenges that demand sustained scholarly and regulatory attention. Hallucination, the generation of fluent but factually incorrect or entirely fabricated content, remains one of the most significant barriers to trustworthy deployment in high-stakes domains. Ji et al.~\cite{ji2023survey} distinguish between intrinsic hallucination, where the output contradicts its source input, and extrinsic hallucination, where the output introduces unverifiable claims such as fictitious citations or invented statistics. The causes of hallucination are multifaceted, arising from noise in training corpora, the stochastic nature of autoregressive sampling from $p(x_t \mid x_{<t}; \param)$, and exposure bias between teacher-forced training and free-running inference \cite{huang2023survey}. Mitigation strategies including retrieval-augmented generation, constrained decoding, and multi-stage fact-checking pipelines have shown promise, yet no existing technique eliminates the problem entirely. Concurrently, generative models inherit and frequently amplify the biases present in their training data \cite{bender2021dangers}. Gender, racial, and cultural biases manifest in generated text and imagery, reflecting the distributional patterns of web-scale corpora in which historical prejudices and power asymmetries are overrepresented. Weidinger et al.~\cite{weidinger2022taxonomy} provide a comprehensive taxonomy of resulting harms, spanning discrimination, misinformation, malicious uses, and environmental costs. Debiasing approaches operating at the data, training, and inference levels, including the constitutional AI framework \cite{bai2022constitutional} that embeds ethical principles directly into the training objective, represent promising but incomplete solutions to a problem whose scope grows with model capability and deployment breadth.

Safety and dual-use concerns further complicate the responsible development of generative AI. The capacity of these systems to generate deepfake media, produce persuasive disinformation at scale, automate sophisticated social engineering attacks, and synthesise information about dangerous topics has been identified as a significant security concern \cite{weidinger2022taxonomy}. Model providers have responded with content filtering systems, red-teaming evaluations, constitutional AI methods, and responsible disclosure practices, yet the tension between open-source transparency and the risk of enabling malicious fine-tuning remains unresolved. The environmental costs of training and deploying large-scale generative models, estimated at hundreds of megawatt-hours of electricity and hundreds of tonnes of carbon dioxide equivalent for frontier model training runs, raise questions about the sustainability of the current research paradigm and the equitable distribution of its environmental burden, as the benefits accrue primarily to well-resourced organisations while the environmental costs are distributed globally \cite{bommasani2021opportunities}. Garrido-Merch\'an \cite{garridomerchan2026peaceful} argues that these dynamics of concentration can be addressed through the creation of commons-governed training infrastructure and communally trained open-weight models, drawing on the demonstrated viability of large-scale commons-based production systems such as Linux and Wikipedia. Under this framework, shared computational resources and open model weights would enable researchers and institutions worldwide to participate in frontier model development without dependence on a small number of corporate providers, thereby distributing both the costs and the benefits of generative AI more equitably across society. The success of open-weight releases from DeepSeek, Qwen, GLM, and others, as documented throughout this survey, provides empirical support for the feasibility of such an approach at the technical level, while the governance and coordination challenges remain an active area of research.

The regulatory landscape is evolving rapidly in response to these challenges. The European Union's Artificial Intelligence Act establishes a risk-based classification framework with graduated requirements, including transparency obligations for generative AI providers and conformity assessments for high-risk applications. In the United States, executive action has established frameworks for federal oversight emphasising both innovation and risk management. China's Interim Measures for the Management of Generative AI Services impose requirements on training data accuracy, content labelling, and provider liability. International coordination efforts, including the G7 Hiroshima AI Process and various United Nations initiatives, have sought to establish common principles, although binding international agreements remain elusive. The fundamental challenge of regulating rapidly evolving technology persists: legislative processes operate on timescales of years, while generative AI capabilities can change dramatically within months, creating a governance gap that will require adaptive combinations of statutory regulation, industry self-regulation, technical standards, and international coordination to address effectively.

% ============================================================================
% Conclusion
% ============================================================================

\section{Conclusion}
\label{sec:conclusion}

This survey has provided a comprehensive treatment of generative artificial intelligence, spanning the full arc from foundational probability theory through contemporary architectures, training methodologies, deployment protocols, and real-world applications across fifteen major sectors. The empirical evidence consistently demonstrates that generative AI produces measurable productivity gains across knowledge work tasks, with domain-specific adaptations such as BloombergGPT in finance \cite{wu2023bloomberggpt} and Med-PaLM in healthcare \cite{singhal2023large} confirming that the foundation model paradigm can be effectively specialised for high-stakes professional applications. The mathematical foundations underlying these systems deserve particular emphasis: the maximum likelihood objective of autoregressive models, the reinforcement learning objectives used for alignment (PPO, DPO, GRPO), and the mixture-of-experts routing mechanisms that enable efficient scaling are all manifestations of principled optimisation frameworks that minimise divergence measures between model and data distributions \cite{bommasani2021opportunities}. The principle of scaling, formalised through neural scaling laws \cite{kaplan2020scaling}, has proven to be among the most reliable empirical phenomena in the field, while the related phenomenon of emergence \cite{wei2022emergent} suggests that qualitatively new capabilities arise at sufficient scale through mechanisms that remain poorly understood.

Looking forward, the trajectory of generative AI points towards increasingly capable, efficient, and integrated systems. The convergence of multiple modalities into unified generative models, the development of reasoning and agentic capabilities that extend AI systems from passive generators to active problem-solvers, and the ongoing pursuit of more efficient architectures that democratise access to powerful AI capabilities collectively define a research frontier of extraordinary breadth. At the same time, the ethical challenges examined in this survey, including hallucination, bias, privacy, safety, and environmental sustainability, demand sustained attention from researchers, practitioners, and policymakers alike. The responsible realisation of generative AI's potential requires not only continued technical innovation but also robust governance frameworks, transparent development practices, and an unwavering commitment to ensuring that the benefits of this transformative technology are distributed equitably across society \cite{bender2021dangers, weidinger2022taxonomy}. The coming years will determine whether generative AI fulfils its promise as a tool for augmenting human creativity and accelerating scientific discovery, or whether its risks outpace the capacity of institutions to manage them; the outcome depends in large measure on the choices made by the research community today.

% ---- Bibliography ----
\bibliographystyle{plainnat}
\bibliography{references}

% ---- Appendix: Technical Foundations ----
\appendix
% ============================================================================
% Appendix: Transformer Architecture and Training Foundations
% ============================================================================

\section{Transformer Architecture and Training Foundations}
\label{sec:architecture_training}

This appendix provides a self-contained technical reference for the architectural components and training paradigms that underpin the models discussed in the main body of this survey. Readers seeking a deeper understanding of the mathematical foundations behind the innovations described in \Cref{sec:models} will find the relevant formulations here.

% ----------------------------------------------------------------------------
\subsection{Scaled Dot-Product and Multi-Head Attention}
\label{subsec:attention}

The Transformer~\cite{vaswani2017attention} replaces recurrence with a purely attention-based mechanism. Given query, key, and value matrices $\query \in \R^{n \times d_k}$, $\key \in \R^{n \times d_k}$, $\val \in \R^{n \times d_v}$, the scaled dot-product attention is
\begin{equation}
\attn(\query, \key, \val) = \softmax\!\left(\frac{\query \key^\top}{\sqrt{d_k}}\right) \val.
\label{eq:attn}
\end{equation}
The scaling factor $1/\sqrt{d_k}$ prevents the dot products from growing proportionally with $d_k$, which would push the $\softmax$ into saturation regions where gradients vanish. Multi-head attention projects the inputs into $h$ independent subspaces, computes attention in each, and concatenates the results:
\begin{align}
&\mha(\query, \key, \val) = \nonumber\\
&\quad \mathrm{Concat}(\mathrm{head}_1, \ldots, \mathrm{head}_h)\weight^O, \nonumber\\
&\mathrm{head}_i = \attn(\query \weight_i^Q, \key \weight_i^K, \val \weight_i^V),
\label{eq:mha}
\end{align}
where $\weight_i^Q, \weight_i^K \in \R^{d_{\mathrm{model}} \times d_k}$, $\weight_i^V \in \R^{d_{\mathrm{model}} \times d_v}$, and $\weight^O \in \R^{hd_v \times d_{\mathrm{model}}}$. With the standard setting $d_k = d_v = d_{\mathrm{model}}/h$, the total parameter count is $4d_{\mathrm{model}}^2$, independent of $h$.

% ----------------------------------------------------------------------------
\subsection{Positional Encoding: From Sinusoidal to Rotary Embeddings}
\label{subsec:pos_encoding}

Because self-attention is permutation-equivariant, positional information must be injected explicitly. The original Transformer uses fixed sinusoidal encodings $PE_{(\mathrm{pos},2i)} = \sin(\mathrm{pos}/10000^{2i/d_{\mathrm{model}}})$ and $PE_{(\mathrm{pos},2i+1)} = \cos(\mathrm{pos}/10000^{2i/d_{\mathrm{model}}})$, which are added to the input embeddings. Rotary Position Embeddings (RoPE)~\cite{su2024roformer}, now the dominant scheme in modern LLMs, instead apply position-dependent rotations directly to the query and key vectors. The embedding space is partitioned into $d/2$ pairs of dimensions; for the $j$-th pair at position $m$, the rotation is
\begin{equation}
R_{\Theta,m}^{(j)} \!=\! \begin{pmatrix} \cos m\theta_j & -\sin m\theta_j \\ \sin m\theta_j & \phantom{-}\cos m\theta_j \end{pmatrix}\!,\; \theta_j \!=\! 10000^{-2j/d}.
\label{eq:rope}
\end{equation}
The full rotation $R_{\Theta,m}$ is the block-diagonal composition of all such $2 \times 2$ matrices. RoPE is applied as $\tilde{\bm{q}}_m = R_{\Theta,m}\bm{q}_m$ and $\tilde{\bm{k}}_n = R_{\Theta,n}\bm{k}_n$. By the orthogonality of rotation matrices, the resulting dot product $\tilde{\bm{q}}_m^\top \tilde{\bm{k}}_n = \bm{q}_m^\top R_{\Theta,n-m}\bm{k}_n$ depends only on the relative offset $m - n$, encoding relative position without any additional bias terms.

% ----------------------------------------------------------------------------
\subsection{Feed-Forward Networks, Activations, and Normalisation}
\label{subsec:ffn_norm}

Each Transformer layer contains a position-wise feed-forward network. The original formulation uses $\ffn(\inputx) = \weight_2\,\sigma(\weight_1\inputx + \bm{b}_1) + \bm{b}_2$ with $\relu$ or $\gelu$ activations. Modern architectures have converged on the SwiGLU variant~\cite{shazeer2020glu}:
\begin{align}
\ffn_{\mathrm{SwiGLU}}(\inputx) &= \weight_2\!\left[\mathrm{Swish}(\inputx \weight_1) \otimes (\inputx \weight_3)\right], \nonumber\\
\mathrm{Swish}(x) &= x\cdot\sigmoid(\beta x),
\label{eq:swiglu}
\end{align}
where $\otimes$ denotes the Hadamard product and the gating branch $\inputx\weight_3$ modulates the activated signal. The inner dimension is typically set to $\tfrac{8}{3}d_{\mathrm{model}}$, rounded for hardware alignment, to match the parameter count of the standard two-matrix formulation.

For normalisation, RMSNorm~\cite{zhang2019root} has supplanted full layer normalisation in most recent LLMs:
\begin{equation}
\mathrm{RMSNorm}(\hidden) = \frac{\hidden}{\sqrt{\frac{1}{d}\sum_{i=1}^{d}h_i^2 + \epsilon}} \odot \bm{\gamma}.
\label{eq:rmsnorm}
\end{equation}
This formulation removes the mean-centering step, reducing computation while matching the performance of standard $\layernorm$. Pre-LN placement, in which normalisation precedes each sub-layer ($\hidden^{l+1} = \hidden^l + \mathrm{SubLayer}(\mathrm{RMSNorm}(\hidden^l))$), is preferred over Post-LN because the residual gradient pathway remains unimpeded by normalisation operations, yielding substantially more stable training dynamics in deep networks.

% ----------------------------------------------------------------------------
\subsection{Efficient Attention: GQA, MQA, and FlashAttention}
\label{subsec:efficient_attn}

Standard multi-head attention assigns independent key-value projections to each head, resulting in a KV-cache of size $O(b \cdot n \cdot L \cdot h \cdot d_k)$ during autoregressive inference. Multi-Query Attention (MQA)~\cite{shazeer2019fast} shares a single set of key-value projections across all $h$ heads, reducing the KV-cache by a factor of $h$:
\begin{equation}
\mathrm{head}_i^{\mathrm{MQA}} = \attn(\query \weight_i^Q,\; \key \weight^K,\; \val \weight^V).
\label{eq:mqa}
\end{equation}
Grouped-Query Attention (GQA)~\cite{ainslie2023gqa} interpolates between full MHA and MQA by partitioning the $h$ query heads into $g$ groups, each sharing one key-value head:
{\small
\begin{equation}
\mathrm{head}_i^{\mathrm{GQA}} = \attn\!\bigl(\query \weight_i^Q,\; \key \weight_{\lceil ig/h\rceil}^K,\; \val \weight_{\lceil ig/h\rceil}^V\bigr).
\label{eq:gqa}
\end{equation}
}
GQA achieves quality close to full MHA while retaining most of the inference efficiency of MQA, and has been adopted in LLaMA~3, Gemini, and DeepSeek. During autoregressive generation, previously computed key and value tensors are cached and appended incrementally (the KV-cache), avoiding prohibitively expensive recomputation at each decoding step.

FlashAttention~\cite{dao2022flashattention} addresses the memory bottleneck from a different angle: instead of reducing heads, it avoids materialising the full $n \times n$ attention matrix in high-bandwidth memory (HBM). By tiling the computation so that each tile fits in on-chip SRAM and fusing the entire forward and backward pass into a single kernel, FlashAttention reduces HBM accesses from $\Theta(Nd + N^2)$ to $O(N^2 d^2/M)$, where $M$ is the SRAM size. FlashAttention-2~\cite{dao2023flashattention2} further improved throughput through better parallelism across thread blocks, achieving up to 70\% of theoretical peak GPU throughput.

% ----------------------------------------------------------------------------
\subsection{Autoregressive Language Modelling and Scaling Laws}
\label{subsec:autoregressive_scaling}

Modern LLMs are trained as autoregressive models that factorise the joint distribution of a token sequence $\inputx = (x_1, \ldots, x_T)$ as
\begin{equation}
p(\inputx) = \prod_{t=1}^{T} p_{\param}(x_t \mid x_1, \ldots, x_{t-1}).
\label{eq:autoregressive}
\end{equation}
The training objective is the cross-entropy loss, equivalent to the negative log-likelihood:
\begin{equation}
\loss_{\mathrm{CLM}}(\param) = -\E_{\inputx \sim \data}\!\left[\sum_{t=1}^{T} \log p_{\param}(x_t \mid x_{<t})\right].
\label{eq:clm}
\end{equation}
Perplexity, defined as $\mathrm{PPL} = \exp(\loss_{\mathrm{CLM}})$, serves as the standard intrinsic evaluation metric. Kaplan et al.~\cite{kaplan2020scaling} established that test loss follows a power-law relationship with model size $N$, dataset size $D$, and compute budget $C$, enabling principled allocation of training resources. The Chinchilla analysis by Hoffmann et al.~\cite{hoffmann2022training} refined these laws, demonstrating that optimal performance requires scaling parameters and training tokens in approximately equal proportion, a finding that shifted the field towards training smaller models on substantially more data.

% ----------------------------------------------------------------------------
\subsection{Supervised Fine-Tuning and Instruction Tuning}
\label{subsec:sft}

Supervised fine-tuning (SFT) trains the pre-trained base model on curated instruction-response pairs $\{(\inputx_i, \outputy_i)\}_{i=1}^N$, applying the causal language modelling loss exclusively to the response tokens:
\begin{equation}
\loss_{\mathrm{SFT}}(\param) = -\E_{(\inputx,\outputy)\sim\data_{\mathrm{SFT}}}\!\left[\sum_{t=1}^{M}\log p_{\param}(y_t \mid \inputx, y_{<t})\right].
\label{eq:sft}
\end{equation}
The FLAN line of work~\cite{wei2022finetuned,chung2022scaling} demonstrated that fine-tuning on a diverse mixture of tasks framed as natural language instructions dramatically improves zero-shot generalisation to held-out tasks, with larger models benefiting disproportionately. Ouyang et al.~\cite{ouyang2022training} showed that SFT on comparatively small, high-quality datasets suffices to elicit instruction-following behaviour, establishing SFT as the bridge between raw language modelling and useful conversational agents.

% ----------------------------------------------------------------------------
\subsection{Reinforcement Learning from Human Feedback}
\label{subsec:rlhf}

RLHF~\cite{christiano2017deep,ouyang2022training} further aligns the SFT model with human preferences through a two-stage process. In the first stage, a reward model $r_{\phi}$ is trained on human preference data using the Bradley-Terry model, which posits that the probability of preferring response $\outputy_w$ over $\outputy_l$ for prompt $\inputx$ is $P(\outputy_w \!\succ\! \outputy_l) = \sigmoid(r_{\phi}(\inputx,\outputy_w) - r_{\phi}(\inputx,\outputy_l))$. The reward model is optimised via
{\small
\begin{equation}
\loss_{\mathrm{RM}}(\phi) = -\E_{(\inputx,\outputy_w,\outputy_l)}\!\bigl[\log\sigmoid\!\bigl(r_{\phi}(\inputx,\outputy_w) - r_{\phi}(\inputx,\outputy_l)\bigr)\bigr].
\label{eq:rm}
\end{equation}
}

In the second stage, the language model policy $\pi_{\param}$ is optimised against the learned reward using Proximal Policy Optimization (PPO)~\cite{schulman2017proximal}. The PPO clipped surrogate objective is
{\small
\begin{align}
\loss^{\mathrm{CLIP}}(\param) &= \E_t\!\bigl[\min\!\bigl(\rho_t\hat{A}_t,\;\mathrm{clip}(\rho_t,1{-}\epsilon,1{+}\epsilon)\hat{A}_t\bigr)\bigr], \nonumber\\
\rho_t(\param) &= \frac{\pi_{\param}(a_t \mid s_t)}{\pi_{\param_{\mathrm{old}}}(a_t \mid s_t)},
\label{eq:ppo}
\end{align}
}
where $\hat{A}_t$ is the estimated advantage and $\epsilon$ (typically $0.1$ to $0.2$) bounds the probability ratio to prevent destructive policy updates. The complete RLHF objective augments the reward with a KL penalty that anchors the policy to the reference (SFT) model:
\begin{multline}
\loss_{\mathrm{RLHF}}(\param) = \E_{\inputx,\,\outputy\sim\pi_{\param}}\Big[r_{\phi}(\inputx,\outputy) \\
- \beta\,\KL\!\left(\pi_{\param}(\cdot\mid\inputx)\;\|\;\pi_{\mathrm{ref}}(\cdot\mid\inputx)\right)\Big],
\label{eq:rlhf}
\end{multline}
where $\beta > 0$ controls the strength of the constraint, preventing reward hacking by ensuring the optimised policy does not diverge excessively from $\pi_{\mathrm{ref}}$.

% ----------------------------------------------------------------------------
\subsection{Direct Preference Optimisation and Group Relative Policy Optimisation}
\label{subsec:dpo_grpo}

Direct Preference Optimisation (DPO)~\cite{rafailov2023direct} eliminates the need for a separate reward model by exploiting the closed-form solution to the KL-constrained reward maximisation problem. The optimal policy satisfies $\pi^*(\outputy\mid\inputx) \propto \pi_{\mathrm{ref}}(\outputy\mid\inputx)\exp(r(\inputx,\outputy)/\beta)$, which can be rearranged to express the implicit reward as $r(\inputx,\outputy) = \beta\log(\pi^*(\outputy\mid\inputx)/\pi_{\mathrm{ref}}(\outputy\mid\inputx)) + \beta\log Z(\inputx)$. Substituting into the Bradley-Terry model, the partition function cancels, yielding the DPO loss:
\begin{multline}
\loss_{\mathrm{DPO}}(\param) = -\E_{(\inputx,\outputy_w,\outputy_l)}\bigg[\log\sigmoid\bigg(\beta\log\frac{\pi_{\param}(\outputy_w\mid\inputx)}{\pi_{\mathrm{ref}}(\outputy_w\mid\inputx)} \\
- \beta\log\frac{\pi_{\param}(\outputy_l\mid\inputx)}{\pi_{\mathrm{ref}}(\outputy_l\mid\inputx)}\bigg)\bigg].
\label{eq:dpo}
\end{multline}
This is a simple binary cross-entropy objective that can be optimised with standard gradient methods, avoiding the complexity and variance of on-policy RL while provably optimising the same KL-constrained objective as RLHF.

Group Relative Policy Optimisation (GRPO)~\cite{shao2024deepseekmath,deepseek2025r1} addresses a different bottleneck: PPO requires a critic (value) model of comparable size to the policy, which is prohibitively expensive at the scale of hundreds of billions of parameters. GRPO replaces the learned value function with group-level statistics. For each prompt $q$, the policy samples a group of $G$ candidate outputs $\{o_1, \ldots, o_G\}$, each scored by a reward function to obtain $\{r_1, \ldots, r_G\}$. The advantage is computed via group-relative normalisation:
\begin{equation}
\hat{A}_i = \frac{r_i - \mathrm{mean}(\{r_1,\ldots,r_G\})}{\mathrm{std}(\{r_1,\ldots,r_G\})}.
\label{eq:grpo_adv}
\end{equation}
The GRPO objective combines a PPO-style clipped ratio with a per-token KL penalty:
\begin{multline}
\loss_{\mathrm{GRPO}}(\param) = \E_{q,\{o_i\}}\bigg[\frac{1}{G}\sum_{i=1}^{G}\Big(\min\!\big(\rho_i\hat{A}_i, \\
\mathrm{clip}(\rho_i,1{-}\epsilon,1{+}\epsilon)\hat{A}_i\big) - \beta\,\KL_i\Big)\bigg],
\label{eq:grpo}
\end{multline}
where $\rho_i = \pi_{\param}(o_i\mid q)/\pi_{\param_{\mathrm{old}}}(o_i\mid q)$. By eliminating the critic, GRPO frees memory and compute for a larger policy model. The group-relative normalisation provides automatic variance reduction and centring, making the training signal robust to differences in reward scale across prompts. GRPO served as the primary alignment algorithm for DeepSeek-R1~\cite{deepseek2025r1}.

% ----------------------------------------------------------------------------
\subsection{Regularisation: LoRA, Quantisation, and MoE Load Balancing}
\label{subsec:regularisation}

Low-Rank Adaptation (LoRA)~\cite{hu2022lora} enables parameter-efficient fine-tuning by decomposing the weight update as a low-rank product. For a pre-trained weight matrix $\weight_0 \in \R^{d \times d}$, the adapted weight is
\begin{equation}
\weight = \weight_0 + \tfrac{\alpha}{r}\,\bm{B}\bm{A},\;\; \bm{B} \!\in\! \R^{d \times r},\; \bm{A} \!\in\! \R^{r \times d},\; r \!\ll\! d.
\label{eq:lora}
\end{equation}
The pre-trained matrix $\weight_0$ is frozen; only the low-rank factors $\bm{A}$ (initialised from a Gaussian) and $\bm{B}$ (initialised to zero) are trained, reducing trainable parameters from $d^2$ to $2dr$ per matrix. For $d = 4096$ and $r = 16$, this yields a $128\times$ reduction. The low-rank constraint acts as an implicit regulariser, biasing the update towards the most significant task-specific adaptations while preserving pre-trained knowledge.

Quantisation reduces the numerical precision of weights and activations to lower memory and accelerate inference. INT8 quantisation~\cite{dettmers2022gpt3} preserves model quality nearly perfectly for most LLMs. More aggressive INT4 quantisation, as implemented by GPTQ~\cite{frantar2023gptq}, uses approximate second-order information to compensate remaining weights for quantisation error, achieving 4-bit precision with minimal perplexity degradation. FP8 training, as employed in DeepSeek-V3~\cite{deepseek2024v3}, performs matrix multiplications in 8-bit floating-point during both forward and backward passes using the E4M3 format for all tensors, made possible by a fine-grained quantisation strategy with tile-wise and block-wise scaling factors that accumulate in higher precision, thereby reducing compute cost during training itself.

In Mixture-of-Experts (MoE) models, a routing mechanism assigns each token to a subset of experts. Without regularisation, routing degenerates as a few experts receive disproportionate traffic. The standard auxiliary load-balancing loss~\cite{fedus2022switch} encourages uniform expert utilisation:
\begin{equation}
\loss_{\mathrm{balance}} = \alpha_{\mathrm{aux}} \cdot E \cdot \sum_{i=1}^{E} f_i \cdot p_i,
\label{eq:moe_balance}
\end{equation}
where $f_i$ is the fraction of tokens routed to expert $i$, $p_i$ is the average routing probability for expert $i$, and $\alpha_{\mathrm{aux}}$ controls the penalty strength. Tuning $\alpha_{\mathrm{aux}}$ is delicate: too large forces uniform routing regardless of input content, while too small allows degenerate collapse. DeepSeek-V3~\cite{deepseek2024v3} introduced an auxiliary-loss-free strategy that instead adds a dynamically adjusted bias $b_i$ to each expert's routing score before top-$k$ selection. The bias is increased for underutilised experts and decreased for overloaded ones, achieving balanced routing without introducing any gradient that conflicts with the primary training objective.

% ---- Figure: Three-Stage Training Pipeline ----
\begin{figure*}[t]
\centering
\definecolor{s1col}{RGB}{41,98,255}
\definecolor{s2col}{RGB}{0,150,80}
\definecolor{s3col}{RGB}{192,57,43}
\begin{tikzpicture}[
    node distance=0.6cm and 1.6cm,
    stage/.style={
        rectangle,
        rounded corners=8pt,
        minimum width=3.0cm,
        minimum height=1.6cm,
        align=center,
        font=\small\sffamily,
        line width=1.2pt,
        drop shadow={shadow xshift=1pt, shadow yshift=-1pt, opacity=0.18}
    },
    arrow/.style={
        -{Stealth[length=8pt, width=6pt]},
        line width=1.4pt,
        deepblue!55
    },
    datalabel/.style={
        font=\footnotesize\sffamily\itshape,
        text=gray!70,
    }
]

% Stage 1: Pre-training
\node[stage, fill=s1col!10, draw=s1col!65] (pt) {
    Pre-training\\[4pt]
    {\footnotesize $\displaystyle\loss = -\sum_t \log p_{\param}(x_t \mid x_{<t})$}
};

% Stage 2: Supervised Fine-Tuning
\node[stage, fill=s2col!10, draw=s2col!65, right=1.6cm of pt] (sft) {
    Supervised\\Fine-Tuning\\[4pt]
    {\footnotesize Instruction-response pairs}
};

% Stage 3: Alignment
\node[stage, fill=s3col!10, draw=s3col!65, right=1.6cm of sft] (align) {
    Alignment\\[4pt]
    {\footnotesize RLHF\,/\,DPO\,/\,GRPO}
};

% Arrows between stages
\draw[arrow] (pt.east) -- (sft.west)
    node[midway, above=3pt, font=\footnotesize\sffamily, text=deepblue!75] {Base model};
\draw[arrow] (sft.east) -- (align.west)
    node[midway, above=3pt, font=\footnotesize\sffamily, text=deepblue!75] {SFT model};

% Output
\node[right=1.0cm of align, font=\small\sffamily, align=center, text=s3col!75] (out) {Aligned\\Model};
\draw[arrow, s3col!55] (align.east) -- (out.west);

% Data labels below
\node[datalabel, below=0.6cm of pt] {Trillions of tokens};
\node[datalabel, below=0.6cm of sft] {Thousands of curated examples};
\node[datalabel, below=0.6cm of align] {Preference comparisons};

\end{tikzpicture}
\caption{The three-stage training pipeline for modern large language models. Pre-training via next-token prediction establishes linguistic competence; supervised fine-tuning teaches instruction following; alignment via RLHF, DPO, or GRPO optimises against human preferences.}
\label{fig:training_pipeline}
\end{figure*}

\end{document}